\documentclass[
  journal=medium,
  manuscript=article-type,
  year=2020,
  volume=37,
]{cup-journal}

\usepackage{caption}
\usepackage{capt-of}
\usepackage{times}
\usepackage{latexsym}
\usepackage{comment}
\usepackage{amsmath}
\usepackage{graphicx}
\usepackage{subcaption}
\usepackage{hyperref}
\usepackage{array}
\usepackage[table]{xcolor}
\usepackage{placeins}
\usepackage[utf8]{inputenc}
\usepackage{xurl} 
\usepackage{hyperref}
\hypersetup{breaklinks=true}
\usepackage{placeins}

\usepackage{tabularx}
\usepackage{array}

\usepackage{inconsolata}

\usepackage{graphicx}
\usepackage{amssymb}
\usepackage{subcaption}
\usepackage{xurl}
\hypersetup{breaklinks=true}

\usepackage{amsmath}
\usepackage[nopatch]{microtype}
\usepackage{booktabs}

\usepackage{etoolbox}  
\usepackage{graphicx}
\usepackage{subcaption}

 \title{Evaluating Explainable AI Attribution Methods in Neural Machine Translation via Attention-Guided Knowledge Distillation}

\author{Aria Nourbakhsh}
\affiliation{Department of Computer Science, University of Luxembourg,  4364 Esch-sur-Alzette, Luxembourg}
\email[F. Author]{aria.nourbakhsh@uni.lu}

\author{Salima Lamsiyah}
\affiliation{Department of Computer Science, University of Luxembourg,  4364 Esch-sur-Alzette, Luxembourg}

\author{Adelaide Danilov}
\affiliation{Department of Computer Science, University of Luxembourg,  4364 Esch-sur-Alzette, Luxembourg}

\author{Christoph Schommer}
\affiliation{Department of Computer Science, University of Luxembourg,  4364 Esch-sur-Alzette, Luxembourg}

\keywords{Explainability, Sequence-to-Sequence Model, Attribution Maps, Transformer, Knowledge Distillation}

\begin{document}

\begin{abstract}
The study of the attribution of input features to the output of neural network models is an active area of research. While numerous Explainable AI (XAI) techniques have been proposed to interpret these models, the systematic and automated evaluation of these methods in sequence-to-sequence (seq2seq) models is less explored. This paper introduces a new approach for evaluating explainability methods in transformer-based seq2seq models, building upon forward simulation of XAI methods. We use teacher-derived attribution maps as a structured side signal to guide a student model, and quantify the utility of different attribution methods through the student’s ability to simulate targets. Using the Inseq library, we extract attribution scores over source–target sequence pairs and inject these scores into the attention mechanism of a student transformer model under four composition operators (addition, multiplication, averaging, and replacement). Across three language pairs (de–en, fr–en, ar–en) and attributions from Marian-MT and mBART models, Attention, Value Zeroing, and Layer Gradient $\times$ Activation consistently yield the largest gains in BLEU (and corresponding improvements in chrF) relative to baselines. In contrast, other gradient-based methods (Saliency, Integrated Gradients, DeepLIFT, Input $\times$ Gradient, GradientShap) lead to smaller and less consistent improvements. These results suggest that different attribution methods capture distinct signals and that attention-derived attributions better capture alignment between source and target representations in seq2seq models. Finally, we introduce an Attributor transformer that, given a source–target pair, learns to reconstruct the teacher’s attribution map. Our findings demonstrate that the more accurately the Attributor can reproduce attribution maps, the more useful an injection of those maps is for the downstream task. The source code can be found on GitHub.\footnote{\url{https://github.com/ariana2011/seq2seq_xai_attributions/}}.

\end{abstract}

\section{Introduction}

In recent years, natural language processing (NLP) generative models have advanced rapidly and found applications in a wide range of domains \autocite{chang2024survey, kalyan2024survey}. These models, especially transformer-based sequence-to-sequence (seq2seq) architectures \autocite{sutskever2014sequence, vaswani2017attention}, excel at capturing complex relationships between input and output sequences. However, they are typically built on complex neural network architectures, which are often described as `black boxes' due to their opaque internal mechanisms~\autocite{dayhoff2001artificial, burkart2021survey}. To address this challenge, the field of Explainable AI (XAI) has attracted researchers seeking to improve transparency and interpretability, and to explain models' behavior. One of the main objectives of XAI is to assess, quantify, or characterize the importance (or attribution) of input features in shaping the final outputs of these models~\autocite{arya2019one, vieira2022machine, SAEED2023110273}. Several XAI methods have been developed and applied specifically to NLP models to evaluate the contribution of input information to the model's output across various tasks~\autocite{madsen2022post,mohammadi2025explainabilitypracticesurveyexplainable}. Despite these advances, identifying which explanation methods more accurately reflect model reasoning remains challenging, especially in seq2seq settings, which are characterized by intricate encoding-decoding dynamics and many-to-many mappings~\autocite{li-etal-2020-evaluating,gurrapu2023rationalization}.

The calculation and extraction of the decision-making process of neural networks per input features, which are also applied to Neural Machine Translation (NMT) models~\autocite{li-etal-2019-word, he-etal-2019-towards,eksi-etal-2021-explaining,fomicheva-etal-2022-translation, perrella-etal-2024-beyond}, are often grouped into three broad families of gradient-based, model-based, and perturbation-based methods~\autocite{dwivedi2023explainable,sarti-etal-2023-inseq, fantozzi2024explainability}. Gradient-based approaches estimate the contribution of each input by computing derivatives of the model output with respect to the input or intermediate representations; examples include Saliency~\autocite{simonyan2014deepinsideconvolutionalnetworks} and Integrated gradients~\autocite{sundararajan2017axiomatic}. Model-based approaches rely on components that already produce interpretable signals, such as the attention mechanism. Perturbation-based methods, instead, modify or remove parts of the input and measure the resulting change in model output; LIME~\autocite{ribeiro2016should} and Value Zeroing \autocite{mohebbi-etal-2023-quantifying} fall into this category. However, the boundaries between these families are not always strict. Some techniques combine properties of multiple categories, such as GradientSHAP~\autocite{NIPS2017_8a20a862}, which blends gradient information with stochastic perturbations. Nevertheless, because these methods rest on different assumptions about how models encode and use information, they compute feature importance in different ways and can therefore produce divergent explanations for the same prediction, raising the question of which methods best reflect the model’s behavior.

Despite the proliferation of explainability methods, comprehensive and scalable evaluation of XAI methods in NLP remains limited. Current evaluation practices that rely on human-centered validation, which, although insightful, are costly and difficult to scale \autocite{leiter2203towards, kim2024human}. Automated evaluation frameworks that are common in computer vision \autocite{ribeiro2016should, chang2018explaining, hooker2019benchmark} are underrepresented in NLP and NMT, and existing work typically focuses on a small number of explanation methods. This signifies the need for systematic, model-based evaluation approaches capable of objectively comparing diverse explainability techniques in seq2seq settings. Prior evaluations have, for example, compared attribution maps with human-annotated word alignments (see Section~\ref{subsec:XAI_ev_related_work}). Yet, such alignments only approximate the underlying translation dynamics and may not represent the information flow within modern NMT systems.

In this work, we address these gaps by proposing an automated evaluation framework based on the simulatability of XAI methods \autocite{doshi2017towards,hase-bansal-2020-evaluating}, specifically designed to assess and compare multiple attribution methods within seq2seq models for NMT. 
Intuitively, if an attribution method captures a model’s input–output dependencies, it should provide useful guidance for a student model to make better predictions. We operationalize this idea through a teacher-student setup: attribution maps are extracted from a pre-trained teacher NMT model and injected into the attention mechanism of a smaller, untrained student model. Concretely, we treat the attribution maps as attention priors within the encoder-decoder architecture and explore several ways to combine them with the student's own attention scores. The resulting student performance provides an automated, task-specific measure for evaluating different explanation methods. Within this framework, higher-quality explanations produce more informative attribution maps, which in turn allow the student to make more accurate predictions, thereby serving as a proxy for judging the effectiveness of the XAI attribution method. We apply this framework across three language pairs, using Marian-MT ~\autocite{tiedemann-thottingal-2020-opus} and mBART~\autocite{liu2020multilingual} models' attributions.

As part of our analysis, we deliberately compute attribution maps with respect to the gold reference translation, and in another part, we compute the attributions based on the teacher's generation. This defines an oracle setting in which the explanation is allowed to depend on the true target sequence and the resulting translation. Therefore, the model sees the source and attribution maps for all target tokens during encoding and at each autoregressive step.
We use this oracle setup to address two questions. First, to what extent can attribution-guided attention priors help a student model reproduce the gold, human-generated target when the student model has access to the attributions? Second, given a fixed teacher-generated translation, which attribution methods produce maps that are most helpful for a student model to approximate that teacher, i.e., which explanations best capture the teacher's input-output behavior under our idealized conditions? In this way, the oracle setting serves as a controlled environment for comparing attribution methods, and we interpret the reported BLEU/chrF scores as relative indicators of explanation quality rather than as standard test-set performance. 

To interpret the discrepancy in student performance under different attribution methods, we propose a hypothesis that their behavior can be explained by the closeness of their mappings to what a transformer is able to produce. To investigate that, we design a separate encoder-decoder transformer, named \textit{Attributor}, and train it to reconstruct the teacher's attribution map based on the respective source-target pair. Our experiments confirm the claim and highlight that the Attributor's ability to reproduce scores of top-3 salient tokens per column of the attribution maps very strongly correlates with the student performance in the MT task utilizing those maps.

Beyond the primary goal of evaluating XAI attribution methods in NMT, this work also provides insight into the behavior of the attention mechanism itself. We show that attributions derived from the teacher model’s attention tend to be more effective in guiding the student model. It is aligned with the fact that attention maps are the easiest for the Attributor to reproduce.  We also observe an interesting and somewhat unintuitive pattern in how the student model responds to externally injected attribution signals. In particular, the intervention is more effective when applied to the encoder attention~(see Section \ref{sec:methodology}). A detailed discussion of these findings appears later in the paper.

In summary, the main contributions of this work are as follows:
\begin{itemize}

\item We propose an evaluation framework that uses knowledge distillation to systematically assess and compare explainability methods by integrating attribution explanations into seq2seq model architectures.

\item We conduct extensive experiments exploring multiple strategies for incorporating explanations within the Transformer architecture and systematically compare their effects on model performance across various language pairs.

\item We provide empirical evidence that XAI attribution methods influence the performance of seq2seq models. Our findings demonstrate that the quality and type of explanations can enhance or degrade model output relative to baseline models without attribution guidance.

\item Finally, we investigate reasons why each attribution mapping yields different results when used within the student model for NMT tasks and show a strong correlation of those results with the ability of a transformer to approximate top-3 salient scores per target token of such mappings.
\end{itemize}

The paper is organized as follows: Section~\ref{sec:related_work} reviews the background and related work. Section~\ref{sec:methodology} describes the proposed approach. Sections \ref{sec:results} and \ref{discussion} present and discuss the obtained results. In section \ref{sec:attributore} we present the Attributor network. Section~\ref{conclusion} concludes the paper and outlines the limitations of this work, highlighting directions for future research.

\section{Related work} \label{sec:related_work}

In this section, we first summarize related work on evaluating and analyzing XAI     attribution methods, with a particular focus on NMT. We then briefly review the seq2seq NMT architecture and outline the XAI methods used in this study.

\subsection{Evaluation of Explanations and their Application in NMT}\label{subsec:XAI_ev_related_work}

The XAI literature distinguishes several dimensions of what explanations can provide. A central distinction is between plausibility and faithfulness/fidelity: plausibility refers to how well an explanation aligns with human intuition, whereas faithfulness describes how accurately an explanation reflects the model’s actual decision-making process \autocite{arrieta2020explainable, jacovi-goldberg-2020-towards}. Doshi-Velez and Kim \autocite{doshi2017towards,doshi2018considerations} have proposed three approaches to evaluate XAI methods, one of which is a functionally grounded evaluation protocol. In this protocol, explanations are assessed by automatic task performance metrics rather than human judgments. This paradigm has been adopted in NLP for evaluating the faithfulness of saliency\footnote{In the broader literature, feature-importance methods are sometimes referred to collectively as “saliency methods.” In this paper, however, later, we use Saliency to denote a specific gradient-based attribution method.} methods. For example, \cite{arras-etal-2016-explaining,nguyen-2018-comparing,deyoung-etal-2020-eraser,atanasova-etal-2020-diagnostic,nauta2023anecdotal} proposed automatic, task-based metrics for classification models that quantify how well feature attributions capture model behavior.

Our focus is on the NMT task, in particular, transformer-based~\autocite{vaswani2017attention} seq2seq models~\autocite{sutskever2014sequence}. Compared to standard classification, NMT introduces additional challenges for explanation because it decomposes prediction into a sequence of conditional next-token decisions, one per decoding timestep, and the importance of each source token depends not only on the current target token but also on the previous target prefix \autocite{stahlberg2020neural,shakil2024abstractive}. This temporal and source–target coupling complicates both the design and the evaluation of token-level attribution methods.

A large body of work has studied the attention mechanism as an interpretable component of NMT, using encoder–decoder attention weights to estimate word importance and to approximate word alignments \autocite{ghader-monz-2017-attention,raganato-tiedemann-2018-analysis,kobayashi-etal-2020-attention,ferrando-costa-jussa-2021-attention-weights}. These studies show that attention patterns correlate with, but do not faithfully reproduce, traditional word alignments, and that attention weights also reflect how the model balances source information against the evolving target prefix. On the other hand, other work has questioned the plausibility and faithfulness of attention as an explanation, arguing that attention weights should be interpreted cautiously based on the task and model, and in some cases, augmented with more explicit attribution mechanisms \autocite{jain2019attention,meister-etal-2021-sparse,madsen-etal-2022-evaluating}.

Beyond attention, other explanation methods exist to compute per-token importance. The most common metrics to evaluate token importance are comprehensiveness (does removing the highlighted tokens reduce the model’s confidence or translation quality?) and sufficiency (are the highlighted tokens alone sufficient to preserve model performance?)~\autocite{deyoung-etal-2020-eraser,nauta2023anecdotal}. In NMT, such approaches have been explored to measure the drop in log-probability of the chosen next token after input perturbations or at the sequence level, where changes in BLEU scores after manipulating important source tokens, according to XAI methods, were used as a test bed for assessing whether attribution maps identify tokens that are necessary and/or sufficient for the model’s predictions~\autocite{he-etal-2019-towards,moradi-etal-2021-measuring}.

Word alignment \autocite{brown_word_alignment} has also been widely used as a proxy for the plausibility of model attributions in NMT. \textcite{li-etal-2019-word} showed that attention-based alignments have clear limitations, motivating alternative alignment models and prediction-difference techniques to improve alignment quality. In parallel, \textcite{zenkel2019addinginterpretableattentionneural} proposed augmenting NMT models with dedicated alignment layers, treating alignment as an auxiliary prediction task rather than a by-product of standard attention. Their approach improves alignment accuracy compared to classical tools such as GIZA++~\autocite{och2003systematic} and FastAlign~\autocite{dyer2013simple}. \textcite{ding-etal-2019-saliency} introduced saliency-driven, gradient-based methods that yield more interpretable alignment signals without modifying the underlying NMT architecture. Building on these ideas, \textcite{ferrando-costa-jussa-2021-attention-weights} analyzed encoder–decoder attention in detail, highlighting systematic alignment errors and proposing techniques that explicitly quantify the relative contributions of source and target contexts. \textcite{ferrando-etal-2022-towards} further developed ALTI+, an attention-rollout-based framework that traces contributions from both source and target contexts across layers in multilingual Transformer models. ALTI+ has been used as an internal explanation metric for downstream diagnostic tasks; for instance, \textcite{dale-etal-2023-detecting} employ ALTI+-based scores to detect and mitigate hallucinations in NMT outputs. Related work by \textcite{voita-etal-2021-analyzing} used layer-wise propagation (LPR) to analyze the intrinsic contributions of source and target contexts under different training regimes and dataset conditions, while \textcite{kobayashi-etal-2020-attention} performed a norm-based analysis of attention and transformed representations to study internal alignment mechanisms within Transformers. Closer to the current work, \textcite{li-etal-2020-evaluating} evaluated XAI methods in NMT by training surrogate models on the important words identified by the XAI methods and measuring the prediction success of each token $i$ based on the top-k tokens identified as the most contributing tokens to the generation of that token. Other work has explored and compared XAI methods such as gradient and perturbation methods to detect word-level translation errors in NMT \autocite{eksi-etal-2021-explaining, fomicheva-etal-2022-translation}.

\subsection{Simulatability of XAI Methods}

A complementary line of work evaluates explanations through simulatability. Simulatability is how well an explanation helps a user replicate a model’s behavior. \textcite{doshi2017towards,hase-bansal-2020-evaluating} propose human-grounded protocols in which participants are asked to predict a model’s output on a given input, first without and then with access to explanations. The underlying idea is that an explanation method is better if it improves human prediction accuracy of the model’s decisions. These studies implement such user-in-the-loop evaluations on tabular and text classification tasks, using common XAI techniques to generate explanations that are shown to human subjects.

Closer to our setting, \textcite{pruthi2022evaluating} worked on a related idea without relying on human annotators. Building on the notion of simulatability, they transfer important information learned from one model to another. Token-level importance scores are used to guide a new model on several classification tasks, and the resulting change in performance is taken as evidence for the usefulness of the original attributions. In other words, an explanation is considered higher quality if it can be leveraged to train another model that better simulates or reconstructs the original model’s behavior. Our work adopts a similar spirit of model-based simulatability, but in the more complex seq2seq NMT setting.

\subsection{Injection of Knowledge into the Attention Mechanism}

There is also some work on injecting external or structured linguistic knowledge directly into the attention mechanism~\autocite{jiao-etal-2020-tinybert,bai-etal-2022-enhancing,zhao2023knowledge}. In NMT \textcite{bugliarello-okazaki-2020-enhancing} incorporate syntactic information by encoding, for each token, the distance to the syntactic `parent' in a dependency tree, and using this signal to bias attention patterns. In their approach, knowledge injection is applied on the encoder side for the source sentence, enabling attention heads to exploit syntactic structure.~\textcite{slobodkin-etal-2022-semantics} augment encoder attention with syntactic and semantic information in the form of alignment-like constraints over the input. Their architecture modifies the attention computation so that heads are explicitly informed by these external signals, rather than learning them implicitly. They report that enriching attention with semantic information benefits translation quality. Similar to current work \textcite{nourbakhsh2025quantifying}
inject attributions in the form of hard alignments and compare them to linguistic information, such as POS and Dependency information. In this work, we deal with soft attributions and more diverse XAI methods.
These works suggest that attention can serve as a locus for the injection of task-relevant prior knowledge.

Bringing together these strands of work, we propose an evaluation framework for XAI attribution methods to inject attribution matrix scores into the attention mechanism and compare the resulting effects on the translation task itself. Our design is inspired by Remove and Retrain (ROAR) retraining paradigms~\autocite{hooker2019benchmark}, simulatability of XAI methods ~\autocite{hase-bansal-2020-evaluating}, and prior work on knowledge injection into attention, and it contributes to this line of research by providing a comprehensive, functionally grounded comparison of several attribution methods in the seq2seq NMT model.

\subsection{NMT with Sequence-to-sequence Models}
Seq2seq models \autocite{sutskever2014sequence,bahdanau2016neuralmachinetranslationjointly}, originally introduced for machine translation \autocite{cho-etal-2014-learning}, are conditional language models that learn to generate the target sentence token by token, conditioned on the source sentence and the previously generated target tokens. The original Transformer model, a precursor to encoder-based classifiers and decoder-based large language models, was an encoder–decoder \autocite{vaswani2017attention}:

An NMT encoder–decoder model operates on two sequences:
\[
\mathbf{x} = (x_1, x_2, \dots, x_{T_x}), \quad 
\mathbf{y} = (y_1, y_2, \dots, y_{T_y}).
\]

Where $\mathbf{x}$ and $\mathbf{y}$ are the source and target sequences, an NMT model with a seq2seq architecture defines a conditional probability distribution over the target sequence given the source sequence and previous target tokens:
\[
p(\mathbf{y} \mid \mathbf{x}; \theta) 
= \prod_{t=1}^{T_y} p\bigl(y_t \mid y_{<t}, \mathbf{x}; \theta\bigr),
\]
where $y_{<t} = (y_1, \dots, y_{t-1})$ and $\theta$ are all model parameters.

The encoder maps the source sequence to a sequence of continuous contextual representations based on stacked self-attention and feed-forward layers.

\[
\mathbf{H} = ( \mathbf{h}_1, \dots, \mathbf{h}_{T_x} ),
\]
 The decoder is another neural network that, at each time step $t$, takes as input the previously generated target tokens (through masked self-attention over $y_{<t}$) and attends to the encoder representations $\mathbf{H}$ (via cross-attention), producing a decoder representation $\mathbf{s}_t$ from which the next-token distribution $p(y_t \mid y_{<t}, \mathbf{x}; \theta)$ is computed.

Given a training corpus
\[
\mathcal{D} = \{(\mathbf{x}^{(n)}, \mathbf{y}^{(n)})\}_{n=1}^N
\]
of source–target pairs, model parameters $\theta$ are typically learned by maximizing the conditional log-likelihood:
\[
\mathcal{L}(\theta)
= - \sum_{n=1}^N \sum_{t=1}^{T_y^{(n)}} 
\log p\bigl(y_t^{(n)} \mid y_{<t}^{(n)}, \mathbf{x}^{(n)}; \theta\bigr).
\]
During training, teacher forcing is commonly used, meaning that the decoder receives the ground-truth previous token $y_{t-1}^{(n)}$ as input when predicting $y_t^{(n)}$.

\subsection{AI Explainability Methods}\label{exp-method}

\begin{figure}[!t]
  \centering
  \begin{subfigure}{1\textwidth}
    \includegraphics[width=\textwidth]{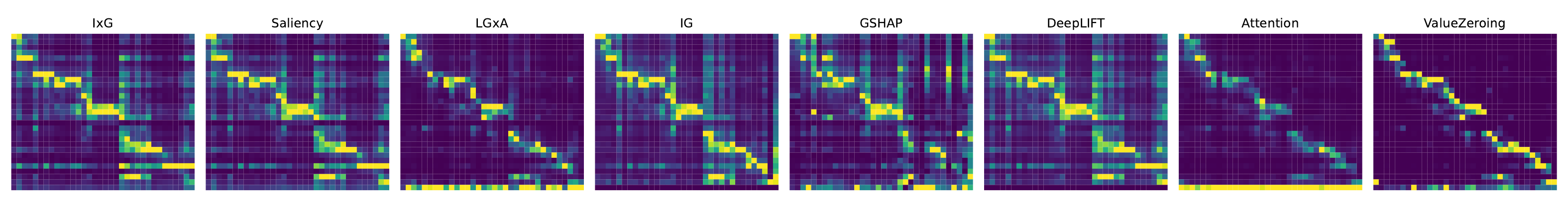}
    \caption{Marian-MT model.}
  \end{subfigure}

  \vspace{1em}

  \begin{subfigure}{1\textwidth}
    \includegraphics[width=\textwidth]{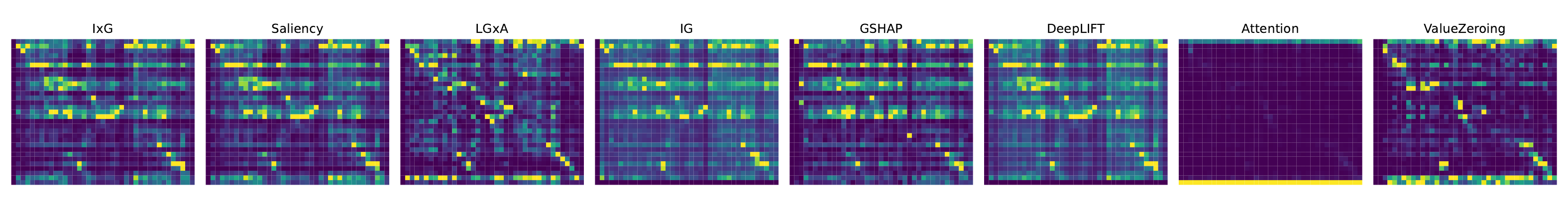}
    \caption{mBART model.}
\end{subfigure}

  \caption{An example of attribution maps derived from different XAI methods. For the source sentence \textit{'Dann gibt es noch Anbieter, die kaum Fahrraderfahrung, jedoch gute Fernostkontakte haben und so an günstige E-Bikes kommen.'} and the target \textit{'Then there are suppliers with little or no experience in the bicycle industry but good contacts in the Far East, thus giving them access to low-cost e-bikes.'}. In the heatmaps, the rows correspond to source tokens, and the columns to target tokens. The heatmaps are generated from the normalized columns using the MinMax normalizer. }
  \label{fig:att-heatmap}
\end{figure}

XAI attribution methods can be broadly categorized into three main types: gradient-based, internal model-based, and perturbation-based approaches. Below, we briefly describe the attribution methods used in this work to extract attribution maps. To extract these attribution maps we used the Inseq Python library ~\autocite{sarti-etal-2023-inseq}.

\paragraph{Saliency:} 
The saliency method was originally introduced for image classification tasks using convolutional neural networks (CNNs) and is one of the earliest gradient-based explanation techniques~\autocite{simonyan2014deepinsideconvolutionalnetworks}. It treats a trained neural network as a locally linear function of its input around a given example. The primary objective is to identify which pixels in an input image are most influential for a particular class prediction. The `saliency map' is defined as the gradient of the class score with respect to each input dimension, often visualized as the element-wise magnitude of this gradient. Intuitively, if a small change in a particular input dimension (e.g., a pixel intensity) leads to a large change in the class score, that dimension is considered important for the model’s decision for class \(c\).

Let \(\mathbf{x} \in \mathbb{R}^d\) be an input and \(S_c(\mathbf{x})\) the score for class \(c\).
The saliency map for class \(c\) at \(\mathbf{x}_0\) is defined as:
\begin{equation}
M_c^{(i)}(\mathbf{x}_0)
= \frac{\partial S_c(\mathbf{x}_0)}{\partial x_i},
\qquad i=1,\dots,d.
\label{eq:saliency}
\end{equation}
In NLP, \(\mathbf{x}\) corresponds to token embeddings; we aggregate per-dimension gradients to obtain a scalar attribution per token (see Section~\ref{sec:methodology}).

\paragraph{Input $\times$ Gradient (I$\times$G):}
It is a simple extension of saliency that combines information about how sensitive a prediction is to a feature with how strongly that feature is present in the input \autocite{denil2014extraction}. As in the saliency method, it considers the gradient of the class score with respect to the input, but instead of using the gradient alone, each input dimension is weighted by its own value. Intuitively, a feature should only be considered important if (i) small changes in that feature have a large effect on the score, and (ii) the feature is actually active in the current example. Raw gradients ignore the importance of the feature itself. I$\times$G takes this information into account by scaling the gradient by the input. In image models, this corresponds to weighting pixel-wise gradients by the pixel intensities. 
Let \(\mathbf{x} \in \mathbb{R}^d\) be an input and \(S_c(\mathbf{x})\) the score for class \(c\).
The Input $\times$ Gradient attribution for class \(c\) at \(\mathbf{x}_0\) is defined as:
\begin{equation}
M_c^{(i)}(\mathbf{x}_0)
= x_{0,i}\,\frac{\partial S_c(\mathbf{x}_0)}{\partial x_i},
\qquad i=1,\dots,d.
\label{eq:ixg}
\end{equation}

\paragraph{Layer Gradient $\times$ Activation (LG$\times$A):}
LG$\times$A applies I$\times$G to a chosen hidden layer. Let \(\mathbf{h}(\mathbf{x}_0)\in\mathbb{R}^{m}\) be its activation vector; the attribution for unit \(j\) is:

\begin{equation}
M_{c}^{(j)}(\mathbf{x}_0)
= h_j(\mathbf{x}_0)\,\frac{\partial S_c(\mathbf{x}_0)}{\partial h_j},
\qquad j=1,\dots,m.
\label{eq:lgxa}
\end{equation}

\paragraph{Integrated Gradients (IG):}
IG is motivated by two axioms: 
(i) Sensitivity, which requires that features responsible for a change in the model output relative to a baseline receive non-zero attribution, and 
(ii) Implementation invariance, which requires that attributions depend only on the input-output function \(S(\mathbf{x})\), not on a particular network parameterization~\autocite{sundararajan2017axiomatic}.
The authors define IG as: 

\begin{equation}
\mathrm{IG}_i(\mathbf{x}_0;\mathbf{x}')
= (x_{0,i}-x'_i)\int_{0}^{1}
\frac{\partial S_c\!\left(\mathbf{x}' + \alpha(\mathbf{x}_0-\mathbf{x}')\right)}{\partial x_i}\,d\alpha.
\label{eq:ig}
\end{equation}

Given an input \(\mathbf{x}\) and a baseline \(\mathbf{x}'\) representing the absence of information (e.g., the zero vector), IG attributes feature \(i\) by integrating gradients along the straight-line path from \(\mathbf{x}'\) to \(\mathbf{x}\), capturing how the prediction changes as the input moves from the baseline to the actual example.

\paragraph{Gradient SHAP (GSHAP):} GSHAP mixes IG with SHAP and estimates SHAP-style~\autocite{NIPS2017_8a20a862} attributions by averaging gradients over randomized reference points. For each of \(n\) samples, it adds small noise to the input, draws a random baseline from a set of baselines, and picks a random interpolation point along the straight line to compute the gradient\footnote{https://github.com/shap/shap} .

\paragraph{Deep Learning Important Features (DeepLIFT):}
DeepLIFT explains a prediction by comparing the network’s response on an input \(\mathbf{x}_0\) to a reference input \(\mathbf{x}'\)~\autocite{shrikumar2019learningimportantfeaturespropagating}. Instead of raw gradients, it propagates \emph{differences from reference} layer by layer and assigns contribution scores to input features that sum to the output difference \(S_c(\mathbf{x}_0)-S_c(\mathbf{x}')\), which helps mitigate gradient saturation and captures both positive and negative influences more reliably than standard gradient-based saliency methods.

\paragraph{Attention:}
In the Transformer model~\autocite{vaswani2017attention}, each token representation \(x_t\in\mathbb{R}^{d_{\text{model}}}\) is linearly projected into a query, key, and value using learned parameter matrices \(W_Q,W_K,W_V\in\mathbb{R}^{d_{\text{model}}\times d_k}\). Scaled dot-product attention computes similarities between queries and keys, normalizes them with a softmax to obtain attention weights, and then uses these weights to form weighted sums of the values. In multi-head attention, \(H\) such projections \(\{W_Q^{(h)},W_K^{(h)},W_V^{(h)}\}_{h=1}^{H}\) are used in parallel, and the concatenated head outputs are linearly projected back to the model dimension with a learned matrix \(W_O\in\mathbb{R}^{(H d_k)\times d_{\text{model}}}\).

Given queries \(Q\in\mathbb{R}^{T_q\times d_k}\), keys \(K\in\mathbb{R}^{T_k\times d_k}\), and values \(V\in\mathbb{R}^{T_k\times d_v}\), scaled dot-product attention is
\begin{equation}
\mathrm{Attention}(Q,K,V)
= \mathrm{softmax}\!\left(\frac{QK^\top}{\sqrt{d_k}}\right)V.
\label{eq:attention}
\end{equation}
In \eqref{eq:attention}, the score matrix \(QK^\top\) can be interpreted as pairwise similarity scores between queries and keys; after softmax normalization, these become attention weights that determine how strongly each query attends to each key.

\paragraph{Value Zeroing (ValueZeroing):}
In the attention mechanism, the value vector \(V_j\) for token \(j\) carries contextual content that is mixed into the representation of other tokens via attention weights derived from the \(QK^\top\) scores. ValueZeroing is an ablation technique that quantifies how much a context token \(j\) contributes to the representation of an output token at position \(i\) by recomputing the model’s hidden representation at \(i\) after zeroing out the value vector of token \(j\), while keeping all keys and queries fixed~\autocite{mohebbi-etal-2023-quantifying}.

Let \(\tilde{\mathbf{x}}_i\) denote the original representation of the output token at position \(i\), and let \(\tilde{\mathbf{x}}_i^{\neg j}\) denote the representation obtained when the value vector of token \(j\) is replaced by the zero vector. The context-mixing score between output position \(i\) and token \(j\) is defined as the cosine distance between these two representations:
\begin{equation}
C_{i,j}
= 1 - \cos\!\bigl(\tilde{\mathbf{x}}_i^{\neg j},\,\tilde{\mathbf{x}}_i\bigr).
\label{eq:valuezeroing}
\end{equation}
Higher values of \(C_{i,j}\) indicate that token \(j\) induces a larger change in the output representation at position \(i\), and therefore has a stronger influence on that output.

In this subsection, we briefly summarised the XAI attribution methods used in our comparison. Our goal was to include representatives from all three major families of attribution methods while also respecting practical constraints on computation. In particular, many perturbation-based methods are computationally expensive, and generating their attribution maps at scale for our datasets would be infeasible within a reasonable runtime.

\section{Methodology}\label{sec:methodology}

\begin{figure*}[!t]
    \centering
    \begin{subfigure}[b]{0.4\textwidth}
        \centering
        \includegraphics[width=\linewidth]{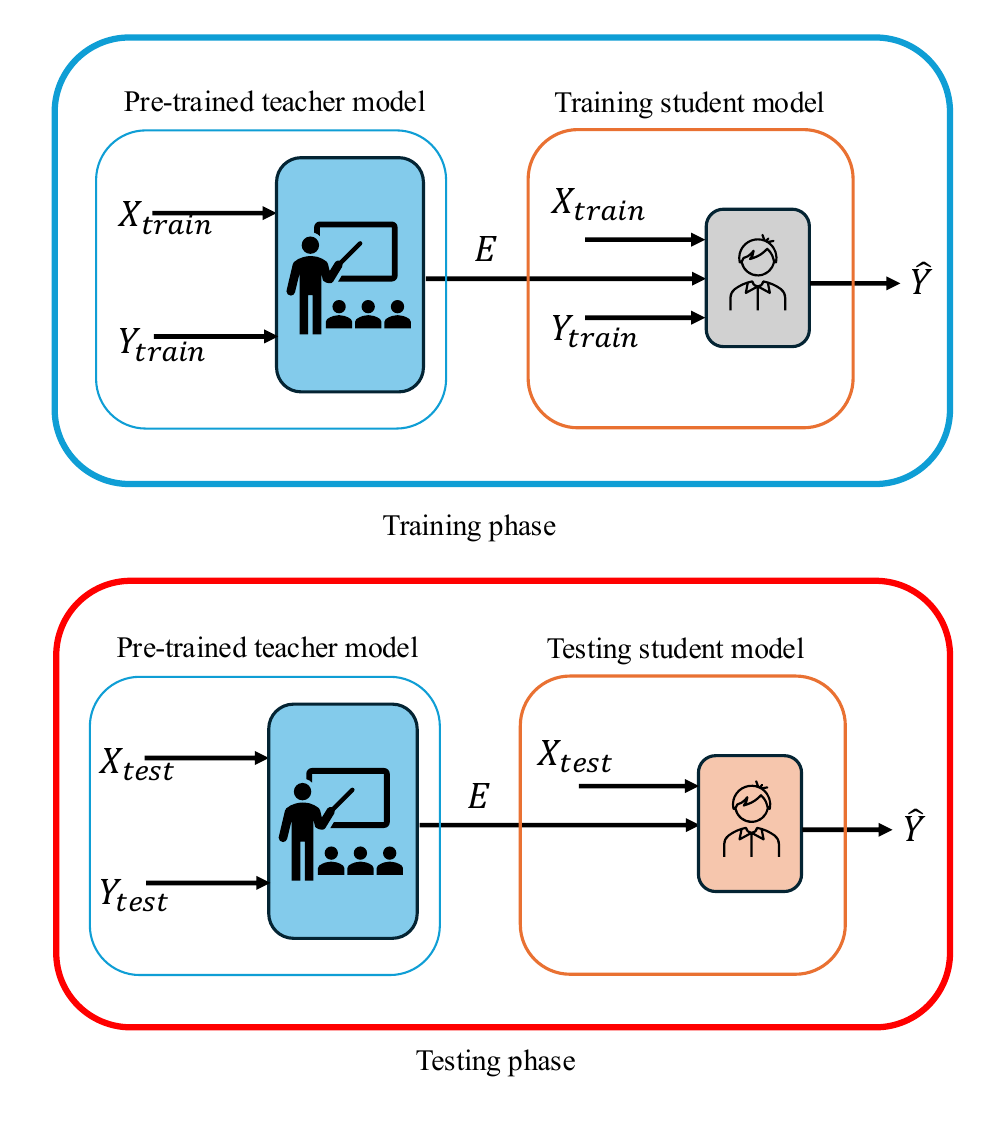}
        \label{fig:img1}
    \end{subfigure}
    \hspace{0.1\textwidth}
    \begin{subfigure}[b]{0.45\textwidth}
        \centering
        \includegraphics[width=\linewidth]{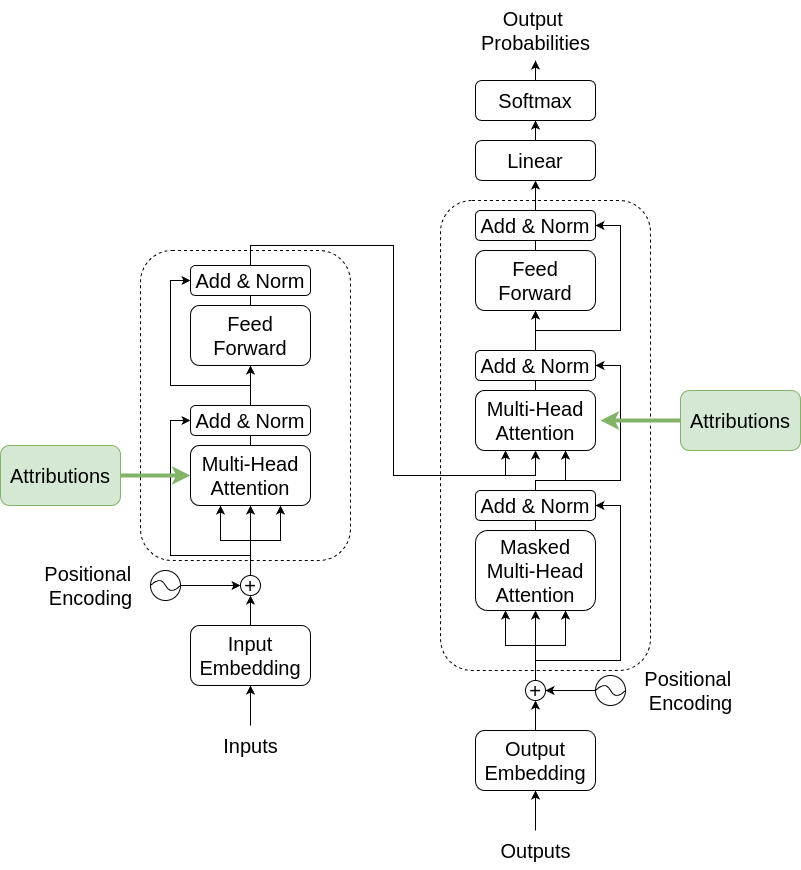}
        \label{fig:img2}
    \end{subfigure}
    \caption{\textbf{(a)} Illustrates the overall design of our approach. The input sequence and the gold output $(\mathbf{x}, \mathbf{y})$ are given to a teacher model, and their attributions $E$ are obtained. Then, a new untrained model is trained using the same $(\mathbf{x}, \mathbf{y}, E)$ triples. In the testing phase, the model gets the $(\mathbf{x}, E)\rightarrow \hat{\mathbf{y}}$. \textbf{(b)} Shows two places where we inject the attributions obtained from XAI methods.}
    \label{fig:architecture}
\end{figure*}

Inspired by the forward simulation of XAI methods \autocite{hase-bansal-2020-evaluating}, we design a pipeline to compare different explainability attribution maps based on their impact on model performance in the NMT task. For this purpose,
we use teacher-student knowledge distillation \autocite{hinton2015distilling}(Fig.~\ref{fig:architecture}). 
In the first step, we use the Inseq library\footnote{https://inseq.org/en/latest/} to extract input-output attribution maps using the eight explainability algorithms specified in subsection~\ref{exp-method}.  
The teacher model receives source-target sample pairs as input, and the output of Inseq is a set of attributions \( (\mathbf{x}, \mathbf{y}) \to E \) mapping the target to the source tokens . Then, the student models are trained under teacher forcing, receiving \( (\mathbf{x}, \mathbf{y}, E) \) for training. During testing, the student model gets the source token and attributions to predict the target \( (\mathbf{x}, E) \to \hat{\mathbf{y}} \).

Gradient-based attributions are in the shape \( e \in \mathbb{R}^{j \times k \times l} \), where \( j \) is the input sequence length, \( k \) is the output sequence length, and \( l \) is the hidden dimension of the model. Gradient-based methods get the weight of the gradient for each individual input feature in the vector space. We aggregate these values along the last dimension by getting the L2 norm $\lVert \mathbf{e}_i \rVert_{2}$ of the token vectors. L2 norm represents the magnitude of a vector and has a non-negative value\footnote{We did experiments with average pooling of the importance vectors; however, L2 norm representation always yielded better results.}. The final result is in the shape of \(e\in\mathbb {R}^{ j\times k} \). For all the attribution methods, we get the scores from the first layer of the transformer model. However, for LG$\times$A, where the attribution from the first layer is the same as I$\times$G, we chose to obtain attributions from the encoder's last layer. Prior work has examined which task-relevant properties are encoded at different layers~\autocite{langedijk-etal-2024-decoderlens}, but for our purposes, the encoder's final layer is the natural choice since its representations are the ones the decoder attends to when producing predictions.

The Attention attributions are extracted in the shape  $e \in \mathbb{R}^{j \times k \times n \times h}$, where \( j \) and \( k \) are the same as before, \( n \) represents the number of layers of the transformers, and \( h \) is the number of attention heads. We then compute the average along both of the last two axes to obtain a final shape of \( e \in \mathbb{R}^{j \times k} \). ValueZeroing yields the importance score for each layer, and therefore the scores are in the shape of  $e \in \mathbb{R}^{j \times k \times n}$. Similarly, we get the average of the scores on the last dimension to reach $e \in \mathbb{R}^{j \times k}$. To normalize and handle negative values in the attribution matrices, we apply the MinMaxScaler\footnote{https://scikit-learn.org/stable/modules/generated/sklearn.preprocessing.MinMaxScaler.html} to the columns of the attribution maps as follows:

\begin{equation}
\mathbf{e}'_{i,j} = \frac{\mathbf{e}_{i,j} - \min\limits_{i} (\mathbf{e}_{:,j})}{\max\limits_{i} (\mathbf{e}_{:,j}) - \min\limits_{i} (\mathbf{e}_{:,j})}
\end{equation}

Minmax transformation rescales values linearly to the $[0,1]$ interval while preserving their rank and relative structure. This choice over the softmax function avoids the additional inductive bias introduced by a softmax transformation, which converts scores into a probability distribution whose shape depends nonlinearly on their scale and forces scores to compete with one another. In our setting, we were interested in comparing the pattern and relative magnitude of attributions across methods and tokens, rather than imposing a probabilistic interpretation. Minmax normalization, therefore, preserves the geometry of the original attribution map \footnote{That said, the choice of the normalizer would affect the results; in our limited experiments, the minmax scaler scored better than the softmax; however, more investigations may be needed.}.

Next, the student model receives as input the triple \((\mathbf{x}, \mathbf{y}, \mathbf{E}')\), where \(\mathbf{E}'\) represents the (normalized) attribution map associated with the source–target pairs.
Subsequently, we perform four distinct operations on the pre-softmax attention scores
\(\mathbf{A}^{(h)} = \frac{Q^{(h)} {K^{(h)}}^\top}{\sqrt{d_k}}\) for each head \(h\):

\begin{align}
\tilde{\text{Attention}}(Q, K, V, \mathbf{E}')
&= \text{softmax}\!\left(
    f\!\left(
        \mathbf{A}, 
        \mathbf{E}'
    \right)
\right) V,
\label{eq:attn_attr}
\end{align}

where \(f\) is one of the following operations (applied elementwise):\\

\textbf{Addition ($+$)}: add attributions to the attention scores:
\begin{equation}
\tilde{\mathbf{A}}^{(h)} = \mathbf{A}^{(h)} + \mathbf{E}'.
\label{eq:addition}
\end{equation}

\textbf{Multiply ($\odot$)}: elementwise multiplication with the attention scores:
\begin{equation}
\tilde{\mathbf{A}}^{(h)} = \mathbf{A}^{(h)} \odot \mathbf{E}'.
\label{eq:multiplication}
\end{equation}

\textbf{Average ($\mu$)}: take the average of attributions and attention scores:
\begin{equation}
\tilde{\mathbf{A}}^{(h)} = \frac{\mathbf{A}^{(h)} + \mathbf{E}'}{2}.
\label{eq:average}
\end{equation}

\textbf{Replace ($R$)}: substitute the attention scores with attribution maps:
\begin{equation}
\tilde{\mathbf{A}}^{(h)} \leftarrow \mathbf{E}'.
\label{eq:replace}
\end{equation}

The last operation replaces $\mathbf{A}^{(h)}$ with $\mathbf{E}'$, completely substituting $\frac{QK^\top}{\sqrt{d_k}}$. The point of applying these simple element-wise operators is to treat the normalized attribution matrices $\mathbf{E}'$ as soft importance weights over the similarity scores $A^{(h)} = QK^\top$ in each attention head. Acting directly on $QK^\top$ (i.e., at the level of the similarity matrix before softmax) rather than on the values $V$ or the hidden states confines the intervention to the alignment structure between source and target tokens, which is precisely what attribution methods aim to characterize.

The four operators correspond to qualitatively different ways of using $\mathbf{E}'$ as a scaling factor for the similarity matrix. The multiplicative update $\tilde{\mathbf{A}}^{(h)} = \mathbf{A}^{(h)} \odot \mathbf{E}'$ implements a gating mechanism where attributions close to zero suppress specific query–key interactions, whereas attributions close to one leave them largely unchanged. Also, from another perspective, multiplication has a more dire effect if the attribution maps are incorrect. In contrast, the additive update $\tilde{\mathbf{A}}^{(h)} = \mathbf{A}^{(h)} + \mathbf{E}'$ behaves like a bias term on the similarity scores; when applied before softmax, it shifts probability mass toward positions preferred by the attribution map while preserving much of the relative structure induced by $QK^\top$. The averaging operator $\tilde{\mathbf{A}}^{(h)} = \tfrac{1}{2}(\mathbf{A}^{(h)} + \mathbf{E}')$ can be seen as a symmetric compromise between the model’s own attention and the external explanation, which smooths extreme scores from both matrices. Finally, the replacement variant, which feeds $\mathbf{E}'$ directly into the attention module in place of $\frac{QK^\top}{\sqrt{d_k}}$, provides a raw testbed in which the attribution map is treated as the only alignment signal. This yields an approximate lower bound on how well a given XAI method can affect the translation task.

By comparing these operators within the same teacher–student framework, we can probe two questions at once:  (i) whether they are strong enough to reliably gate or reroute information flow, and (ii) quantify the influence of attribution maps on the MT task. In all cases, applying $\mathbf{E}'$ as element-wise weights on $QK^\top$ makes a source-target token pair suitable for forward-simulation style evaluation, since changes in translation quality can be traced back to the manipulation of the attention matrix by the XAI attribution maps.

\section{Results}\label{sec:results}

In this section, we first describe our experimental setup, including datasets, metrics, and implementation details. We then analyze the results along four dimensions: (1) the comparison of eight attribution methods based on their influence on translation quality when integrated into the model; (2) the impact of the attribution injection location, comparing encoder self-attention and cross-attention modules; (3) the effect of selectively applying attributions to half of the attention heads (8 Heads vs. 4 Heads); (4) the ability of the student model to approximate the generation from the teacher model.

\subsection{Experimental Setup}

To evaluate the proposed pipeline, we train the Marian-MT model \autocite{tiedemann-thottingal-2020-opus}\footnote{https://huggingface.co/Helsinki-NLP} on three datasets from scratch. We choose two datasets belonging to a more closely related language family: German$\rightarrow$English (de-en) and French$\rightarrow$English (fr-en). For the third dataset, we choose Arabic$\rightarrow$English (ar-en) due to its encoding and linguistic differences from the target language. For de-en and fr-en, we use the WMT14 dataset \autocite{bojar2014findings}, and for ar-en, we use the UN Parallel Corpus~\autocite{ziemski-etal-2016-united}.

We select 220,000 sample pairs from each dataset and preprocess them to suit our experimental setup. Considering the numerous seq2seq models we train from scratch, we impose constraints to efficiently manage the training process. Specifically, we limit both the input and output sequences to at most 128 tokens. Additionally, we discard samples with fewer than ten tokens and filter out pairs where the input-to-output length ratio (or vice versa) exceeds $1.7$ for de-en and fr-en. Since the validation and test sets of the WMT datasets are relatively small, we select an additional 15,000 samples from their training sets (without overlap with our training data). The UN Parallel Corpus does not include separate validation and test sets, so we extract 15,000 samples from the main dataset for validation and testing.

We use two teacher models from which we extract attribution maps: a) monolingual Marian-MT systems and b) multilingual mBART-large~\autocite{tang2020multilingual} models. We focus most of our analysis on Marian-MT, since it is substantially smaller than mBART and, therefore, more tractable for computing and training with attribution maps at scale, given its smaller vocabulary size. Marian-MT is a Transformer model with six encoder layers and six decoder layers, each layer containing eight attention heads. In contrast, mBART-large has 12 encoder and 12 decoder layers, each with 12 attention heads. As a multilingual model, mBART uses a much larger subword vocabulary than Marian-MT (on the order of 500k vs.\ roughly 50k token types), which further increases the computational cost of attribution extraction. 
The student model shares the same overall architecture as Marian-MT, with feed-forward dimensionality 
\( d_{\mathrm{ff}} = 2048 \) and embedding dimensionality \( d_{\mathrm{model}} = 512 \). 
We limit the maximum sequence length to \( L_{\max} = 128 \), while keeping the number of layers 
\( N_{\mathrm{layers}} \) and attention heads \( H \) unchanged. Overall, Marian-MT has around 74 million parameters, and mBART has 610 million parameters.

We train the student Marian-MT models for 20 epochs and apply early stopping after three consecutive epochs without improvement in validation loss. The student model employs the Swish activation function, as proposed by \cite{ramachandran2017searching}, which has been shown to enhance training dynamics and convergence. We used 20 Nvidia V100 GPUs for our experiments.

Throughout our experiments, we report BLEU~\autocite{papineni2002bleu}, which measures n-gram overlap between system outputs and reference translations, and chrF~\autocite{popovic-2015-chrf}, a character-level n-gram F-score. This combination is particularly appropriate in our setting. We deliberately do not report semantic evaluation metrics such as COMET for two reasons. First, we conceptualize attribution maps as an auxiliary `memory' that the student model can use to reconstruct the future target sequence $Y$ from the teacher’s representations. Our objective is thus fidelity to the reference at the token level, for which BLEU and chrF are sufficient. Second, most of our source–target segments are relatively short, and our datasets are modest in size and length. Thus, COMET can be noisy and add limited additional insight beyond n-gram overlap. All in all, consistent changes across both BLEU and chrF provide sufficient evidence that attribution-guided attention priors affect the underlying translation behavior.

\subsection{Effectiveness of Attribution Methods (Encoder Attention)}
\label{subsec:self-att-res}
 
This analysis evaluates the impact of eight XAI attribution methods on translation quality, comparing models with injected attribution maps and the baseline. Tables \ref{tab:de_en_bleu_marian}, \ref{tab:fr_en_bleu_marian}, \ref{tab:ar_en_bleu_marian} present the BLEU and chrF of this setting and their delta compared to the baseline, using Marian-MT attributions, while \ref{tab:de_en_bleu_mabart}, \ref{tab:fr_en_bleu_mbart}, and \ref{tab:ar_en_bleu_mbart} represent the results of mBART attribution maps. The baseline models are trained and evaluated on the same dataset and settings, but without integrating attribution $E$. The difference between the baseline and the models with attribution maps is an indicator that XAI attribution maps changed the results relative to the baseline.

Starting with attribution maps extracted from Marian-MT, across all three language pairs, injecting attribution maps into the encoder's attention mechanism consistently improved translation quality over the baseline models. 
The highest gains come from Attention, ValueZeroing, and LG$\times$A.
The highest BLEU gains ranged up to +20.0 for de-en, +28.8 for fr-en, and +27.9 for ar-en, with corresponding chrF gains up to +13.3, +17.5, and +14.9, respectively. Other gradient-based methods score quite similarly to each other, and among them, GSHAP scores lowest across all three language pairs. Importantly, BLEU and chrF changes were aligned, and the configurations that improved BLEU nearly always improved chrF by a similar margin and vice versa, indicating that the gains are not metric-specific but reflect genuine translation quality.

For mBART attributions, there are more nuances. ValueZeroing scores higher than all other attribution methods in all cases. Attention maps that used to achieve higher scores on Marian-MT now score lower, and even for de-en, they degrade results in three out of four operators. Among the gradient-based methods, LG$\times$A scores highest for all the operators and all the language pairs. GSHAP yields the weakest results. Similarly, in mBART fr-en, Attention and ValueZeroing with the best operator reached 59.7 and 63.0 BLEU (+32.6 and +35.9), while GSHAP-based injections remained close to the baseline or produced only small improvements.

The choice of operator used to combine attributions with the original attention weights had a strong and systematic effect. Across attribution methods, language pairs, and models, the element-wise product operator ($\odot$) consistently yielded the highest BLEU and chrF scores, while averaging ($\mu$) was almost always the worst-performing operator, with $+$ and $R$ lying in between. For instance, in Marian-MT de-en with ValueZeroing, BLEU improved from 25.82 (baseline) to 41.8 (+16.0) with $+$, 40.2 (+14.4) with $\mu$, 41.9 (+16.1) with $R$, and 45.8 (+20) with $\odot$. An analogous pattern appeared with Marian-MT fr-en and ar-en, where $\odot$ systematically dominated the other operators for all strong attribution sources. The same trend held with mBART: for fr-en, Attention with $\odot$ reached 59.7 BLEU (+32.6) above the other operators, and ValueZeroing with $\odot$ reached 63.0 BLEU (+35.9) versus 51.2, 45.2, and 48.9 BLEU for $+,\mu$ and $R$ respectively. The magnitude of the difference between the operator for mBART Attention is higher than that for the others. We observed that mBART Attention assigns high values to the last source token (also visible in Figure \ref{fig:att-heatmap}), and that's why other operators can't contribute to the student model.

Finally, the patterns described above were consistent across de-en, fr-en, and ar-en for Marian-MT and across de-en and fr-en for mBART. Although the absolute baselines and magnitudes of improvement varied by language pair, the relative rankings of attribution sources and operators were nearly identical. This cross-lingual consistency suggests that certain XAI attribution methods provide more faithful target–source alignment signals than others, and that these signals can be leveraged as a reliable inductive bias for attention-guided knowledge distillation.

\FloatBarrier

\begin{table}[h]
\centering
\caption{BLEU and chrF scores for de-en Marian-MT attributions. Scores followed by $\Delta$ over the baseline.}
\label{tab:de_en_bleu_marian}
\setlength{\tabcolsep}{4.5pt} 
\begin{tabular}{lcccccccc}
\hline
\multicolumn{9}{c}{BLEU scores for de-en (Baseline: 25.82)} \\
\hline
Op. & I$\times$G & Saliency & LG$\times$A & IG & GSHAP & DeepLIFT & Attention & ValueZeroing \\
\hline
$+$ & 32.1\textsubscript{+6.3} & 32.3\textsubscript{+6.4} & 35.5\textsubscript{+9.6} & 33.6\textsubscript{+7.8} & 27.5\textsubscript{+1.7} & 32.8\textsubscript{+7.0} & 40.5\textsubscript{+14.7} & 41.8\textsubscript{+16.0} \\
$\mu$ & 29.5\textsubscript{+3.7} & 29.7\textsubscript{+3.9} & 33.5\textsubscript{+7.7} & 32.0\textsubscript{+6.2} & 26.8\textsubscript{+1.0} & 30.5\textsubscript{+4.7} & 37.3\textsubscript{+11.5} & 40.2\textsubscript{+14.4} \\
$\odot$ & 33.7\textsubscript{+7.9} & 33.8\textsubscript{+8.0} & 37.7\textsubscript{+11.8} & 35.4\textsubscript{+9.6} & 29.1\textsubscript{+3.3} & 34.2\textsubscript{+8.3} & \cellcolor{green!20}45.8\textsubscript{+20.0} & 45.8\textsubscript{+19.9} \\
R & 31.4\textsubscript{+5.6} & 31.7\textsubscript{+5.9} & 35.4\textsubscript{+9.5} & 33.0\textsubscript{+7.1} & 26.7\textsubscript{+0.9} & 32.3\textsubscript{+6.5} & 40.2\textsubscript{+14.3} & 41.9\textsubscript{+16.1} \\
\hline
\multicolumn{9}{c}{chrF scores for de-en (Baseline: 49.27)} \\
\hline
Op. & I$\times$G & Saliency & LG$\times$A & IG & GSHAP & DeepLIFT & Attention & ValueZeroing \\
\hline
$+$ & 53.2\textsubscript{+3.9} & 53.3\textsubscript{+4.0} & 55.6\textsubscript{+6.3} & 54.2\textsubscript{+4.9} & 50.0\textsubscript{+0.8} & 53.7\textsubscript{+4.4} & 58.7\textsubscript{+9.4} & 60.0\textsubscript{+10.7} \\
$\mu$ & 51.3\textsubscript{+2.0} & 51.4\textsubscript{+2.2} & 54.1\textsubscript{+4.8} & 53.1\textsubscript{+3.8} & 49.7\textsubscript{+0.4} & 52.0\textsubscript{+2.7} & 56.4\textsubscript{+7.1} & 58.8\textsubscript{+9.5} \\
$\odot$ & 54.3\textsubscript{+5.0} & 54.4\textsubscript{+5.1} & 57.2\textsubscript{+7.9} & 55.5\textsubscript{+6.3} & 51.1\textsubscript{+1.8} & 54.7\textsubscript{+5.4} & \cellcolor{green!20}62.6\textsubscript{+13.3} & 62.3\textsubscript{+13.0} \\
R & 52.7\textsubscript{+3.5} & 52.9\textsubscript{+3.6} & 55.5\textsubscript{+6.3} & 53.7\textsubscript{+4.5} & 49.6\textsubscript{+0.3} & 53.3\textsubscript{+4.1} & 58.5\textsubscript{+9.3} & 60.1\textsubscript{+10.8} \\
\hline
\end{tabular}
\end{table}


\begin{table}[h]
\centering
\caption{BLEU and chrF scores for fr-en Marian-MT attributions. Scores followed by $\Delta$ over the baseline.}
\label{tab:fr_en_bleu_marian}
\setlength{\tabcolsep}{4.5pt} 
\begin{tabular}{lcccccccc}
\hline
\multicolumn{9}{c}{BLEU scores for fr-en (Baseline: 27.01)} \\
\hline
Op. & I$\times$G & Saliency & LG$\times$A & IG & GSHAP & DeepLIFT & Attention & ValueZeroing \\
\hline
$+$ & 35.5\textsubscript{+8.5} & 35.7\textsubscript{+8.7} & 39.1\textsubscript{+12.1} & 37.6\textsubscript{+10.6} & 31.5\textsubscript{+4.5} & 35.9\textsubscript{+8.8} & 45.1\textsubscript{+18.1} & 47.8\textsubscript{+20.8} \\
$\mu$ & 32.8\textsubscript{+5.8} & 33.0\textsubscript{+6.0} & 36.0\textsubscript{+9.0} & 34.8\textsubscript{+7.8} & 28.8\textsubscript{+1.8} & 33.4\textsubscript{+6.4} & 40.8\textsubscript{+13.8} & 45.6\textsubscript{+18.6} \\
$\odot$ & 37.6\textsubscript{+10.6} & 37.9\textsubscript{+10.9} & 42.1\textsubscript{+15.1} & 40.1\textsubscript{+13.1} & 33.5\textsubscript{+6.5} & 37.7\textsubscript{+10.7} & \cellcolor{green!20}55.8\textsubscript{+28.8} & 54.1\textsubscript{+27.1} \\
R & 34.6\textsubscript{+7.5} & 34.6\textsubscript{+7.6} & 38.2\textsubscript{+11.2} & 37.2\textsubscript{+10.1} & 30.4\textsubscript{+3.4} & 35.1\textsubscript{+8.1} & 44.4\textsubscript{+17.4} & 47.1\textsubscript{+20.1} \\
\hline
\multicolumn{9}{c}{chrF scores for fr-en (Baseline: 53.01)} \\
\hline
Op. & I$\times$G & Saliency & LG$\times$A & IG & GSHAP & DeepLIFT & Attention & ValueZeroing \\
\hline
$+$ & 57.6\textsubscript{+4.6} & 57.6\textsubscript{+4.6} & 60.0\textsubscript{+7.0} & 58.8\textsubscript{+5.8} & 55.0\textsubscript{+2.0} & 57.8\textsubscript{+4.8} & 63.4\textsubscript{+10.4} & 65.5\textsubscript{+12.5} \\
$\mu$ & 55.8\textsubscript{+2.8} & 55.9\textsubscript{+2.9} & 57.8\textsubscript{+4.8} & 57.0\textsubscript{+4.0} & 53.3\textsubscript{+0.3} & 56.1\textsubscript{+3.1} & 60.4\textsubscript{+7.4} & 63.9\textsubscript{+10.9} \\
$\odot$ & 58.9\textsubscript{+5.9} & 59.1\textsubscript{+6.1} & 61.9\textsubscript{+8.9} & 60.4\textsubscript{+7.4} & 56.2\textsubscript{+3.2} & 58.9\textsubscript{+5.9} & \cellcolor{green!20}70.5\textsubscript{+17.5} & 69.1\textsubscript{+16.1} \\
R & 56.9\textsubscript{+3.9} & 56.9\textsubscript{+3.9} & 59.4\textsubscript{+6.4} & 58.5\textsubscript{+5.5} & 54.2\textsubscript{+1.2} & 57.3\textsubscript{+4.2} & 63.0\textsubscript{+9.9} & 65.1\textsubscript{+12.1} \\
\hline
\end{tabular}
\end{table}

\begin{table}
\centering
\caption{BLEU and chrF scores for ar-en Marian-MT attributions. Scores followed by $\Delta$ over the baseline.}
\label{tab:ar_en_bleu_marian}
\setlength{\tabcolsep}{4.5pt} 
\begin{tabular}{lcccccccc}
\hline
\multicolumn{9}{c}{BLEU scores for ar-en (Baseline: 40.68)} \\
\hline
Op. & I$\times$G & Saliency & LG$\times$A & IG & GSHAP & DeepLIFT & Attention & ValueZeroing \\
\hline
$+$ & 52.0\textsubscript{+11.2} & 51.7\textsubscript{+10.9} & 54.9\textsubscript{+14.1} & 52.1\textsubscript{+11.3} & 46.4\textsubscript{+5.6} & 52.7\textsubscript{+11.9} & 60.8\textsubscript{+19.9} & 62.9\textsubscript{+22.1} \\
$\mu$ & 47.3\textsubscript{+6.5} & 46.9\textsubscript{+6.1} & 51.3\textsubscript{+10.5} & 47.3\textsubscript{+6.5} & 43.7\textsubscript{+2.9} & 47.4\textsubscript{+6.6} & 55.7\textsubscript{+14.9} & 60.8\textsubscript{+20.0} \\
$\odot$ & 54.1\textsubscript{+13.2} & 54.2\textsubscript{+13.4} & 57.5\textsubscript{+16.7} & 54.2\textsubscript{+13.4} & 48.1\textsubscript{+7.3} & 54.2\textsubscript{+13.4} & 66.4\textsubscript{+25.6} & \cellcolor{green!20}68.7\textsubscript{+27.9} \\
R & 51.1\textsubscript{+10.3} & 51.0\textsubscript{+10.2} & 54.4\textsubscript{+13.6} & 51.1\textsubscript{+10.3} & 45.1\textsubscript{+4.3} & 51.9\textsubscript{+11.1} & 60.2\textsubscript{+19.4} & 62.7\textsubscript{+21.9} \\
\hline
\multicolumn{9}{c}{chrF scores for ar-en (Baseline: 66.07)} \\
\hline
Op. & I$\times$G & Saliency & LG$\times$A & IG & GSHAP & DeepLIFT & Attention & ValueZeroing \\
\hline
$+$ & 71.4\textsubscript{+5.3} & 71.2\textsubscript{+5.1} & 73.2\textsubscript{+7.1} & 71.4\textsubscript{+5.4} & 68.2\textsubscript{+2.1} & 71.8\textsubscript{+5.7} & 76.2\textsubscript{+10.1} & 77.6\textsubscript{+11.6} \\
$\mu$ & 68.7\textsubscript{+2.6} & 68.5\textsubscript{+2.4} & 71.0\textsubscript{+4.9} & 68.7\textsubscript{+2.6} & 66.8\textsubscript{+0.7} & 68.8\textsubscript{+2.7} & 73.2\textsubscript{+7.1} & 76.4\textsubscript{+10.3} \\
$\odot$ & 72.5\textsubscript{+6.4} & 72.5\textsubscript{+6.4} & 74.5\textsubscript{+8.4} & 72.4\textsubscript{+6.3} & 68.9\textsubscript{+2.9} & 72.5\textsubscript{+6.5} & 79.5\textsubscript{+13.4} & \cellcolor{green!20}81.0\textsubscript{+14.9} \\
R & 70.6\textsubscript{+4.6} & 70.6\textsubscript{+4.5} & 72.7\textsubscript{+6.6} & 70.7\textsubscript{+4.7} & 67.3\textsubscript{+1.3} & 71.1\textsubscript{+5.0} & 75.8\textsubscript{+9.7} & 77.4\textsubscript{+11.4} \\
\hline
\end{tabular}
\end{table}


\begin{table}
\centering
\caption{BLEU and chrF scores for de-en mBART attributions. Scores followed by $\Delta$ over the baseline.}
\label{tab:de_en_bleu_mabart}
\setlength{\tabcolsep}{5pt} 
\begin{tabular}{lcccccccc}
\hline
\multicolumn{9}{c}{BLEU scores for de-en (Baseline: 26.08)} \\
\hline
Op. & I$\times$G & Saliency & LG$\times$A & IG & GSHAP & DeepLIFT & Attention & ValueZeroing \\
\hline
$+$ & 24.9\textsubscript{-1.2} & 25.9\textsubscript{-0.2} & 29.8\textsubscript{+3.8} & 26.0\textsubscript{0.0} & 23.2\textsubscript{-2.8} & 26.2\textsubscript{+0.1} & 25.0\textsubscript{-1.1} & 43.4\textsubscript{+17.3} \\
$\mu$ & 23.1\textsubscript{-3.0} & 23.3\textsubscript{-2.8} & 24.6\textsubscript{-1.5}& 23.5\textsubscript{-2.6} & 22.6\textsubscript{-3.5} & 22.9\textsubscript{-3.2} & 24.2\textsubscript{-1.9} & 40.4\textsubscript{+14.3} \\
$\odot$ & 28.2\textsubscript{+2.1} & 28.4\textsubscript{+2.3} & 32.2\textsubscript{+6.2}& 28.8\textsubscript{+2.7} & 24.9\textsubscript{-1.2} & 28.4\textsubscript{+2.4} & 39.0\textsubscript{+12.9} & \cellcolor{green!20}51.4\textsubscript{+25.3} \\
R & 25.0\textsubscript{-1.1} & 25.4\textsubscript{-0.7} & 29.0\textsubscript{+2.9}& 26.2\textsubscript{+0.1} & 23.9\textsubscript{-2.2} & 25.4\textsubscript{-0.7} & 24.7\textsubscript{-1.4} & 41.4\textsubscript{+15.3} \\
\hline
\multicolumn{9}{c}{chrF scores for de-en (Baseline: 48.4)} \\
\hline
Op. & I$\times$G & Saliency & LG$\times$A & IG & GSHAP & DeepLIFT & Attention & ValueZeroing \\
\hline
$+$ & 46.8\textsubscript{-1.6} & 47.6\textsubscript{-0.8} & 50.5\textsubscript{+2.1} & 47.3\textsubscript{-1.1} & 45.8\textsubscript{-2.6} & 47.8\textsubscript{-0.6} & 46.6\textsubscript{-1.8} & 60.5\textsubscript{+12.1} \\
$\mu$ & 45.3\textsubscript{-3.1} & 45.5\textsubscript{-2.9} & 46.2\textsubscript{-2.2}& 45.6\textsubscript{-2.8} & 45.2\textsubscript{-3.2} & 45.4\textsubscript{-3.0} & 46.3\textsubscript{-2.1} & 58.3\textsubscript{+9.9} \\
$\odot$ & 49.2\textsubscript{+0.8} & 49.2\textsubscript{+0.8} & 52.2\textsubscript{+3.8}& 49.5\textsubscript{+1.1} & 46.9\textsubscript{-1.5} & 49.3\textsubscript{+0.9} & 56.7\textsubscript{+8.3} & \cellcolor{green!20}66.2\textsubscript{+17.8} \\
R & 47.0\textsubscript{-1.4} & 47.3\textsubscript{-1.1} & 49.9\textsubscript{+1.5} & 47.5\textsubscript{-0.9} & 46.4\textsubscript{-2.0} & 47.3\textsubscript{-1.1} & 46.3\textsubscript{-2.1} & 59.1\textsubscript{+10.7} \\
\hline
\end{tabular}
\end{table}

\begin{table}
\centering
\caption{BLEU and chrF scores for fr-en mBART attributions. Scores followed by $\Delta$ over the baseline.}
\label{tab:fr_en_bleu_mbart}
\setlength{\tabcolsep}{5pt} 
\begin{tabular}{lcccccccc}
\hline
\multicolumn{9}{c}{BLEU scores for fr-en (Baseline: 27.12)} \\
\hline
Op. & I$\times$G & Saliency & LG$\times$A & IG & GSHAP & DeepLIFT & Attention & ValueZeroing \\
\hline
$+$ & 33.2\textsubscript{+6.1} & 33.9\textsubscript{+6.8} & 38.9\textsubscript{+11.8} & 34.8\textsubscript{+7.7} & 28.4\textsubscript{+1.3} & 34.0\textsubscript{+6.9} & 31.8\textsubscript{+4.7} & 51.2\textsubscript{+24.1} \\
$\mu$ & 29.5\textsubscript{+2.4} & 30.9\textsubscript{+3.7} & 35.2\textsubscript{+8.1} &29.9\textsubscript{+2.7} & 26.8\textsubscript{-0.3} & 29.9\textsubscript{+2.8} & 30.8\textsubscript{+3.7} & 45.2\textsubscript{+18.1} \\
$\odot$ & 37.5\textsubscript{+10.4} & 38.0\textsubscript{+10.9} & 42.9\textsubscript{+15.8}& 37.5\textsubscript{+10.4} & 32.7\textsubscript{+5.6} & 38.2\textsubscript{+11.0} & 59.7\textsubscript{+32.6} & \cellcolor{green!20}63.0\textsubscript{+35.9} \\
R & 32.2\textsubscript{+5.1} & 32.4\textsubscript{+5.3} & 37.4\textsubscript{+10.2}&  32.3\textsubscript{+5.2} & 28.0\textsubscript{+0.9} & 32.7\textsubscript{+5.5} & 31.3\textsubscript{+4.2} & 48.9\textsubscript{+21.8} \\
\hline
\multicolumn{9}{c}{chrF scores for fr-en (Baseline: 53.01)} \\
\hline
Op. & I$\times$G & Saliency & LG$\times$A & IG & GSHAP & DeepLIFT & Attention & ValueZeroing \\
\hline
$+$ & 55.9\textsubscript{+2.9} & 56.3\textsubscript{+3.3} & 59.6\textsubscript{+6.6}& 56.9\textsubscript{+3.9} & 53.2\textsubscript{+0.2} & 56.3\textsubscript{+3.3} & 54.5\textsubscript{+1.4} & 67.7\textsubscript{+14.7} \\
$\mu$ & 53.6\textsubscript{+0.6} & 54.4\textsubscript{+1.4} & 57.1\textsubscript{+4.1}& 54.4\textsubscript{+1.4} & 52.3\textsubscript{-0.7} & 54.0\textsubscript{+1.0} & 54.0\textsubscript{+1.0} & 63.7\textsubscript{+10.7} \\
$\odot$ & 58.4\textsubscript{+5.4} & 58.7\textsubscript{+5.7} & 62.1\textsubscript{+9.0}& 58.3\textsubscript{+5.3}& 55.3\textsubscript{+2.3} & 58.8\textsubscript{+5.8} & 73.2\textsubscript{+20.2} & \cellcolor{green!20}75.5\textsubscript{+22.5} \\
R & 55.1\textsubscript{+2.1} & 55.2\textsubscript{+2.2} & 58.4\textsubscript{+5.4}& 55.4\textsubscript{+2.4} & 52.8\textsubscript{-0.2} & 55.4\textsubscript{+2.4} & 53.9\textsubscript{+0.9} & 66.1\textsubscript{+13.1} \\
\hline
\end{tabular}
\end{table}

\begin{table}
\centering
\caption{BLEU and chrF scores for ar-en mBART attributions. Scores followed by $\Delta$ over the baseline.}
\label{tab:ar_en_bleu_mbart}
\setlength{\tabcolsep}{5pt} 
\begin{tabular}{lcccccccc}
\hline
\multicolumn{9}{c}{BLEU scores for ar-en (Baseline: 40.68)} \\
\hline
Op. & I$\times$G & Saliency & LG$\times$A & IG & GSHAP & DeepLIFT & Attention & ValueZeroing \\
\hline
$+$ & 50.6\textsubscript{+9.9} & 51.2\textsubscript{+10.5} & 55.3\textsubscript{+14.6}& 50.5\textsubscript{+9.8} & 44.0\textsubscript{+3.3} & 50.3\textsubscript{+9.6} & 49.0\textsubscript{+8.3} & 63.6\textsubscript{+22.9} \\
$\mu$ & 45.1\textsubscript{+4.4} & 44.7\textsubscript{+4.0} & 49.7\textsubscript{+9.1}& 45.0\textsubscript{+4.3} & 42.1\textsubscript{+1.5} & 44.2\textsubscript{+3.6} & 45.7\textsubscript{+5.0} & 58.4\textsubscript{+17.7} \\
$\odot$ & 53.7\textsubscript{+13.0} & 54.5\textsubscript{+13.8} & 59.5\textsubscript{+18.9}& 53.5\textsubscript{+12.8} & 48.3\textsubscript{+7.6} & 54.2\textsubscript{+13.6} & 73.4\textsubscript{+32.7} &\cellcolor{green!20} 74.4\textsubscript{+33.7} \\
R & 48.3\textsubscript{+7.6} & 48.7\textsubscript{+8.0} & 52.8\textsubscript{+12.1}& 48.5\textsubscript{+7.7} & 42.0\textsubscript{+1.3} & 48.6\textsubscript{+7.9} & 47.6\textsubscript{+6.9} & 60.4\textsubscript{+19.7} \\
\hline
\multicolumn{9}{c}{chrF scores for ar-en (Baseline: 65.41)} \\
\hline
Op. & I$\times$G & Saliency & LG$\times$A & IG & GSHAP & DeepLIFT & Attention & ValueZeroing \\
\hline
$+$ & 69.9\textsubscript{+4.5} & 70.3\textsubscript{+4.9} & 72.7\textsubscript{+7.3}& 69.1\textsubscript{+3.7} & 66.6\textsubscript{+1.2} & 69.8\textsubscript{+4.4} & 68.6\textsubscript{+3.2} & 77.7\textsubscript{+12.3} \\
$\mu$ & 67.0\textsubscript{+1.6} & 66.8\textsubscript{+1.4} & 69.5\textsubscript{+4.1}& 67.3\textsubscript{+1.9}& 66.0\textsubscript{+0.6} & 66.6\textsubscript{+1.2} & 67.0\textsubscript{+1.5} & 74.4\textsubscript{+9.0} \\
$\odot$ & 71.4\textsubscript{+6.0} & 71.9\textsubscript{+6.5} & 75.1\textsubscript{+9.7}& 73.1\textsubscript{+7.7} & 68.1\textsubscript{+2.7} & 71.7\textsubscript{+6.3} & 83.8\textsubscript{+18.4} & \cellcolor{green!20}84.3\textsubscript{+18.9} \\
R & 68.2\textsubscript{+2.8} & 68.4\textsubscript{+3.0} & 70.9\textsubscript{+5.5}& 68.3\textsubscript{+2.9} & 65.0\textsubscript{-0.4} & 68.3\textsubscript{+2.9} & 67.1\textsubscript{+1.7} & 75.6\textsubscript{+10.2} \\
\hline
\end{tabular}
\end{table}

\FloatBarrier

\paragraph{Overall ranking of attribution methods (best to worst).}\leavevmode\par
For de-en, we obtain:

\begin{itemize}
    \item \textbf{Marian-MT}: Attention \(\approx\) ValueZeroing \(>\) LG$\times$A \(>\) IG \(>\) DeepLIFT \(>\) Saliency \(>\) I$\times$G \(>\) GSHAP
    \item \textbf{mBART}: ValueZeroing \(>\) Attention \(>\) LG$\times$A \(>\) IG \(>\) DeepLIFT \(>\) Saliency \(>\) I$\times$G \(>\) GSHAP
\end{itemize}

For fr-en, we obtain:
\begin{itemize}
    \item \textbf{Marian-MT}: Attention \(>\) ValueZeroing \(>\) LG$\times$A \(>\) IG \(>\) Saliency \(>\) DeepLIFT \(>\) I$\times$G \(>\) GSHAP
    \item \textbf{mBART}: ValueZeroing \(>\) Attention \(>\) LG$\times$A \(>\) DeepLIFT \(>\) Saliency \(>\) I$\times$G \(\approx\) IG \(>\) GSHAP
\end{itemize}

For ar-en, we obtain:
\begin{itemize}
    \item \textbf{Marian-MT}: ValueZeroing \(>\) Attention \(>\) LG$\times$A \(>\) DeepLIFT \(\approx\) IG  \(\approx\) Saliency \(>\) I$\times$G \(>\) GSHAP
    \item \textbf{mBART}: ValueZeroing \(>\) Attention \(>\) LG$\times$A \(>\) Saliency \(>\)DeepLIFT  \(>\) I$\times$G \(>\) IG \(>\) GSHAP
\end{itemize}

\subsection{Encoder Self-Attention vs. Cross-Attention Injection}

In contrast to the encoder self-attention experiments, injecting attributions into cross-attention rarely improved translation quality and often degraded it (Tables \ref{tab:de_en_bleu_marian_cross}, \ref{tab:fr_en_bleu_marian_cross}, and \ref{tab:ar_en_bleu_marian_cross}). Across de-en, fr-en, and ar-en, most attribution–operator combinations reduced BLEU and chrF relative to the baseline. The only consistent but modest gains were observed for de-en and fr-en when using gradient-based attributions (IG, LG$\times$A) with a multiplicative operator ($\odot$), yielding up to +4.7 BLEU and +1.0 chrF. ValueZeroing and teacher Attention, which were highly effective for encoder attention, provided little benefit in cross-attention and frequently harmed performance, while GSHAP was consistently detrimental. Replacement of cross-attention weights (R) was particularly destructive, often leading to large drops in BLEU and chrF. These patterns suggest that cross-attention is substantially more brittle than encoder attention, and its alignment structure can only tolerate very limited attribution guidance, and even then, the resulting gains are small and not always reflected consistently across evaluation settings. While the exact reason that attribution injection to the cross-attention does not work is obscure and hard to pinpoint, our primary hypothesis is that during autoregressive inference, due to a decoding strategy such as beam search, the fixed sequence of the attributions for generated targets confuses the model. In other words, the model can deviate from the ground-truth sentence during autoregressive generation, but attributions are added for the fixed sequence of tokens, which not only no longer matches the tokens generated by the model but also actively prevents it from being able to correct itself\footnote{To reduce energy consumption and training time, we omit training with mBART attributions when using 4-head encoder and cross-attention.}.

\FloatBarrier
\begin{table}
\centering
\caption{BLEU and chrF scores for de-en Cross-attention Marian-MT attributions (Baseline: 25.82). Scores followed by $\Delta$ over the baseline.}
\label{tab:de_en_bleu_marian_cross}
\setlength{\tabcolsep}{5pt} 
\begin{tabular}{lcccccccc}
\hline
\multicolumn{9}{c}{BLEU scores for de-en (Baseline: 25.82)} \\
\hline
Op. & I$\times$G & Saliency & LG$\times$A & IG & GSHAP & DeepLIFT & Attention & ValueZeroing \\
\hline
$+$ & 25.3\textsubscript{-0.5} & 25.4\textsubscript{-0.4} & 24.9\textsubscript{-0.9} & 24.8\textsubscript{-1.0} & 22.4\textsubscript{-3.4} & 25.6\textsubscript{-0.3} & 16.9\textsubscript{-8.9} & 23.9\textsubscript{-1.9} \\
$\mu$ & 24.8\textsubscript{-1.0} & 25.0\textsubscript{-0.8} & 26.0\textsubscript{+0.2} & 23.8\textsubscript{-2.0} & 22.3\textsubscript{-3.5} & 24.8\textsubscript{-1.1} & 20.0\textsubscript{-5.8} & 23.0\textsubscript{-2.8} \\
$\odot$ & 26.1\textsubscript{+0.3} & 26.1\textsubscript{+0.3} & \cellcolor{green!20}28.6\textsubscript{+2.8} & 28.5\textsubscript{+2.6} & 21.8\textsubscript{-4.1} & 26.3\textsubscript{+0.5} & 25.6\textsubscript{-0.2} & 24.7\textsubscript{-1.1} \\
R & 24.2\textsubscript{-1.6} & 24.2\textsubscript{-1.6} & 22.6\textsubscript{-3.2} & 23.0\textsubscript{-2.9} & 18.2\textsubscript{-7.6} & 24.4\textsubscript{-1.4} & 15.7\textsubscript{-10.1} & 21.0\textsubscript{-4.8} \\
\hline
\multicolumn{9}{c}{chrF scores for de-en (Baseline: 49.27)} \\
\hline
Op. & I$\times$G & Saliency & LG$\times$A & IG & GSHAP & DeepLIFT & Attention & ValueZeroing \\
\hline
$+$ & 47.5\textsubscript{-1.8} & 47.6\textsubscript{-1.6} & 48.7\textsubscript{-0.6} & 48.1\textsubscript{-1.1} & 46.0\textsubscript{-3.3} & 47.4\textsubscript{-1.9} & 44.1\textsubscript{-5.1} & 47.3\textsubscript{-1.9} \\
$\mu$ & 47.0\textsubscript{-2.3} & 47.3\textsubscript{-1.9} & 48.7\textsubscript{-0.5} & 47.8\textsubscript{-1.4} & 45.8\textsubscript{-3.5} & 47.0\textsubscript{-2.2} & 45.3\textsubscript{-4.0} & 46.6\textsubscript{-2.6} \\
$\odot$ & 47.6\textsubscript{-1.7} & 47.4\textsubscript{-1.9} & \cellcolor{green!20}50.3\textsubscript{+1.0} & 49.6\textsubscript{+0.4} & 45.1\textsubscript{-4.2} & 47.2\textsubscript{-2.1} & 47.9\textsubscript{-1.4} & 47.9\textsubscript{-1.4} \\
R & 45.9\textsubscript{-3.4} & 45.9\textsubscript{-3.4} & 46.9\textsubscript{-2.3} & 45.9\textsubscript{-3.4} & 41.4\textsubscript{-7.9} & 44.1\textsubscript{-5.2} & 42.1\textsubscript{-7.2} & 45.3\textsubscript{-4.0} \\
\hline
\end{tabular}
\end{table}

\begin{table}
\centering
\caption{BLEU and chrF scores for fr-en Cross-attention Marian-MT attributions (Baseline: 27.01). Scores followed by $\Delta$ over the baseline.}
\label{tab:fr_en_bleu_marian_cross}
\setlength{\tabcolsep}{5pt} 
\begin{tabular}{lcccccccc}
\hline
\multicolumn{9}{c}{BLEU scores for fr-en (Baseline: 27.01)} \\
\hline
Op. & I$\times$G & Saliency & LG$\times$A & IG & GSHAP & DeepLIFT & Attention & ValueZeroing \\
\hline
$+$ & 26.9\textsubscript{-0.1} & 26.9\textsubscript{-0.1} & 26.1\textsubscript{-0.9} & 28.2\textsubscript{+1.2} & 25.2\textsubscript{-1.8} & 25.3\textsubscript{-1.7} & 21.0\textsubscript{-6.0} & 27.6\textsubscript{+0.6} \\
$\mu$ & 26.4\textsubscript{-0.6} & 26.6\textsubscript{-0.4} & 26.9\textsubscript{-0.1} & 27.6\textsubscript{+0.6} & 24.7\textsubscript{-2.3} & 24.5\textsubscript{-2.6} & 22.2\textsubscript{-4.8} & 25.6\textsubscript{-1.4} \\
$\odot$ & 27.7\textsubscript{+0.6} & 27.8\textsubscript{+0.8} & 29.1\textsubscript{+2.1} & \cellcolor{green!20}31.7\textsubscript{+4.7} & 25.2\textsubscript{-1.8} & 27.7\textsubscript{+0.7} & 19.9\textsubscript{-7.2} & 26.3\textsubscript{-0.7} \\
R & 25.4\textsubscript{-1.6} & 25.5\textsubscript{-1.5} & 23.3\textsubscript{-3.7} & 26.2\textsubscript{-0.8} & 21.5\textsubscript{-5.5} & 25.1\textsubscript{-1.9} & 16.6\textsubscript{-10.4} & 24.5\textsubscript{-2.5} \\
\hline
\multicolumn{9}{c}{chrF scores for fr-en (Baseline: 53.01)} \\
\hline
Op. & I$\times$G & Saliency & LG$\times$A & IG & GSHAP & DeepLIFT & Attention & ValueZeroing \\
\hline
$+$ & 50.9\textsubscript{-2.1} & 50.8\textsubscript{-2.2} & 51.8\textsubscript{-1.2} & 52.1\textsubscript{-0.9} & 50.1\textsubscript{-2.9} & 49.4\textsubscript{-3.6} & 48.3\textsubscript{-4.7} & 53.0\textsubscript{-0.0} \\
$\mu$ & 50.4\textsubscript{-2.6} & 50.6\textsubscript{-2.4} & 51.8\textsubscript{-1.2} & 52.0\textsubscript{-1.0} & 49.7\textsubscript{-3.4} & 48.8\textsubscript{-4.2} & 48.2\textsubscript{-4.8} & 50.2\textsubscript{-2.9} \\
$\odot$ & 51.3\textsubscript{-1.7} & 51.2\textsubscript{-1.8} & 52.8\textsubscript{-0.2} & \cellcolor{green!20}53.5\textsubscript{+0.5} & 49.9\textsubscript{-3.1} & 51.1\textsubscript{-2.0} & 50.1\textsubscript{-2.9} & 51.1\textsubscript{-1.9} \\
R & 48.7\textsubscript{-4.3} & 48.5\textsubscript{-4.6} & 49.9\textsubscript{-3.1} & 49.9\textsubscript{-3.1} & 46.2\textsubscript{-6.8} & 47.9\textsubscript{-5.1} & 43.8\textsubscript{-9.2} & 49.0\textsubscript{-4.0} \\
\hline
\end{tabular}
\end{table}

\begin{table}
\centering
\caption{BLEU and chrF scores for ar-en Cross-attention Marian-MT attributions (Baseline: 40.68). Scores followed by $\Delta$ over the baseline.}
\label{tab:ar_en_bleu_marian_cross}
\setlength{\tabcolsep}{5pt} 
\begin{tabular}{lcccccccc}
\hline
\multicolumn{9}{c}{BLEU scores for ar-en (Baseline: 40.68)} \\
\hline
Op. & I$\times$G & Saliency & LG$\times$A & IG & GSHAP & DeepLIFT & Attention & ValueZeroing\\
\hline
$+$ & 36.2\textsubscript{-4.7} & 41.9\textsubscript{+1.1} & 35.9\textsubscript{-4.9} & 36.2\textsubscript{-4.7} & 36.0\textsubscript{-4.9} & 36.1\textsubscript{-4.8} & 36.1\textsubscript{-4.7} & 40.6\textsubscript{-0.2} \\
$\mu$ & 35.3\textsubscript{-5.5} & 41.7\textsubscript{+0.9} & 35.2\textsubscript{-5.6} & 35.3\textsubscript{-5.5} & 35.2\textsubscript{-5.6} & 35.3\textsubscript{-5.5} & 35.3\textsubscript{-5.5} & 40.5\textsubscript{-0.3} \\
$\odot$ & 28.5\textsubscript{-12.3} & 42.2\textsubscript{+1.4} & 26.5\textsubscript{-14.3} & 29.6\textsubscript{-11.2} & 28.3\textsubscript{-12.5} & 29.3\textsubscript{-11.5} & 26.3\textsubscript{-14.5} & \cellcolor{green!20}42.9\textsubscript{+2.1} \\
R & 14.3\textsubscript{-26.5} & 40.0\textsubscript{-0.8} & 13.9\textsubscript{-26.9} & 14.5\textsubscript{-26.3} & 14.3\textsubscript{-26.5} & 14.2\textsubscript{-26.6} & 13.7\textsubscript{-27.1} & 39.1\textsubscript{-1.7} \\
\hline
\multicolumn{9}{c}{chrF scores for ar-en (Baseline: 66.07)} \\
\hline
Op. & I$\times$G & Saliency & LG$\times$A & IG & GSHAP & DeepLIFT & Attention & ValueZeroing \\
\hline
$+$ & 62.7\textsubscript{-3.4} & 64.1\textsubscript{-1.9} & 62.7\textsubscript{-3.3} & 62.6\textsubscript{-3.5} & 62.7\textsubscript{-3.4} & 62.7\textsubscript{-3.4} & 62.8\textsubscript{-3.3} & 64.4\textsubscript{-1.7} \\
$\mu$ & 62.3\textsubscript{-3.8} & 64.1\textsubscript{-2.0} & 62.3\textsubscript{-3.8} & 62.3\textsubscript{-3.8} & 62.2\textsubscript{-3.8} & 62.3\textsubscript{-3.7} & 62.3\textsubscript{-3.8} & 64.1\textsubscript{-2.0} \\
$\odot$ & 55.4\textsubscript{-10.6} & 64.3\textsubscript{-1.7} & 53.7\textsubscript{-12.4} & 56.1\textsubscript{-9.9} & 55.4\textsubscript{-10.7} & 56.0\textsubscript{-10.1} & 53.4\textsubscript{-12.7} & \cellcolor{green!20}65.3\textsubscript{-0.8} \\
R & 43.4\textsubscript{-22.6} & 62.6\textsubscript{-3.5} & 43.3\textsubscript{-22.8} & 43.5\textsubscript{-22.6} & 43.4\textsubscript{-22.7} & 43.4\textsubscript{-22.7} & 43.2\textsubscript{-22.9} & 63.0\textsubscript{-3.1} \\
\hline
\end{tabular}
\end{table}

\FloatBarrier

\subsection{Selective Attribution Injection (8 vs. 4 Heads)}

In another setting, as an ablation study, we applied the attribution methods to only 4 heads out of the 8 heads of the attention only using Marian-MT attributions. We conduct these experiments only on the encoder side, as previous results showed that these attribution mappings can be useful at this part of the architecture. Tables ~\ref{tab:de_en_bleu_enc_4h},~\ref{tab:fr_en_bleu_enc_4h} and \ref{tab:ar_en_bleu_enc_4h}  show the results of this comparison. This analysis investigates how reducing the number of attention heads affects the model's performance when integrating attribution scores. By selectively applying attributions to only four heads (every other head), we assess whether information flow can still be captured and whether the model retains its translation quality. The changes in BLEU and chrF scores between 8-head and 4-head settings are relatively minor. However, some methods and operators show slight improvements. The results suggest that combining standard attention mechanisms with attribution-based operators can be a `best of both worlds' approach, as some heads learn the standard attention mechanisms while others utilize the attribution maps.

\subsection{Faithfulness (predicting the model's output)}

Up to this point, our experiments have relied on the assumption that attribution maps should help the student model better predict the gold (human) translation. Faithfulness, in contrast, relates to how precisely an attribution method captures the model’s internal reasoning process that produces a particular output \autocite{jacovi-goldberg-2020-towards}. Put differently, a faithful attribution provides a close approximation of the model’s actual decision-making procedure that produced the prediction.

To investigate this notion in a teacher-student setting, we replace the human reference with the teacher’s behavior as the supervision signal. Given a source input \(\mathbf{x}\), we run the teacher model to produce a translation \(\hat{\mathbf{y}}\) and compute the corresponding attribution map \(\hat{\mathbf{E}}\) with respect to this generated output, yielding training triples \((\mathbf{x}, \hat{\mathbf{y}}, \hat{\mathbf{E}})\). The student model is then trained on \((\mathbf{x}, \hat{\mathbf{y}}, \hat{\mathbf{E}})\) exactly as in our earlier experiments, injecting \(\hat{\mathbf{E}}\) into its attention mechanism. At test time, we provide the student model with \((\mathbf{x}, \hat{\mathbf{E}})\) and generate a student translation \(\hat{\mathbf{y}}'\), i.e., \((\mathbf{x}, \hat{\mathbf{E}}) \mapsto \hat{\mathbf{y}}'\). We then compare \(\hat{\mathbf{y}}'\) with \(\hat{\mathbf{y}}\) to quantify how well the student model can reproduce the teacher’s outputs when guided by a given attribution method. Under this setup, we hypothesize that attribution maps that better capture the teacher’s input--output dependencies will provide more useful guidance and, in turn, enable the student to generate translations \(\hat{\mathbf{y}}'\) that are closer to \(\hat{\mathbf{y}}\). The resulting agreement between \(\hat{\mathbf{y}}'\) and \(\hat{\mathbf{y}}\) therefore serves as an empirical proxy for the utility of the underlying attribution maps for simulating the teacher model.

Here, we present the results of this setup for the Marian-MT model in Tables~\ref{tab:de-en-generated}–\ref{tab:ar_en_generated}. The first observation is that the baseline scores are substantially higher for predicting the teacher's generation compared to the gold data (BLEU scores 54.55 vs 25.82 for de-en, 53.38 vs 27.01 for fr-en, 59.45 vs 40.68 for ar-en). As in the previous setting, injecting attribution maps into the encoder attention substantially improves performance over the baseline student for Attention and Value Zeroing. For de-en, the best configuration reaches 75.4~BLEU and 85.1~chrF (vs.\ 54.55/72.77 for the baseline), for fr-en 83.0~BLEU and 90.3~chrF (vs.\ 53.38/74.05), and for ar-en 82.8~BLEU and 90.0~chrF (vs.\ 59.45/77.96). In these cases, attribution-guided encoder attention markedly reduces the student–teacher gap, with absolute BLEU gains of +20--30 points, comparable to the gains observed when training on human references, albeit from a higher baseline.

At a high level, the hierarchy of attribution maps remains similar to the human-reference setting. Across de-en, fr-en, and ar-en, the largest gains again come from Attention and ValueZeroing, followed by the gradient-based methods, with GSHAP consistently being the weakest. Aggregating across language pairs, the approximate ordering under the faithfulness objective is
$\text{Attention} \approx \text{ValueZeroing} > \text{LG$\times$A} > \text{IG} > \text{Saliency} > \text{I$\times$G} \approx \text{DeepLIFT} > \text{GSHAP}$ with slight nuances for ar-en where IG ranks lower.
Because the baseline is already higher, some methods and operators fail to help the student model. For example, in Saliency for fr-en, addition does not change the result from the baseline, whereas multiplication yields improvements. I$\times$G and DeepLIFT can also degrade the results for fr-en with ($\mu$) operator. From the operator's point of view, the element-wise product operator ($\odot$) yields the largest changes, simple averaging ($\mu$) is almost always the weakest operator, and $+$ and $R$ lie in between. For example, in fr-en, the best $\odot$ configuration with ValueZeroing reaches 83.0~BLEU (+29.6) and 90.2~chrF (+16.2) relative to the baseline student, whereas the corresponding $+$, $\mu$, and $R$ configurations with ValueZeroing remain between 74.3--80.6~BLEU and 85.2--89.0~chrF. A similar pattern holds for de-en and ar-en. 

\FloatBarrier

\begin{table}[H]
\centering
\caption{BLEU and chrF scores for de-en generated by the Marian-MT model. Scores followed by $\Delta$ over the baseline.}
\label{tab:de-en-generated}
\setlength{\tabcolsep}{4.5pt} 
\begin{tabular}{lcccccccc}
\hline
\multicolumn{9}{c}{BLEU scores for de-en (Baseline: 54.55)} \\
\hline
Op. & I$\times$G & Saliency & LG$\times$A & IG & GSHAP & DeepLIFT & Attention & ValueZeroing \\
\hline
$+$ & 64.5\textsubscript{+9.9} & 64.6\textsubscript{+10.1} & 68.5\textsubscript{+13.9} & 65.5\textsubscript{+11.0} & 58.1\textsubscript{+3.5} & 65.1\textsubscript{+10.5} & 72.0\textsubscript{+17.4} & 73.7\textsubscript{+19.1} \\
$\mu$ & 58.4\textsubscript{+3.8} & 59.7\textsubscript{+5.2} & 65.2\textsubscript{+10.7} & 61.8\textsubscript{+7.3} & 55.1\textsubscript{+0.5} & 61.1\textsubscript{+6.6} & 68.8\textsubscript{+14.3} & 71.6\textsubscript{+17.0} \\
$\odot$ & 66.5\textsubscript{+11.9} & 66.6\textsubscript{+12.0} & 70.0\textsubscript{+15.5} & 67.3\textsubscript{+12.8} & 59.7\textsubscript{+5.1} & 66.3\textsubscript{+11.8} & \cellcolor{green!20}75.4\textsubscript{+20.9} & 74.2\textsubscript{+19.6} \\
R & 63.4\textsubscript{+8.8} & 63.1\textsubscript{+8.6} & 67.9\textsubscript{+13.3} & 64.4\textsubscript{+9.8} & 56.0\textsubscript{+1.4} & 64.2\textsubscript{+9.7} & 72.0\textsubscript{+17.4} & 73.4\textsubscript{+18.9} \\
\hline
\multicolumn{9}{c}{chrF scores for de-en (Baseline: 72.77)} \\
\hline
Op. & I$\times$G & Saliency & LG$\times$A & IG & GSHAP & DeepLIFT & Attention & ValueZeroing \\
\hline
$+$ & 78.3\textsubscript{+5.5} & 78.3\textsubscript{+5.6} & 80.8\textsubscript{+8.0} & 78.9\textsubscript{+6.2} & 74.4\textsubscript{+1.6} & 78.7\textsubscript{+6.0} & 82.8\textsubscript{+10.0} & 84.0\textsubscript{+11.2} \\
$\mu$ & 74.5\textsubscript{+1.7} & 75.4\textsubscript{+2.6} & 78.7\textsubscript{+5.9} & 76.6\textsubscript{+3.9} & 72.8\textsubscript{+0.0} & 76.2\textsubscript{+3.4} & 80.8\textsubscript{+8.0} & 82.6\textsubscript{+9.8} \\
$\odot$ & 79.5\textsubscript{+6.8} & 79.6\textsubscript{+6.9} & 81.8\textsubscript{+9.0} & 80.1\textsubscript{+7.3} & 75.3\textsubscript{+2.5} & 79.5\textsubscript{+6.7} & \cellcolor{green!20}85.1\textsubscript{+12.3} & 84.0\textsubscript{+11.2} \\
R & 77.6\textsubscript{+4.8} & 77.4\textsubscript{+4.7} & 80.4\textsubscript{+7.7} & 78.2\textsubscript{+5.4} & 73.2\textsubscript{+0.4} & 78.1\textsubscript{+5.3} & 82.8\textsubscript{+10.0} & 83.8\textsubscript{+11.0} \\
\hline
\end{tabular}
\end{table}

\begin{table}[H]
\centering
\caption{BLEU and chrF scores for fr-en generated by the Marian-MT model. Scores followed by $\Delta$ over the baseline.}
\label{tab:fr_en_generated}
\setlength{\tabcolsep}{4.5pt} 
\begin{tabular}{lcccccccc}
\hline
\multicolumn{9}{c}{BLEU scores for fr-en (Baseline: 53.38)} \\
\hline
Op. & I$\times$G & Saliency & LG$\times$A & IG & GSHAP & DeepLIFT & Attention & ValueZeroing \\
\hline
$+$ & 53.7\textsubscript{+0.3} & 53.6\textsubscript{+0.2} & 72.6\textsubscript{+19.2} & 70.7\textsubscript{+17.3} & 52.8\textsubscript{-0.6} & 53.6\textsubscript{+0.2} & 76.3\textsubscript{+22.9} & 80.6\textsubscript{+27.2} \\
$\mu$ & 52.8\textsubscript{-0.6} & 52.9\textsubscript{-0.4} & 63.6\textsubscript{+10.2} & 64.0\textsubscript{+10.7} & 52.4\textsubscript{-1.0} & 52.8\textsubscript{-0.6} & 69.5\textsubscript{+16.1} & 74.3\textsubscript{+20.9} \\
$\odot$ & 62.2\textsubscript{+8.8} & 73.4\textsubscript{+20.1} & 76.5\textsubscript{+23.1} & 75.8\textsubscript{+22.5} & 58.7\textsubscript{+5.3} & 62.9\textsubscript{+9.5} & 82.9\textsubscript{+29.5} & \cellcolor{green!20}83.0\textsubscript{+29.6} \\
R & 54.1\textsubscript{+0.7} & 66.5\textsubscript{+13.1} & 72.8\textsubscript{+19.4} & 70.9\textsubscript{+17.5} & 52.2\textsubscript{-1.2} & 54.2\textsubscript{+0.8} & 77.4\textsubscript{+24.0} & 80.5\textsubscript{+27.1} \\
\hline
\multicolumn{9}{c}{chrF scores for fr-en (Baseline: 74.05)} \\
\hline
Op. & I$\times$G & Saliency & LG$\times$A & IG & GSHAP & DeepLIFT & Attention & ValueZeroing \\
\hline
$+$ & 73.4\textsubscript{-0.7} & 73.3\textsubscript{-0.7} & 84.4\textsubscript{+10.3} & 83.2\textsubscript{+9.1} & 72.8\textsubscript{-1.3} & 73.2\textsubscript{-0.8} & 86.3\textsubscript{+12.3} & 89.0\textsubscript{+15.0} \\
$\mu$ & 73.1\textsubscript{-1.0} & 73.2\textsubscript{-0.8} & 79.5\textsubscript{+5.5} & 79.7\textsubscript{+5.7} & 72.9\textsubscript{-1.2} & 73.0\textsubscript{-1.0} & 82.5\textsubscript{+8.5} & 85.2\textsubscript{+11.1} \\
$\odot$ & 78.1\textsubscript{+4.1} & 84.9\textsubscript{+10.8} & 86.7\textsubscript{+12.6} & 86.2\textsubscript{+12.1} & 76.1\textsubscript{+2.0} & 78.6\textsubscript{+4.5} & \cellcolor{green!20}90.3\textsubscript{+16.2} & 90.2\textsubscript{+16.2} \\
R & 73.5\textsubscript{-0.5} & 80.8\textsubscript{+6.8} & 84.5\textsubscript{+10.5} & 83.2\textsubscript{+9.2} & 72.4\textsubscript{-1.7} & 73.6\textsubscript{-0.5} & 87.0\textsubscript{+12.9} & 88.9\textsubscript{+14.9} \\
\hline
\end{tabular}
\end{table}

\begin{table}[H]
\centering
\caption{BLEU and chrF scores for ar-en generated by the Marian-MT model. Scores followed by $\Delta$ over the baseline.}
\label{tab:ar_en_generated}
\setlength{\tabcolsep}{4.5pt} 
\begin{tabular}{lcccccccc}
\hline
\multicolumn{9}{c}{BLEU scores for ar-en (Baseline: 59.45)} \\
\hline
Op. & I$\times$G & Saliency & LG$\times$A & IG & GSHAP & DeepLIFT & Attention & ValueZeroing \\
\hline
$+$ & 69.1\textsubscript{+9.6} & 68.6\textsubscript{+9.1} & 73.1\textsubscript{+13.6} & 69.8\textsubscript{+10.3} & 63.7\textsubscript{+4.2} & 70.0\textsubscript{+10.6} & 76.8\textsubscript{+17.4} & 78.9\textsubscript{+19.4} \\
$\mu$ & 65.1\textsubscript{+5.6} & 64.6\textsubscript{+5.1} & 65.3\textsubscript{+5.9} & 63.8\textsubscript{+4.3} & 61.7\textsubscript{+2.3} & 65.1\textsubscript{+5.7} & 67.4\textsubscript{+7.9} & 75.1\textsubscript{+15.7} \\
$\odot$ & 72.6\textsubscript{+13.2} & 72.8\textsubscript{+13.3} & 75.6\textsubscript{+16.1} & 72.4\textsubscript{+13.0} & 66.7\textsubscript{+7.2} & 72.9\textsubscript{+13.5} & 81.6\textsubscript{+22.2} & \cellcolor{green!20}82.8\textsubscript{+23.3} \\
R & 68.9\textsubscript{+9.4} & 68.5\textsubscript{+9.0} & 72.7\textsubscript{+13.3} & 68.9\textsubscript{+9.4} & 63.3\textsubscript{+3.8} & 69.5\textsubscript{+10.1} & 77.0\textsubscript{+17.6} & 78.7\textsubscript{+19.2} \\
\hline
\multicolumn{9}{c}{chrF scores for ar-en (Baseline: 77.96)} \\
\hline
Op. & I$\times$G & Saliency & LG$\times$A & IG & GSHAP & DeepLIFT & Attention & ValueZeroing \\
\hline
$+$ & 82.5\textsubscript{+4.5} & 82.2\textsubscript{+4.2} & 84.8\textsubscript{+6.8} & 82.8\textsubscript{+4.8} & 79.5\textsubscript{+1.6} & 83.0\textsubscript{+5.1} & 86.7\textsubscript{+8.7} & 87.9\textsubscript{+10.0} \\
$\mu$ & 80.3\textsubscript{+2.4} & 80.1\textsubscript{+2.1} & 80.6\textsubscript{+2.6} & 79.6\textsubscript{+1.7} & 78.7\textsubscript{+0.7} & 80.4\textsubscript{+2.4} & 81.6\textsubscript{+3.6} & 85.8\textsubscript{+7.8} \\
$\odot$ & 84.4\textsubscript{+6.4} & 84.5\textsubscript{+6.5} & 86.0\textsubscript{+8.0} & 84.2\textsubscript{+6.2} & 81.0\textsubscript{+3.1} & 84.6\textsubscript{+6.6} & 89.3\textsubscript{+11.3} & \cellcolor{green!20}90.0\textsubscript{+12.0} \\
R & 82.2\textsubscript{+4.3} & 82.0\textsubscript{+4.0} & 84.5\textsubscript{+6.5} & 82.2\textsubscript{+4.2} & 79.3\textsubscript{+1.3} & 82.6\textsubscript{+4.6} & 86.7\textsubscript{+8.7} & 87.7\textsubscript{+9.7} \\
\hline
\end{tabular}
\end{table}
\FloatBarrier

\section{Discussion} \label{discussion}

Because these attribution mappings come from trained models on larger datasets, they encode a learned relation between the source and target pairs, which otherwise cannot be learned by the student model. We conjectured that a better learned mapping can contribute to better results, and hence it can be a way of comparison between XAI attribution methods. In subsection \ref{subsec:self-att-res}, we presented the main findings and contributions of this work. Expanding on our observations, we saw some consistent results for the injection of Marian-MT and mBART attributions to the encoder-attention, with some nuances for the mBART attributions. Particularly for the Marian-MT attributions, we saw improvements in almost all the attribution maps. As a sanity check, we conducted the same experiments with two settings: 1) Injecting random attribution maps taken from the uniform distribution and 2) nearly diagonal matrices. Table \ref{tab:rand-diagonal} shows the results of these experiments:

\begin{table}[H]
\centering
\caption{BLEU scores for random attribution matrices (left) and diagonal attribution matrices (right). Scores followed by $\Delta$ over the baseline.}
\label{tab:rand-diagonal}
\begin{tabular}{lccc|ccc}
\hline
 & \multicolumn{3}{c|}{Random matrix} & \multicolumn{3}{c}{Diagonal matrix} \\
\cline{2-7}
Op. & de-en & fr-en & ar-en & de-en & fr-en & ar-en \\
\hline
$\odot$ & 22.98\textsubscript{-2.8} & 24.85\textsubscript{-2.1} & 37.96\textsubscript{-2.9} 
        & 21.64\textsubscript{-4.2} & 24.11\textsubscript{-2.9} & 37.34\textsubscript{-3.3 } \\
\hline
\end{tabular}
\end{table}

The results of these experiments show that the network still learns, albeit with degraded performance. We conclude that non-sensical attribution maps can indeed reduce results, even when a pattern, such as a quasi-diagonal attribution matrix, is present, and the increase is not accidental. There should be some meaningful alignment encoded by the XAI attribution maps.

While a linguistic and qualitative analysis of the differences between each attribution method and whether there is a gold standard alignment in which these attributions approximate is outside the scope of this work, we would like to examine some topological differences and similarities among the attribution matrices. For this reason, we treat each column of the matrices as a probability distribution, from which we can calculate entropy as a measure of confidence in selecting the most important source tokens for the target tokens. Our manual inspection using heatmap visualization, such as Figure \ref{fig:att-heatmap}, showed that higher-scoring attribution methods exhibit more concentrated values around some tokens, and we observed more linear and diagonal patterns.

\begin{figure}[ht]
    \centering
    \begin{minipage}[b]{0.32\textwidth}
        \centering
        \includegraphics[width=\textwidth]{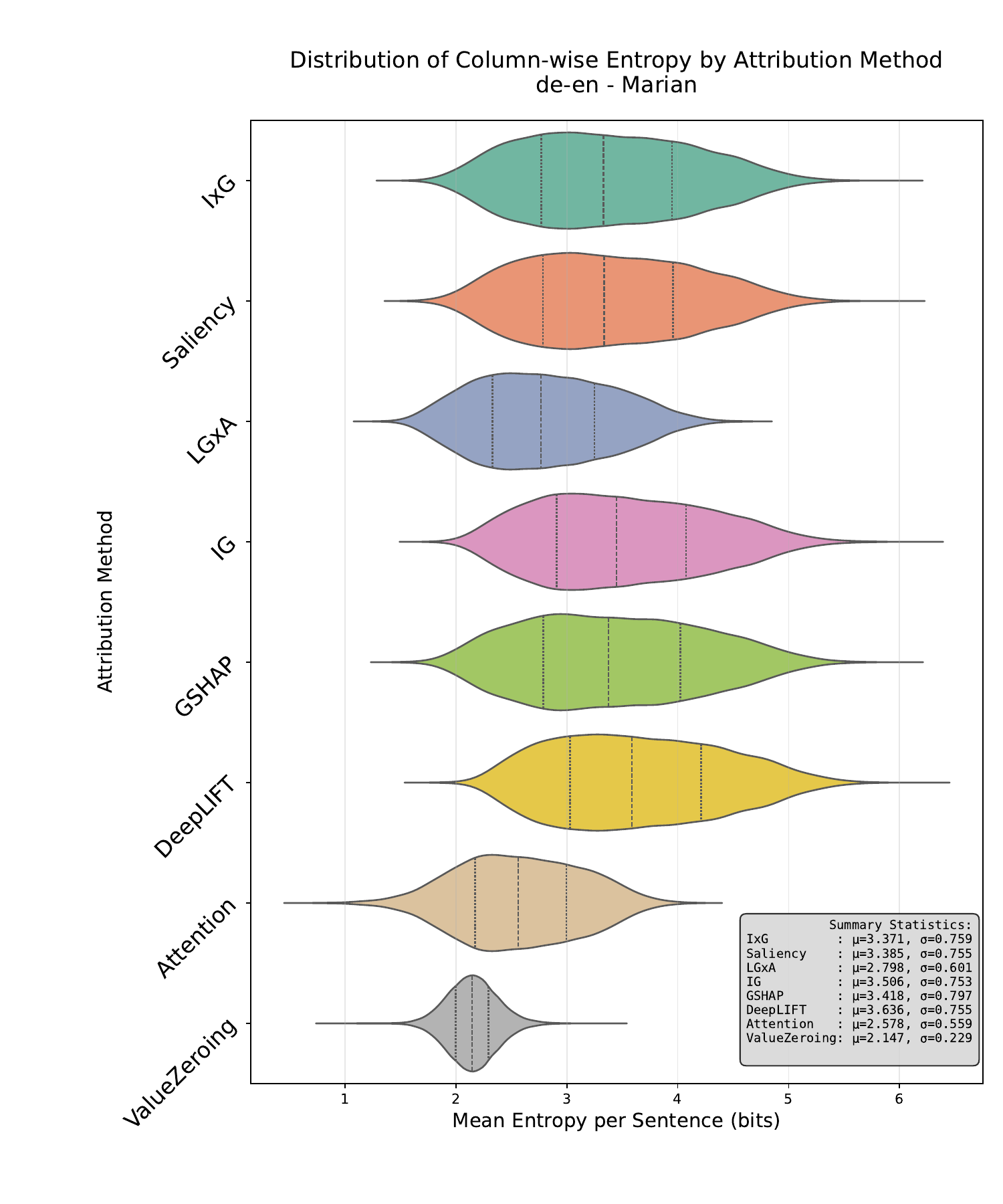}
    \end{minipage}
    \hfill
    \begin{minipage}[b]{0.32\textwidth}
        \centering
        \includegraphics[width=\textwidth]{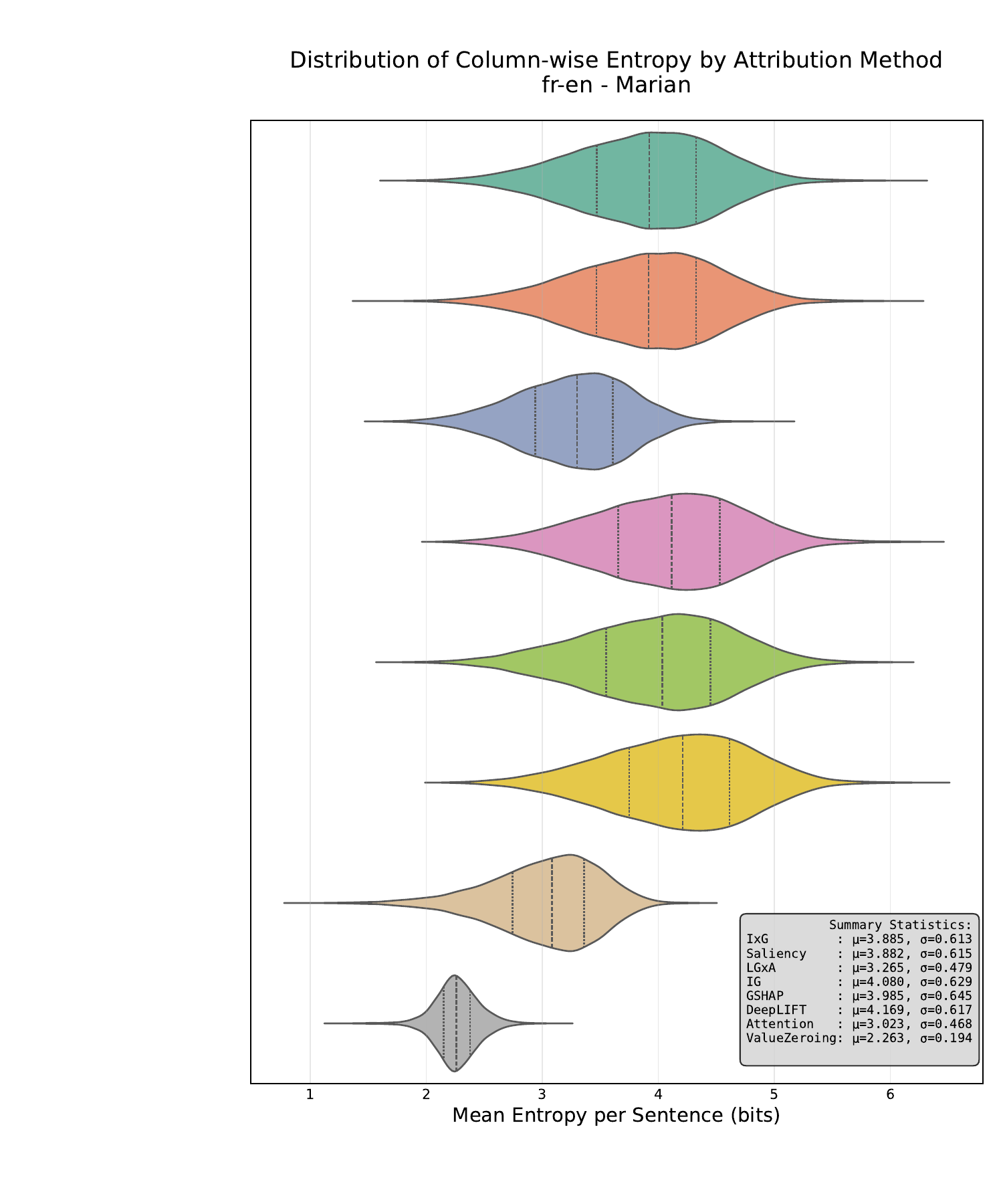}
    \end{minipage}
    \hfill
    \begin{minipage}[b]{0.32\textwidth}
        \centering
        \includegraphics[width=\textwidth]{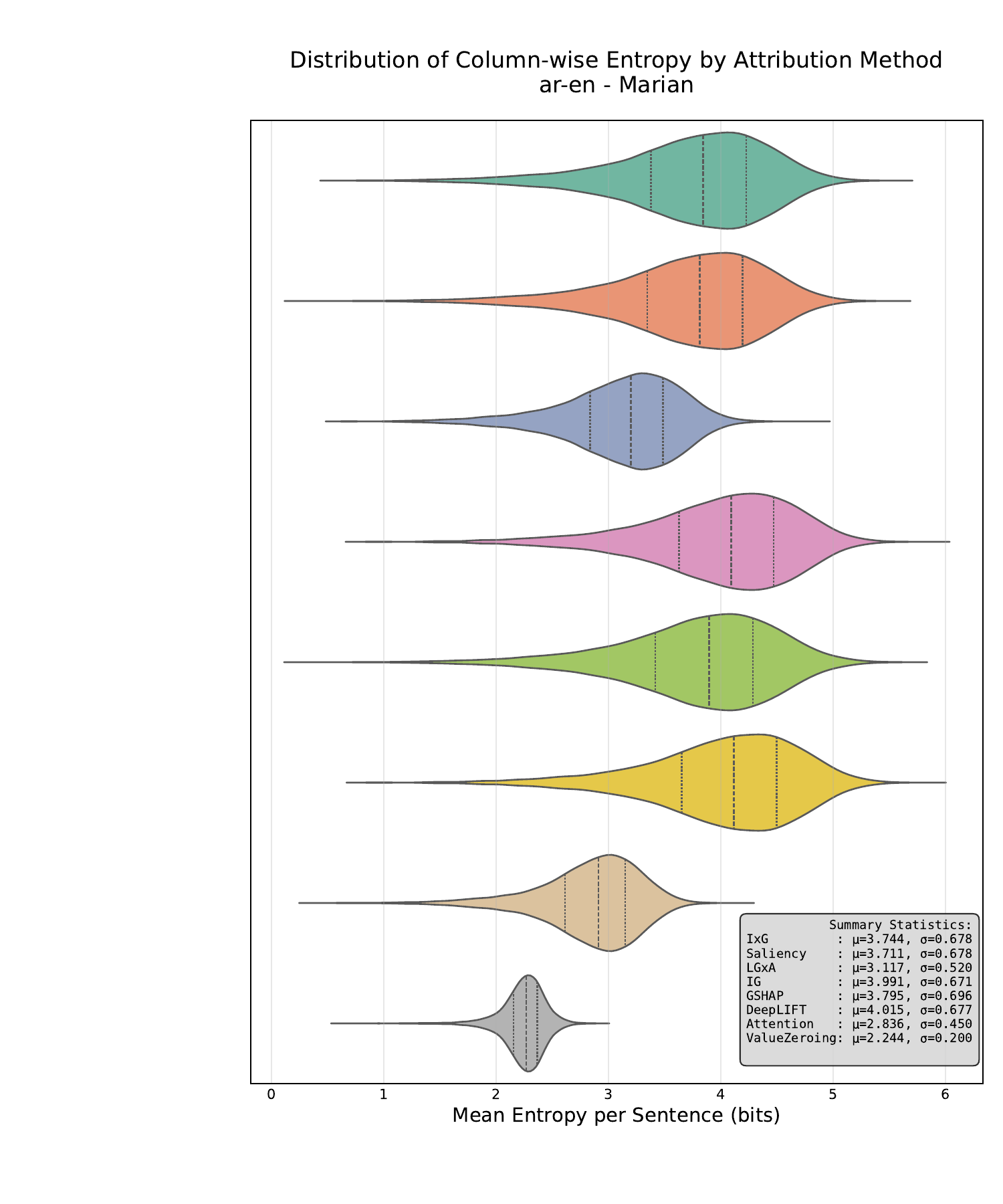}
    \end{minipage}
    \caption{Column-wise Entropy based on Marian-MT attributions.}
    \label{fig:entropy_marian}
\end{figure}

\begin{figure}[ht]
    \centering
    \begin{minipage}[b]{0.32\textwidth}
        \centering
        \includegraphics[width=\textwidth]{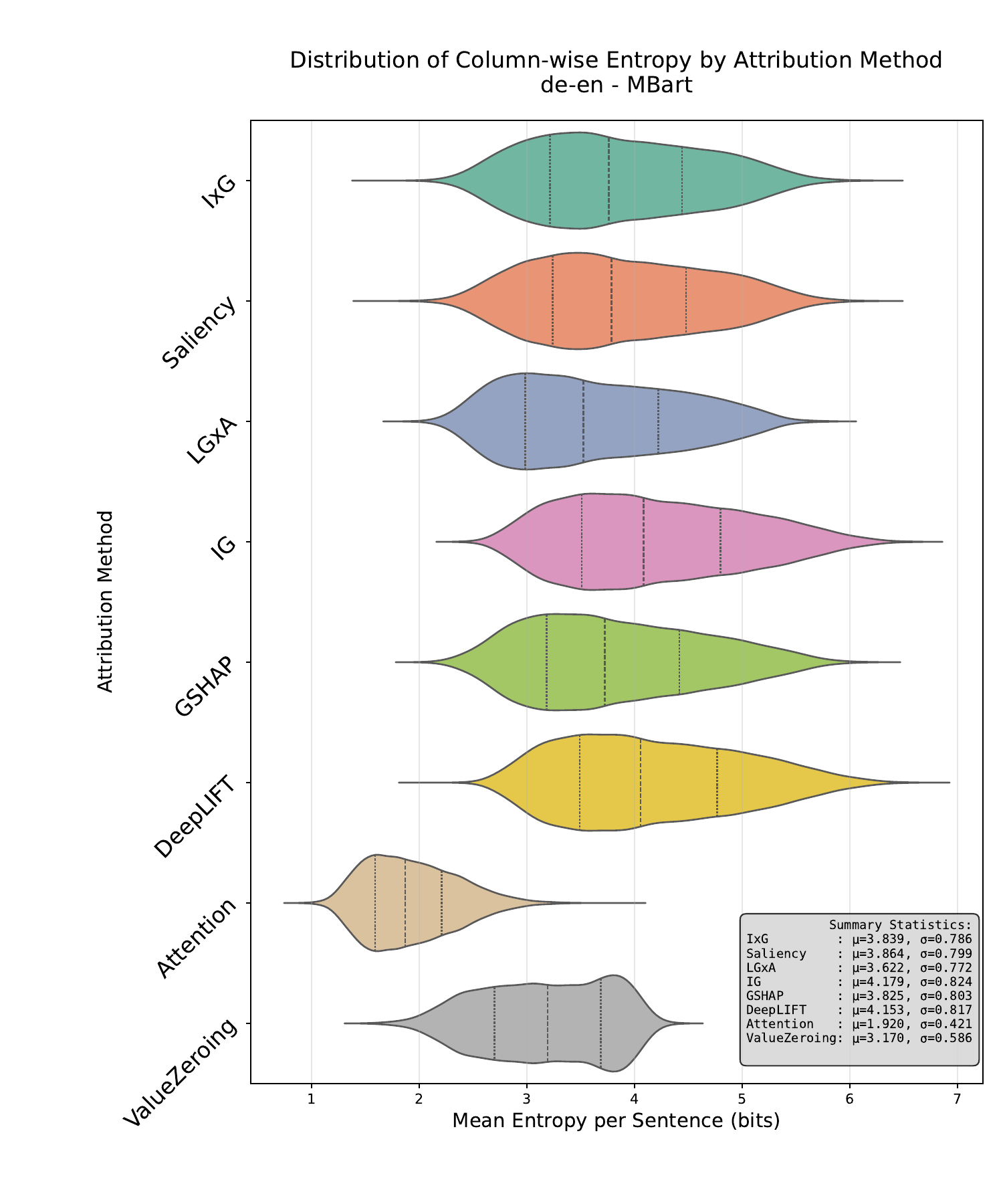}
    \end{minipage}
    \hfill
    \begin{minipage}[b]{0.32\textwidth}
        \centering
        \includegraphics[width=\textwidth]{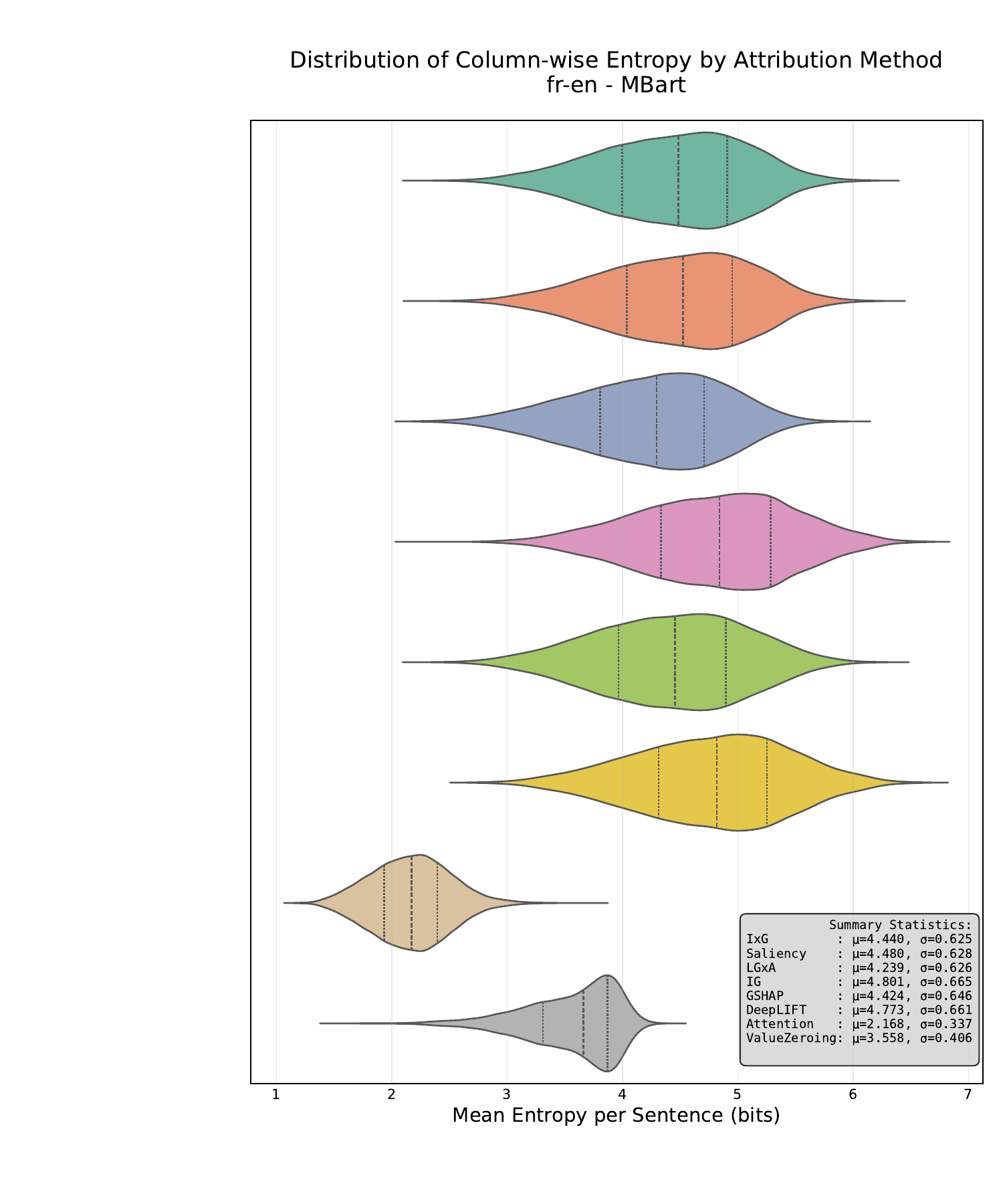}
    \end{minipage}
    \hfill
    \begin{minipage}[b]{0.32\textwidth}
        \centering
        \includegraphics[width=\textwidth]{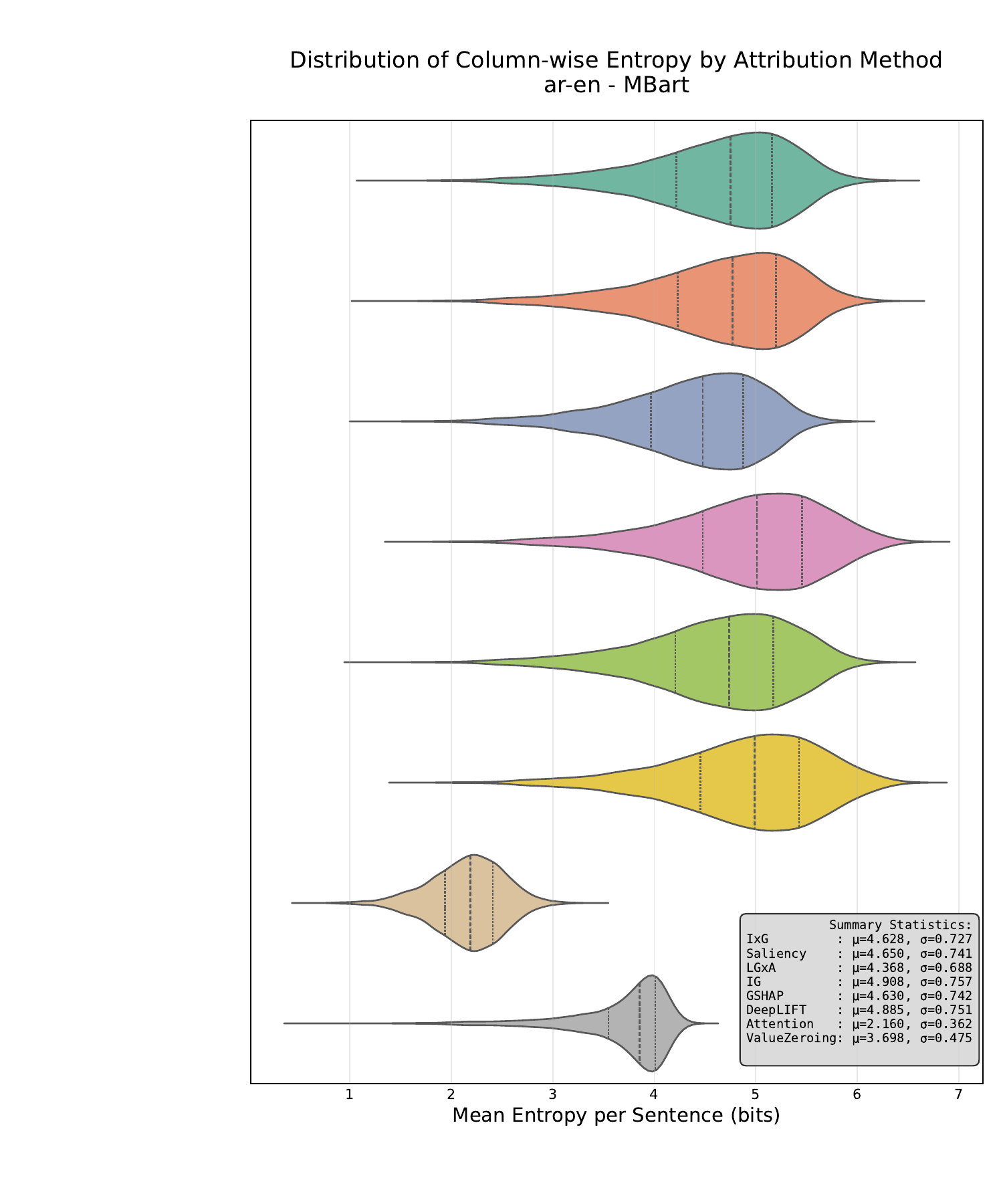}
    \end{minipage}
    \caption{Column-wise Entropy based on mBART attributions.}
    \label{fig:entropy_mbart}
\end{figure}

\begin{figure}[h]
    \centering
    \begin{minipage}[b]{0.32\textwidth}
        \centering
        \includegraphics[width=\textwidth]{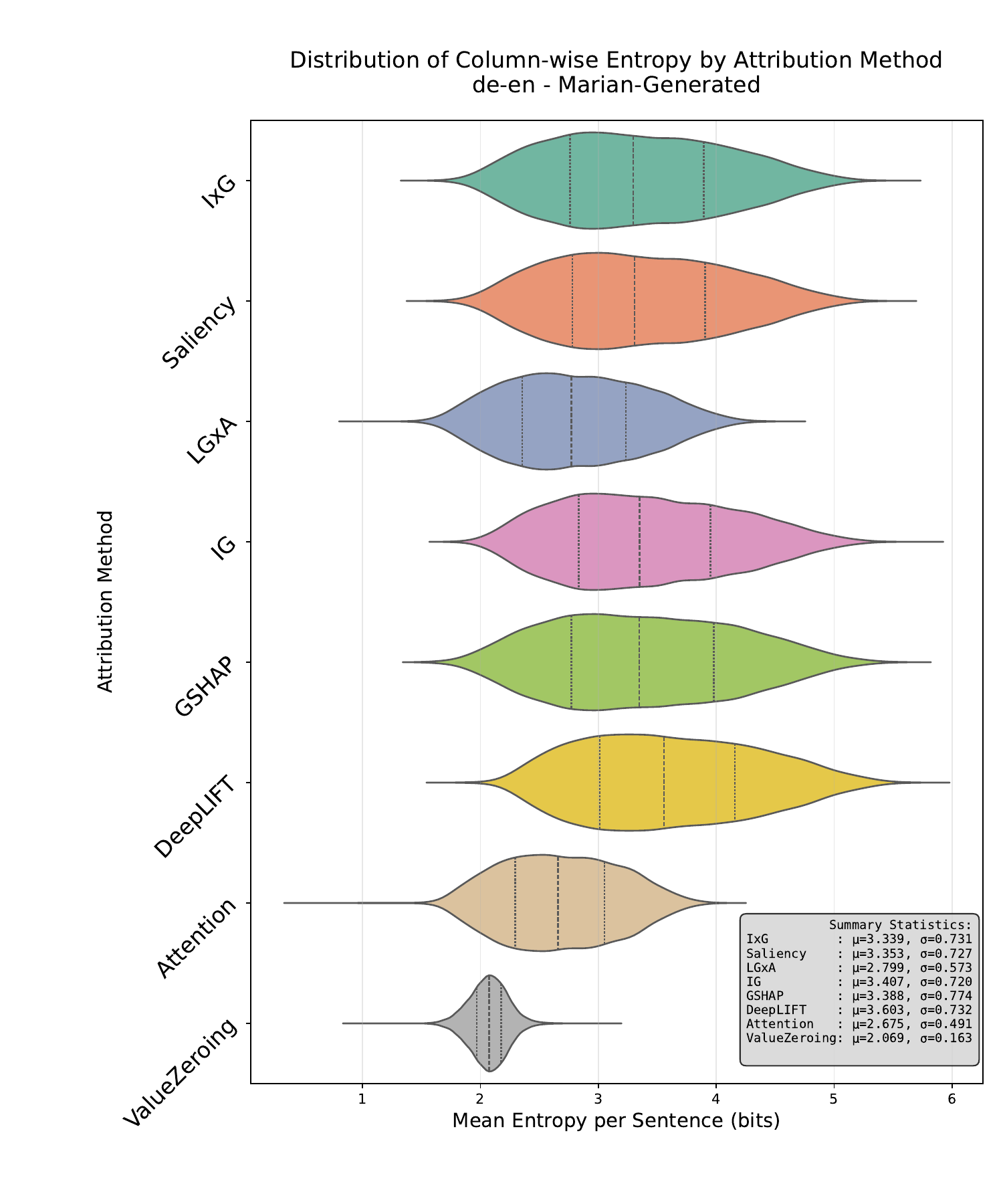}
    \end{minipage}
    \hfill
    \begin{minipage}[b]{0.32\textwidth}
        \centering
        \includegraphics[width=\textwidth]{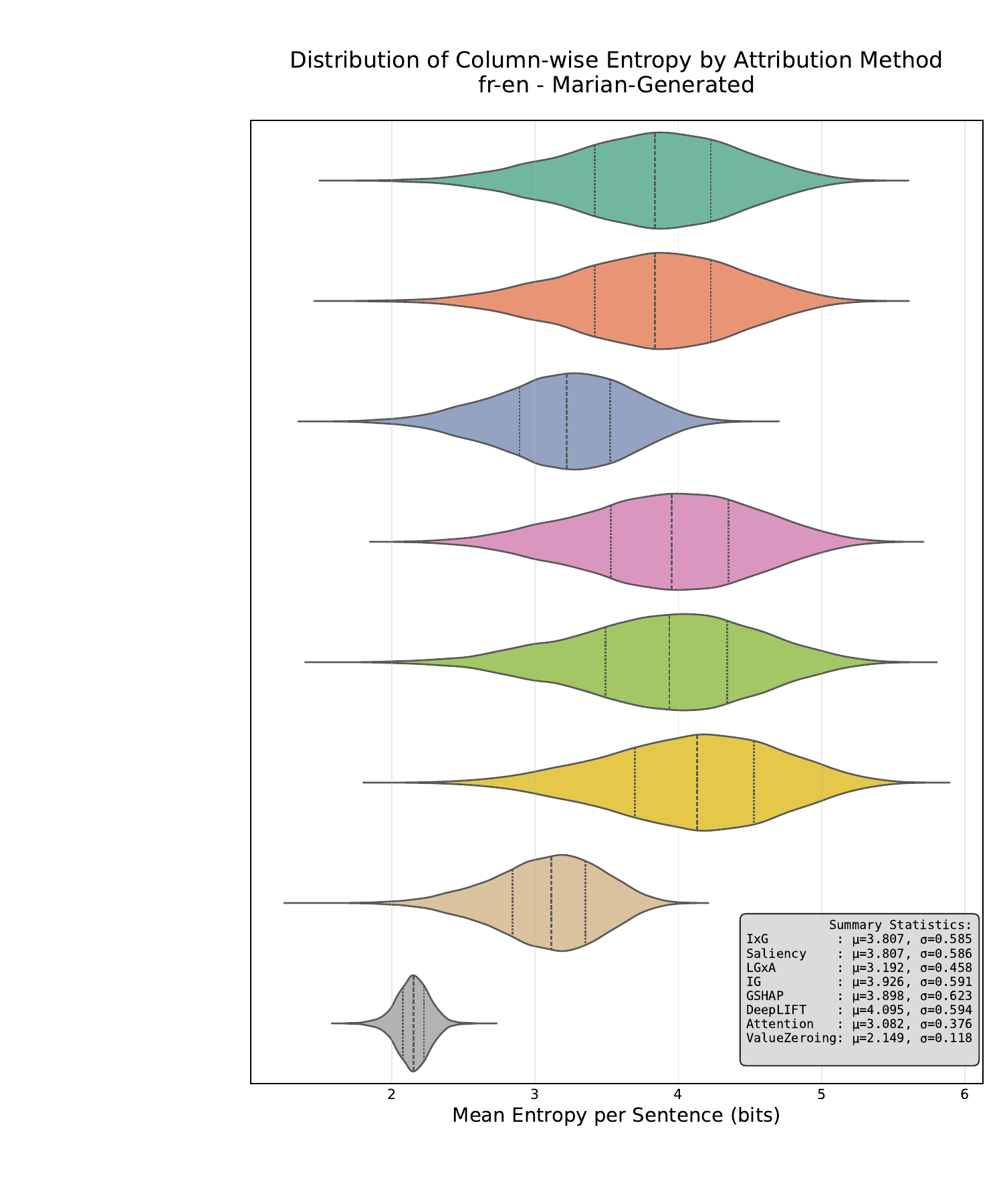}
    \end{minipage}
    \hfill
    \begin{minipage}[b]{0.32\textwidth}
        \centering
        \includegraphics[width=\textwidth]{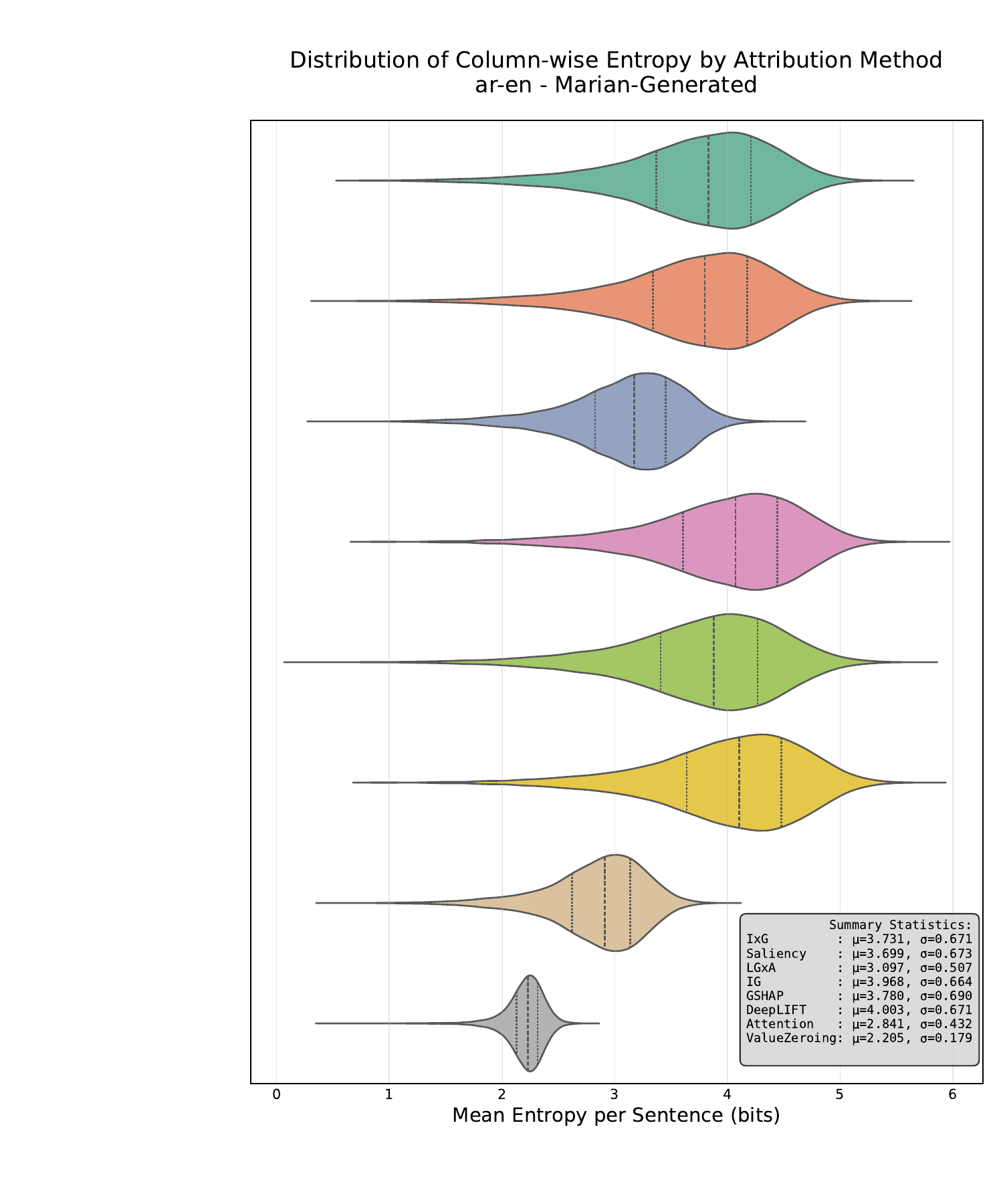}
    \end{minipage}

    \caption{Column-wise entropy based on Marian-MT attributions (Marian-MT-generated targets)}
    \label{fig:entropy_marian_generated}
\end{figure}

Figures \ref{fig:entropy_marian}, \ref{fig:entropy_mbart}, and \ref{fig:entropy_marian_generated} confirm that the three higher attribution maps, ValueZeroing, Attention, and LG$\times$A, have lower entropies compared to other methods. For mBART, the Entropy of gradient-based methods is higher than that of Marian-MT. One plausible explanation is that, in a deeper model with a longer computational graph, gradient signals become less sharp as they propagate toward earlier layers~\autocite{balduzzi2017shattered}. In particular, LG$\times$A at the last layer of the encoder is more informative according to the results. For mBART, we observe that lower-scoring gradient-based attribution maps exhibit higher entropy than those of Marian-MT. Indeed, we could visually confirm that gradient-based methods have more chaotic representations. One point to note is that Entropy does not necessarily completely correlate with the results. For example, for mBART, although the Entropy of Attention is lower, we observe that the last input token receives the most weight for all target tokens, which, in turn, leads to lower entropy. An observation that was confirmed by~\autocite{kobayashi-etal-2020-attention}, and in general, ValueZeroing scored higher for this model.

\FloatBarrier

\FloatBarrier

\section{Attribution Approximator}
\label{sec:attributore}
From the previous experiments, particularly those involving the attention mechanism (Attention attributions and ValueZeroing), we developed a hypothesis that the usefulness of an attribution method is largely determined by how geometrically close its  maps are to those that a transformer can generate. This hypothesis motivates the introduction of a dedicated \textit{Attributor} network. The Attributor is trained to approximate target–source attribution matrices corresponding to different explanation methods. Concretely, given a source sequence and its target sequence, the Attributor network learns to reconstruct the associated attribution matrix. We implement it as an encoder–decoder transformer so that the function class used to model attributions closely matches the inductive biases of the student translation model itself. Intuitively, if the student model can internally reproduce a particular attribution pattern, then it should also be better positioned to exploit that signal when the same attribution is injected into its attention layers, which in turn leads to more effective attribution-guided translation.

\subsection{Attributor implementation}

Our proposed \textit{Attributor} network is a lightweight encoder–decoder transformer. The source sentence is tokenized and mapped to learned token embeddings, which are combined with learned positional embeddings and then processed by a 3-layer, 8-head self-attention encoder. The target sentence is embedded separately with its own learned positional embeddings and passed through 3 causal decoder layers. A standard triangular mask enforces autoregressive ordering to ensure that each target position attends only to previous target tokens. The resulting contextualized source and target embeddings are then fed into a modified cross-attention layer. For each target position of each attention head it returns a row attention score vector over source positions, which are essentially vectors of the per-head attention score matrices.

More formally, let $H$ denote the number of heads and $S$ the number of source positions. For each
target token $t$, the cross-attention block produces per-head attention vectors
$\vec{s}_{t,h} \in \mathbb{R}^{S}$:
\begin{equation}
    \vec{s}_{t,h} = \frac{\vec{q}_{t,h} K_{h}^\top}{\sqrt{d_k}} + \text{mask}, 
    \qquad h = 1,\dots,H,
\end{equation}
where $\vec{q}_{t,h}, K_{h} \in \mathbb{R}^{d_k}$ are the query and key projections.

To aggregate information across heads, we use a weighted mean of these score vectors across the head dimension.
The weights are produced by a small MLP which maps each contextualized target token embedding $h_t \in \mathbb{R}^{d}$ to a probability vector over heads $p_t$:
\begin{equation}
    \vec{p}_t = \operatorname{softmax}\!\bigl(W_2 \,\mathrm{GELU}(W_1 \vec{h}_t)\bigr)
    \in \mathbb{R}^{H},
\end{equation}
Finally, the combined logit vector over source positions for each target token is obtained as a weighted mean of the per-head attention scores vectors for that target token, after which it is passed to softmax to obtain a distribution $\hat{a}_t$ over source tokens:
\begin{equation}
    \hat{\vec{a}}_t = \operatorname{softmax}(\sum_{h=1}^{H} {p_{t}}_h\, \vec{s}_{t,h}) \in \mathbb{R}^{S}.
\end{equation}
The attribution matrix $\hat{A}$ is essentially stacked $\hat{\vec{a}}_t$ for all $t$.

The objective function during the training is to minimize the average row-wise Kullback--Leibler divergence between the predicted
attribution matrix $\hat{A}$ and the gold attribution matrix $A$, summed over non-padded target
positions:
\begin{equation}
    \mathcal{L}_{\text{KL}} =
    \frac{1}{T_{\text{valid}}}
    \sum_{t \in \mathcal{T}_{\text{valid}}}
    \mathrm{KL}\bigl(A_t \ \vert\vert \ \hat{A}_t\bigr),
\end{equation}
where $A_t$ and $\hat{A}_t$ denote the gold and predicted distributions over source tokens for
target position $t$.

\subsection{Attributor Evaluation}

We trained the Attributor on three datasets derived from our earlier experiments, covering all three language pairs. (a) A dataset constructed from Marian-MT attribution maps paired with gold (human) target translations. (b) A dataset constructed from mBART attribution maps paired with gold (human) target translations. (c) A faithfulness-oriented dataset in which Marian-MT attribution maps are paired with targets generated by the Marian-MT teacher model.

As a part of preprocessing, source and target sentences were tokenized according to the respective model's tokenizers, and attribution maps were normalized in the source dimension to represent a valid probability distribution (to accommodate the KL-divergence loss function). To assess the accuracy of the approximation of their attribution maps by the Attributor, we selected the following metrics:
\begin{itemize}
    \item{\textbf{KL-divergence}}. Being the minimization target, it is the first choice to quantify the difference between the approximated and the gold attributions
    \item{\textbf{Overlap@3}} The overlap between sets of top-3 valued source token indices per target token for the approximated and gold attribution maps. As highlighted by the~\textcite{nourbakhsh2025quantifying} we hypothesized that most of the useful signal attributions per target token comes from a small number of the highest-scoring source tokens. The overlap@3 is calculated as 
    \begin{equation}
        \text{overlap@3}(t) = \frac{| \text{Top}_3(a_t) \cap \text{Top}_3(\hat a_t) |}{3}
    \end{equation} 
    \item{\textbf{Tau@3}} Kendall's $\tau$ calculated on top-3 values over the source dimension of the gold attribution and values with the same indices but from the approximated attributions. This metric accompanies overlap@3 by quantifying the rank correlation. 
\end{itemize}

\subsection{Attributor Results}

\begin{figure}[H]
    \centering
    \includegraphics[width=\linewidth]{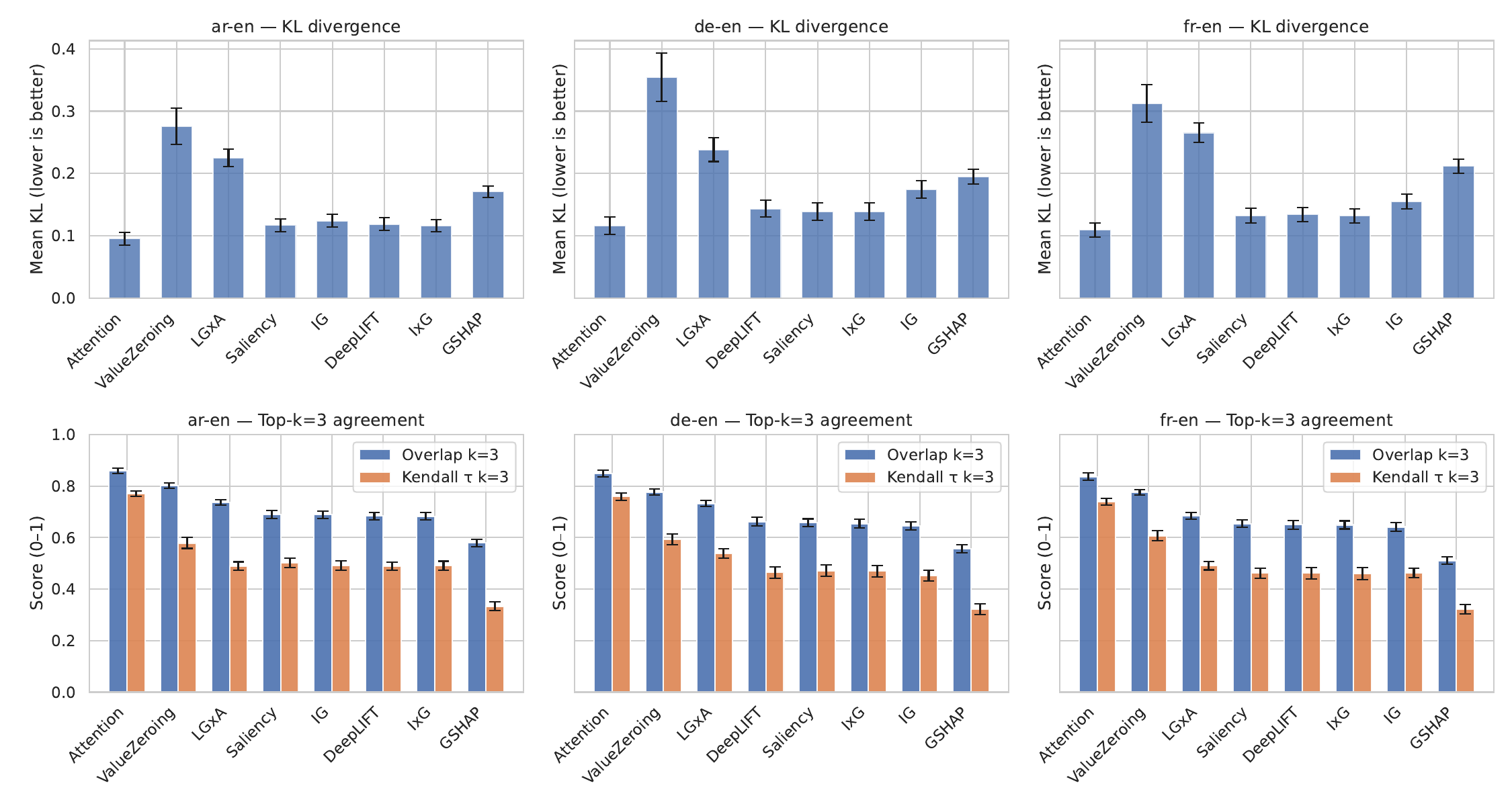}
    \caption{KL-divergence, Overlap@3, and $\tau@3$ for the prediction of attribution on Marian-MT attribution and gold (human) data.}
    \label{fig:graal_human}
\end{figure}

\begin{figure}[H]
    \centering
    \includegraphics[width=\linewidth]{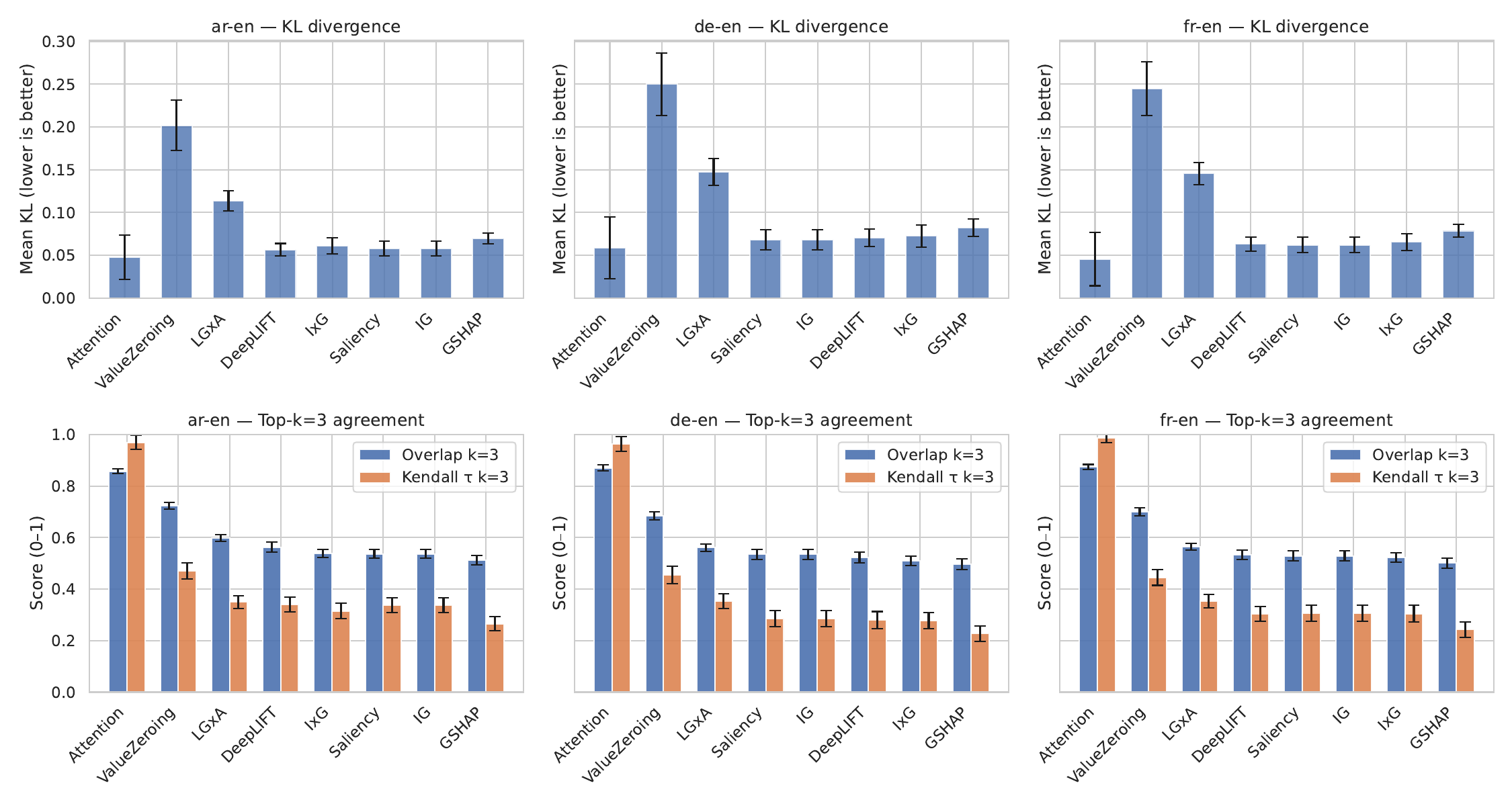}
    \caption{KL-divergence, Overlap@3, and $\tau@3$ for the prediction of attribution on mBART attribution and gold (human) data.}
    \label{fig:graal_mbart}
\end{figure}

\begin{figure}[H]
    \centering
    \includegraphics[width=\linewidth]{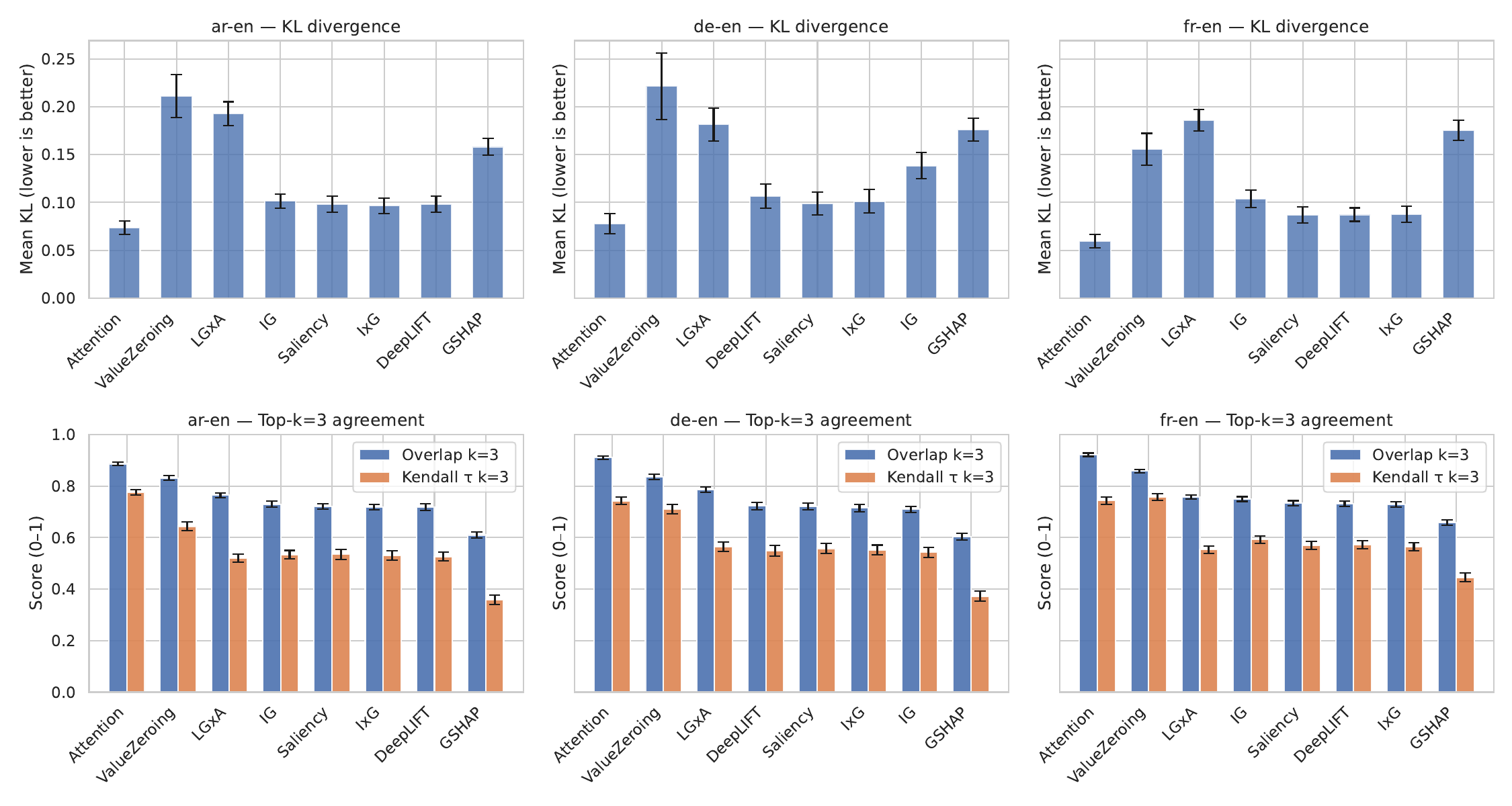}
    \caption{KL-divergence, Overlap@3, and $\tau@3$ for the prediction of attribution on Marian-MT attribution generated target.}
    \label{fig:graal_generated}
\end{figure}

Obtained values for de-en, fr-en, and ar-en pairs are presented in the three settings mentioned above and can be seen in Figures \ref{fig:graal_human}, \ref{fig:graal_mbart}, and \ref{fig:graal_generated}.

\FloatBarrier 

To quantify the link between approximability and downstream gains, we correlate the BLEU scores obtained by Marian-MT when augmented with each attribution method with our three approximability metrics (mean KL divergence, Overlap@3 and Kendall’s $\tau@3$ between gold and predicted target–source maps). We compute BLEU separately for each injection operator (add, average, multiply, replace) and also consider the best BLEU per method across operators.

Across all three language pairs (ar-en, de-en, fr-en) and all operators, BLEU shows a very strong positive correlation with the `top-3' alignment metrics. Pearson’s (r) between BLEU and Overlap@3 lies in the range (r $\approx$ 0.88-0.97), and Kendall’s ($\tau@3$) achieves (r $\approx$ 0.74–0.95). In other words, between roughly 75\% and 90\% of the variance in BLEU across attribution methods can be explained by how well their target–source patterns can be reconstructed by the Attributor. Rank correlations show identical results: Spearman’s ($\rho$) between BLEU and Overlap@3/($\tau@3$) is consistently high (typically $\rho$ $\approx$ 0.65–0.85), and for the fr-en pair with multiplication injection the ranking of methods by BLEU is \emph{exactly} the same as the ranking by $\tau@3$ ($\rho$ = 1.0).

In contrast, KL divergence exhibits only weak and unstable association with BLEU. Pearson correlations between BLEU and mean KL are low-moderate and positive (r $\approx$ 0.27–0.56),  while Spearman’s ($\rho$) tends to be low negative and even changes sign for some operators ($\rho$ $\approx$ -0.26 - 0.11). This behaviour, especially for Pearson, is largely driven by outliers such as ValueZeroing, which combines very high KL divergence with strong BLEU gains. These results suggest that global distribution similarity is a poor predictor of downstream effectiveness, whereas agreement on the locations and ordering of a few top-k source positions is highly predictive.

Taken together, these findings support our central claim: attribution methods whose target-source maps are closer to what an encoder-decoder transformer (Attributor) can reproduce are exactly the ones that yield the largest BLEU gains when injected into the student model. We first measured this closeness with full-distribution KL divergence, but the correlation
analysis shows that the \emph{best predictors of BLEU} are Overlap@3 and Kendall's $\tau@3$ between the Attributor's predictions and the gold attribution maps, which correlate very strongly with BLEU, whereas KL shows only weak and unstable correlations. These results suggest that an attribution method tends to be useful precisely when a transformer can reliably recover the same few most-salient source tokens per target token as the teacher. In contrast, matching the full attention distribution beyond the top positions appears much less informative for predicting BLEU gains.

For the other two settings of generated targets, the manifested trend is the same. For mBART, the highest correlation is observed with the multiplication operator. It should be noted that the approximation accuracy is the highest for Attention from mBART: the last source token consistently gets the highest attribution scores and their reconstruction is easier. However, there are still residual scores for other tokens whose presence can guide the translation. The regression plot of the above approximation metrics for the three language pairs and the corresponding outcomes is provided in the figures \ref{fig:Marian_3x3}, \ref{fig:mBART_3x3}, and \ref{fig:Marian_gen_3x3}. 

In short, the student benefits from what it can imitate: attribution methods for which target tokens a transformer can reliably reconstruct the same top-3 source tokens are precisely those that give student models the largest gains.

\section{Conclusion and Future work} \label{conclusion}
In this work, we demonstrate that XAI attribution maps derived from a learned teacher model can be injected into a student model, and that the resulting changes in translation behavior provide a practical signal of the relative quality of the XAI attribution methods. We conducted an extensive evaluation across German–English, French–English, and Arabic–English language pairs, comparing eight XAI methods and multiple composition strategies for integrating attribution scores into Transformer attention (addition, multiplication, averaging, and replacement). We further examined where the scores are applied (encoder self-attention versus cross-attention) and assessed robustness across two distinct teacher models, mBART and Marian-MT.

Our results indicate that Attention and ValueZeroing, as well as LG$\times$A extracted from the final encoder layer, consistently produced the largest gains in BLEU and chrF. We also observed that injecting source–target attributions into the encoder self-attention can yield significant improvements, which is initially counterintuitive given that the encoder is designed to attend only within the source sequence. A plausible explanation is that the non-auto-regressive encoder attention setup makes future information accessible to the entire sequence. 

We also showed that getting better results from the different attribution methods is not accidental: 1) Attribution maps with lower entropy tend to score higher.2) We set up a transformer model, \textit{Attributor}, which shows that the most beneficial attribution maps are the ones for whose target tokens the transformer model successfully recreates top-3 salient source tokens. These findings offer the NLP community a plausible explanation for the utility of attribution maps and highlight the importance of the attribution's most salient tokens. 

There are some limitations to this work worth noting.
First, this study provides an extensive exploration of attribution transfer between models. However, the overall pipeline is computationally expensive, both in deriving attributions at scale and in retraining student models. For this reason, we focused our comparison on attribution signals produced by a range of explainability methods, most of which are gradient-based. This choice was driven largely by practicality. Extracting attributions with perturbation-based approaches such as LIME~\autocite{ribeiro2016should} and reAGent~\autocite{zhao2024reagentmodelagnosticfeatureattribution} is substantially more resource-intensive and time-consuming. In addition, some methods available in Inseq produce (self-)attributions on the decoder side of seq2seq models. In this work, we restrict our experiments to encoder self-attention and encoder–decoder cross-attention, leaving decoder-side attribution transfer for future study.

More ablation studies are needed. In this work, we tried with cross-attention, 4 attention heads, and different operators. However, the effects of layer-wise injection, broader choices of head selection and head count, and alternative placement strategies remain underexplored. We also used a coarse attribution extraction procedure. We took attention-based attributions from all the layers and averaged attention scores, and for the gradient methods, our only alteration was the last encoder layer. More systematic investigation of layer-wise signals and aggregation strategies is needed. Moreover, it is important to note that, for the Attributor experiments, we used the top-3 tokens for overlap@3 and Kendall’s $\tau$; it may also be beneficial to extend the experiments to other top-k tokens to draw a line for the most salient tokens more precisely. Also, in this work, our focus was limited to machine translation tasks. Machine translation tasks are characterized by the alignments between source and target pairs. Future work could extend this evaluation framework to other generative models, including those applied in question answering and text summarization.

\paragraph{Funding Statement.}
This work is supported by the Luxembourg National Research Fund (FNR) as part of the project C21 - Collaboration 21:
IPBG2020/IS/14839977/C21.

\paragraph{Competing Interests.} The authors declare that the manuscript complies with the ethical standards of the journal and that there are no competing interests to disclose.

\paragraph{Acknowledgment.} We thank Richard Albrecht for early assistance and helpful discussions during project initiation.

\paragraph{Artificial Intelligence Usage.} We used Large Language Models, such as ChatGPT and Copilot, for code autocompletion, proofreading, and spellchecking of the manuscript.



\begingroup
  \sloppy                      
  \setlength\emergencystretch{3em}
  \printbibliography
\endgroup

\clearpage
\FloatBarrier

\appendix
\label{sec:appendix}

\begin{figure}[t]
\centering

\begin{subfigure}{\textwidth}
\centering
\begin{minipage}{0.32\textwidth}\centering
\includegraphics[width=\linewidth]{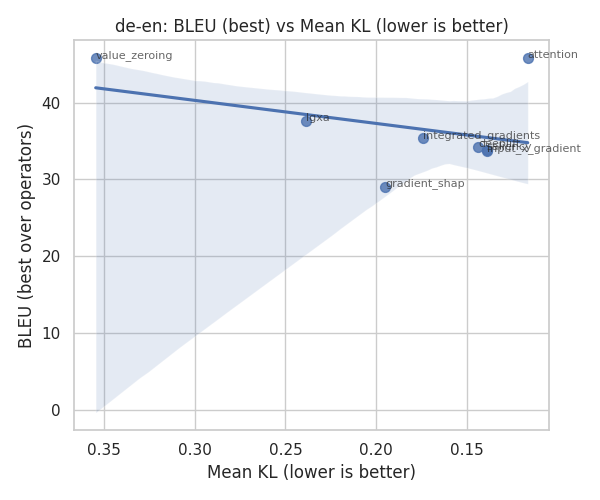}
\end{minipage}\hfill
\begin{minipage}{0.32\textwidth}\centering
\includegraphics[width=\linewidth]{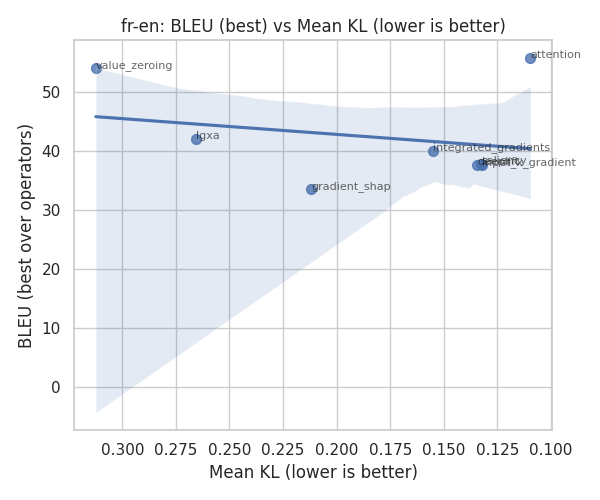}
\end{minipage}\hfill
\begin{minipage}{0.32\textwidth}\centering
\includegraphics[width=\linewidth]{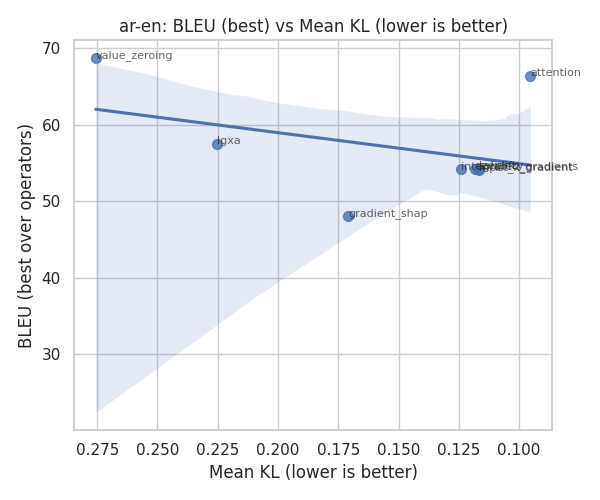}
\end{minipage}
\subcaption{Mean KL (de-en, fr-en, ar-en).}
\label{fig:row_kl}
\end{subfigure}

\vspace{0.6em}

\begin{subfigure}{\textwidth}
\centering
\begin{minipage}{0.32\textwidth}\centering
\includegraphics[width=\linewidth]{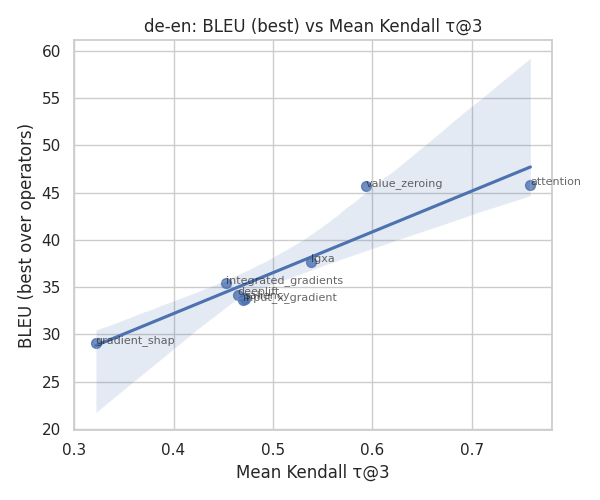}
\end{minipage}\hfill
\begin{minipage}{0.32\textwidth}\centering
\includegraphics[width=\linewidth]{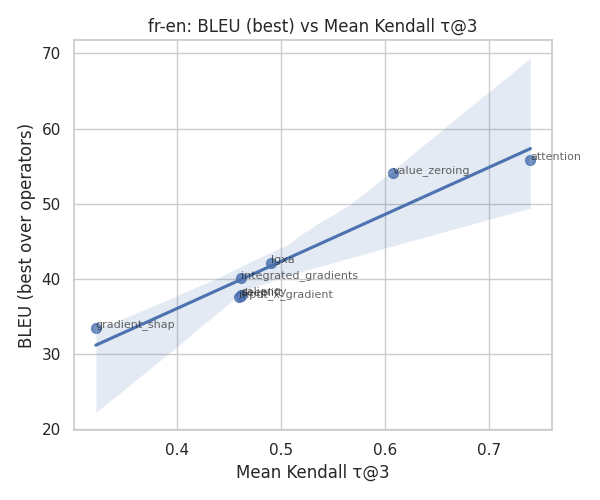}
\end{minipage}\hfill
\begin{minipage}{0.32\textwidth}\centering
\includegraphics[width=\linewidth]{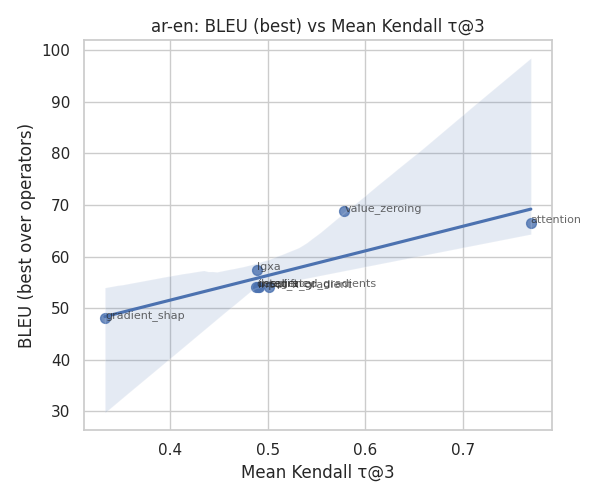}
\end{minipage}
\subcaption{Mean Kendall $\tau$@3 (de-en, fr-en, ar-en).}
\label{fig:row_kts}
\end{subfigure}

\vspace{0.6em}

\begin{subfigure}{\textwidth}
\centering
\begin{minipage}{0.32\textwidth}\centering
\includegraphics[width=\linewidth]{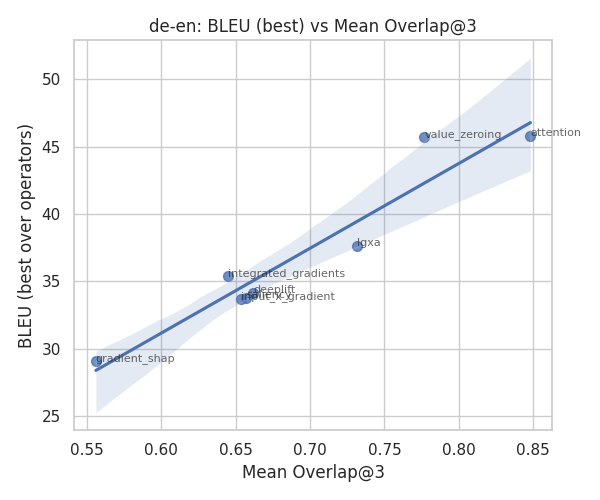}
\end{minipage}\hfill
\begin{minipage}{0.32\textwidth}\centering
\includegraphics[width=\linewidth]{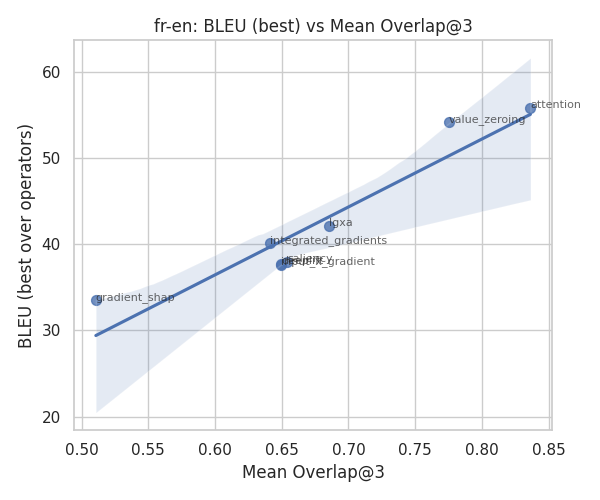}
\end{minipage}\hfill
\begin{minipage}{0.32\textwidth}\centering
\includegraphics[width=\linewidth]{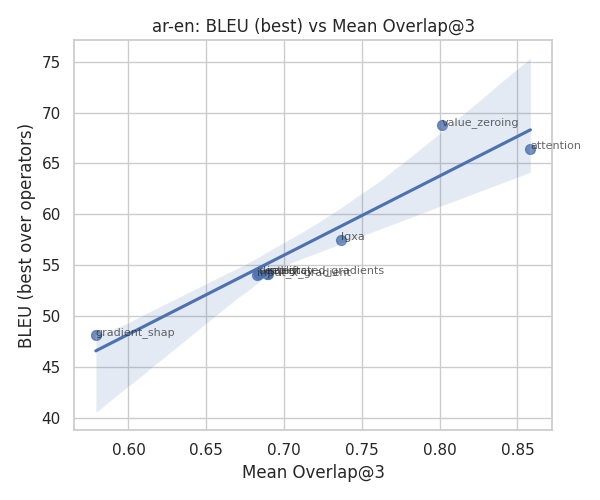}
\end{minipage}
\subcaption{Mean Overlap@3 (de-en, fr-en, ar-en).}
\label{fig:row_overlap}
\end{subfigure}

\caption{Regression plots of the Marian-MT attributions with gold (human) target sentence.}
\label{fig:Marian_3x3}
\end{figure}


\begin{figure}[t]
\centering

\begin{subfigure}{\textwidth}
\centering
\begin{minipage}{0.32\textwidth}\centering
\includegraphics[width=\linewidth]{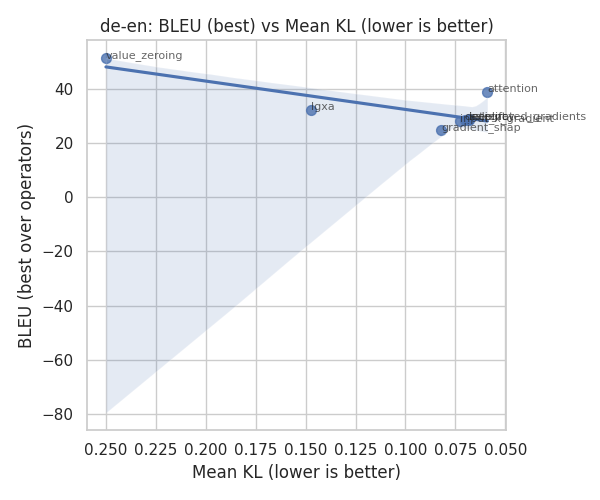}
\end{minipage}\hfill
\begin{minipage}{0.32\textwidth}\centering
\includegraphics[width=\linewidth]{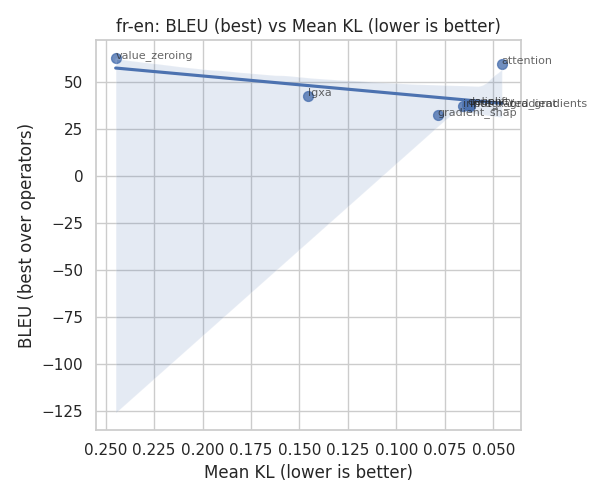}
\end{minipage}\hfill
\begin{minipage}{0.32\textwidth}\centering
\includegraphics[width=\linewidth]{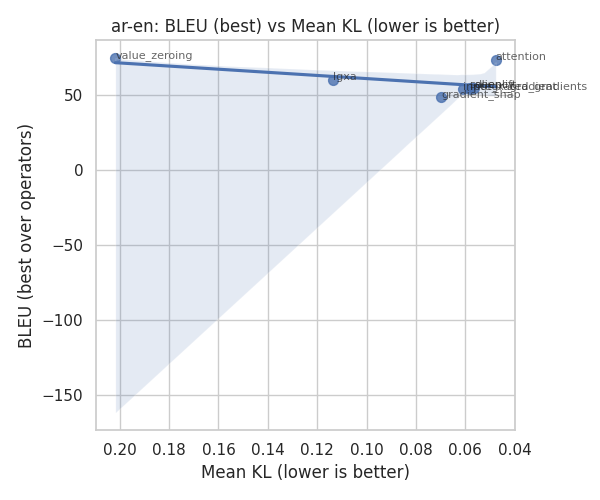}
\end{minipage}
\subcaption{Mean KL (de-en, fr-en, ar-en).}
\label{fig:row_kl}
\end{subfigure}

\vspace{0.6em}

\begin{subfigure}{\textwidth}
\centering
\begin{minipage}{0.32\textwidth}\centering
\includegraphics[width=\linewidth]{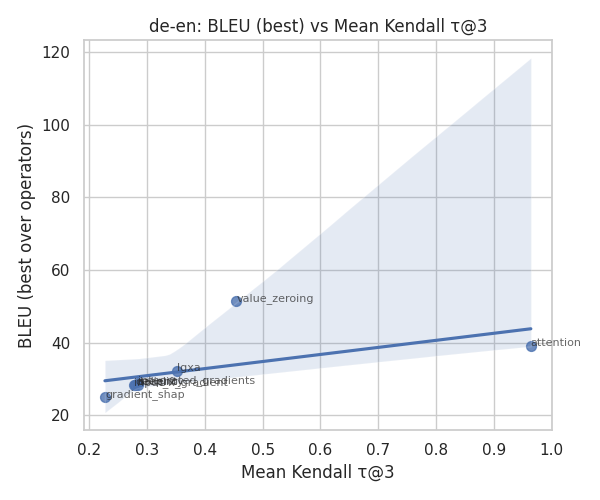}
\end{minipage}\hfill
\begin{minipage}{0.32\textwidth}\centering
\includegraphics[width=\linewidth]{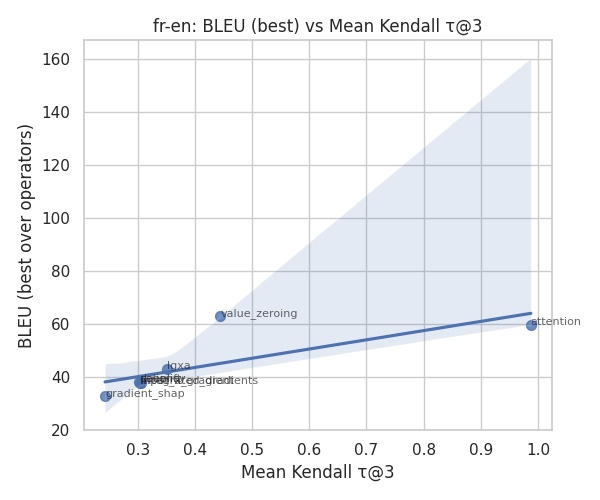}
\end{minipage}\hfill
\begin{minipage}{0.32\textwidth}\centering
\includegraphics[width=\linewidth]{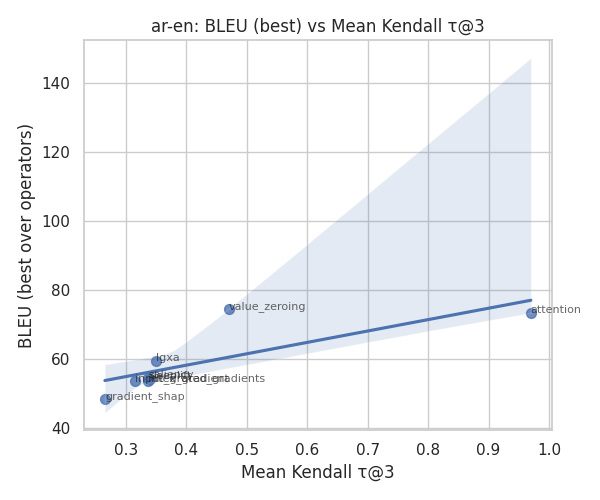}
\end{minipage}
\subcaption{Mean Kendall $\tau$@3 (de-en, fr-en, ar-en).}
\label{fig:row_kts}
\end{subfigure}

\vspace{0.6em}

\begin{subfigure}{\textwidth}
\centering
\begin{minipage}{0.32\textwidth}\centering
\includegraphics[width=\linewidth]{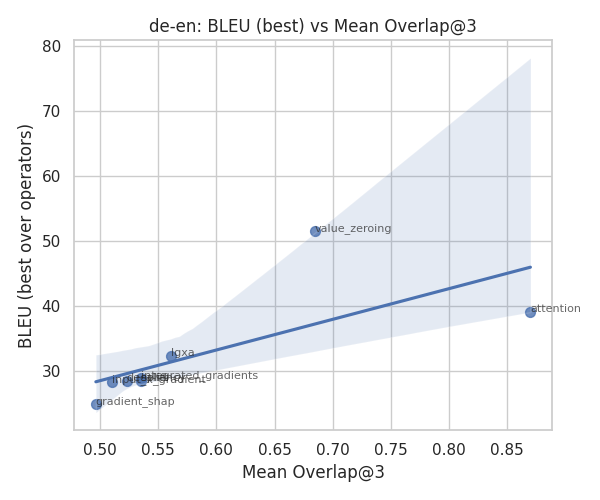}
\end{minipage}\hfill
\begin{minipage}{0.32\textwidth}\centering
\includegraphics[width=\linewidth]{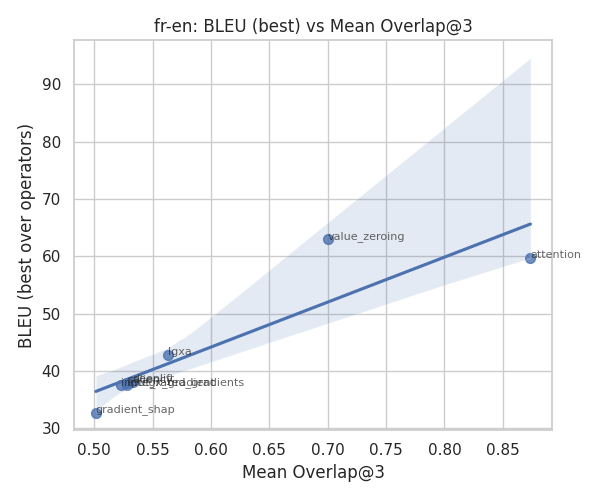}
\end{minipage}\hfill
\begin{minipage}{0.32\textwidth}\centering
\includegraphics[width=\linewidth]{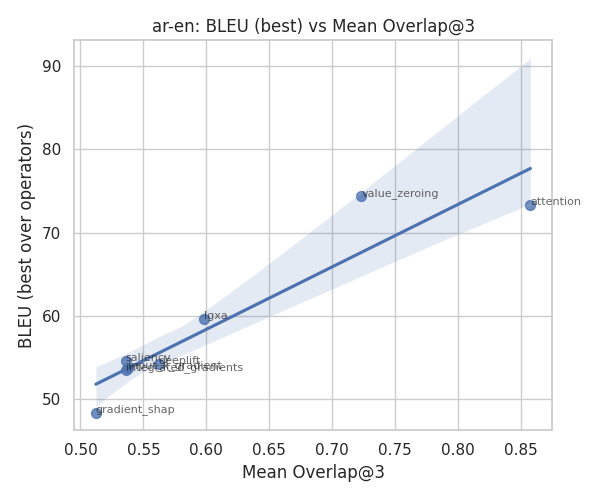}
\end{minipage}
\subcaption{Mean Overlap@3 (de-en, fr-en, ar-en).}
\label{fig:row_overlap}
\end{subfigure}

\caption{Regression plots of the mBART attributions with gold (human) target sentence.}
\label{fig:mBART_3x3}
\end{figure}

\begin{figure}[t]
\centering

\begin{subfigure}{\textwidth}
\centering
\includegraphics[width=\linewidth]{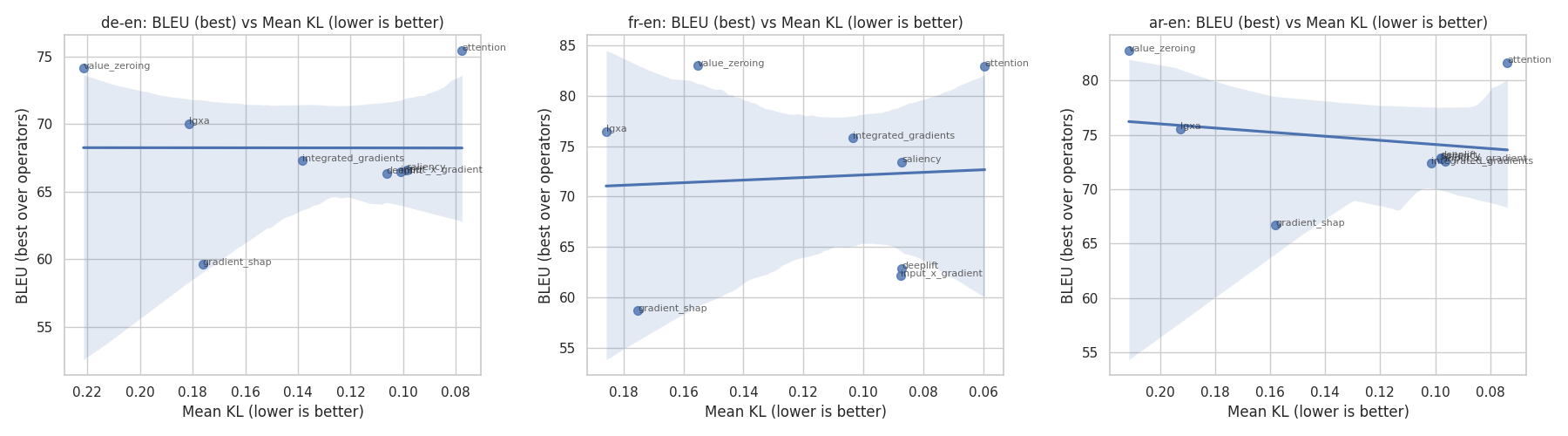}
\subcaption{Mean KL (de-en, fr-en, ar-en).}
\label{fig:row_kl}
\end{subfigure}

\vspace{0.6em}

\begin{subfigure}{\textwidth}
\centering
\includegraphics[width=\linewidth]{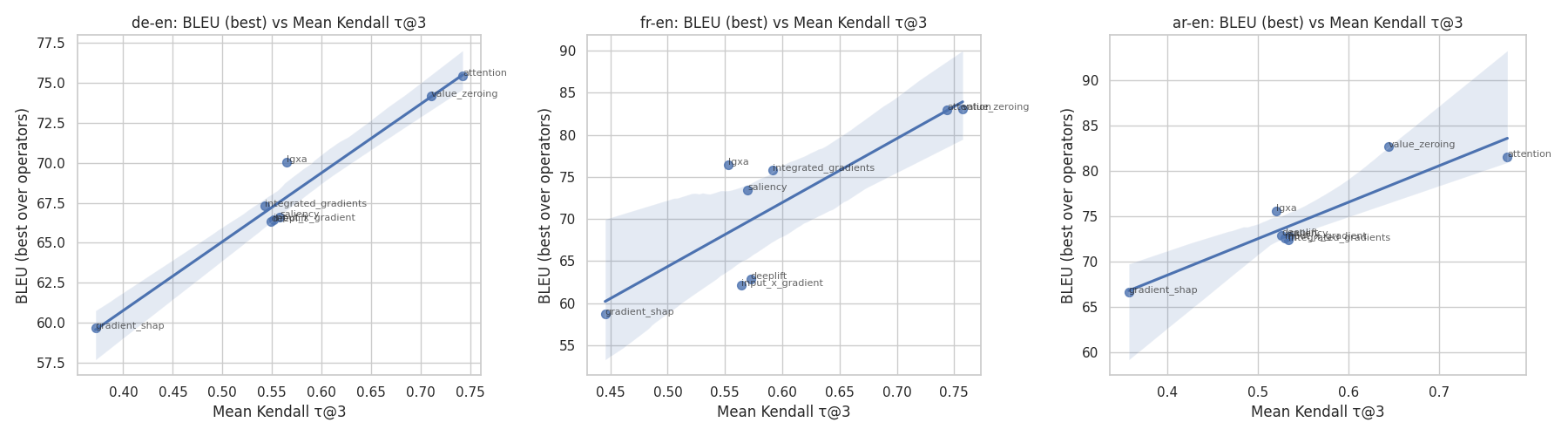}
\subcaption{Mean Kendall $\tau$@3 (de-en, fr-en, ar-en).}
\label{fig:row_kts}
\end{subfigure}

\vspace{0.6em}

\begin{subfigure}{\textwidth}
\centering
\includegraphics[width=\linewidth]{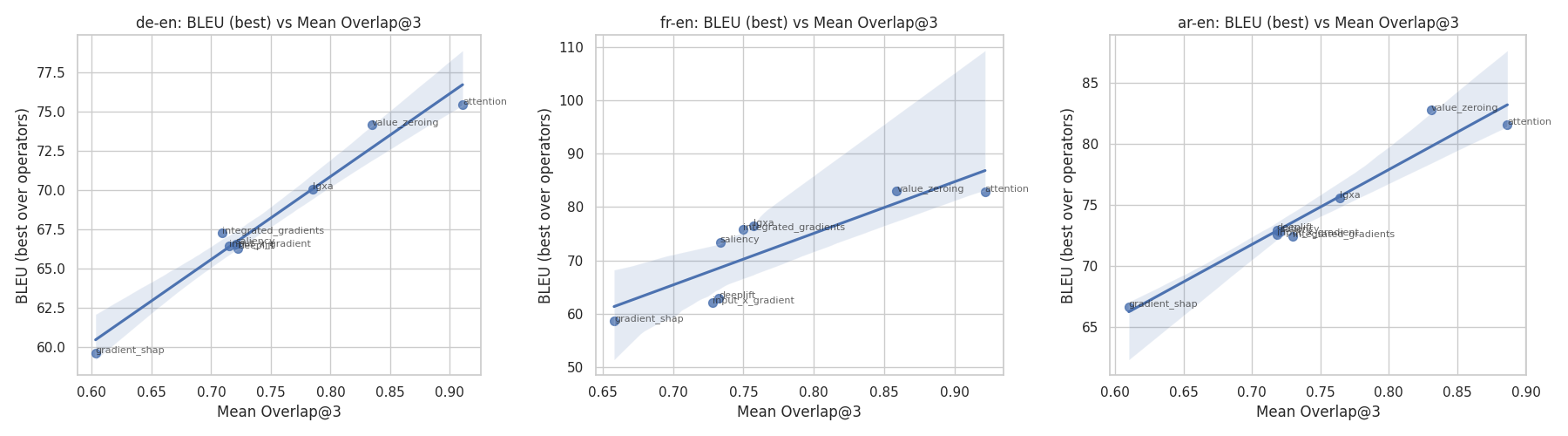}
\subcaption{Mean Overlap@3 (de-en, fr-en, ar-en).}
\label{fig:row_overlap}
\end{subfigure}

\caption{Regression plot of the Marian-MT attributions with generated target sentences.}
\label{fig:Marian_gen_3x3}
\end{figure}


\FloatBarrier
\begin{table}[t]
\centering
\caption{BLEU and chrF scores for de-en Marian-MT 4 head attribution injection. Scores followed by $\Delta$ over the baseline.}
\label{tab:de_en_bleu_enc_4h}
\setlength{\tabcolsep}{4.5pt} 
\begin{tabular}{lcccccccc}
\hline
\multicolumn{9}{c}{BLEU scores for de-en (Baseline: 25.82)} \\
\hline
Op. & I$\times$G & Saliency & LG$\times$A & IG & GSHAP & DeepLIFT & Attention & ValueZeroing \\
\hline
$+$ & 32.7\textsubscript{+6.8} & 32.4\textsubscript{+6.6} & 35.7\textsubscript{+9.9} & 33.7\textsubscript{+7.8} & 27.8\textsubscript{+2.0} & 32.9\textsubscript{+7.1} & 40.0\textsubscript{+14.1} & 41.8\textsubscript{+16.0} \\
$\mu$ & 29.9\textsubscript{+4.0} & 29.6\textsubscript{+3.8} & 33.5\textsubscript{+7.7} & 32.1\textsubscript{+6.3} & 26.2\textsubscript{+0.4} & 30.5\textsubscript{+4.7} & 36.1\textsubscript{+10.2} & 40.5\textsubscript{+14.7} \\
$\odot$ & 34.4\textsubscript{+8.5} & 34.7\textsubscript{+8.9} & 38.3\textsubscript{+12.5} & 36.0\textsubscript{+10.2} & 29.9\textsubscript{+4.0} & 34.6\textsubscript{+8.8} &\cellcolor{green!20} 47.2\textsubscript{+21.4} & 44.8\textsubscript{+18.9} \\
R & 32.4\textsubscript{+6.6} & 32.3\textsubscript{+6.5} & 35.9\textsubscript{+10.1} & 33.8\textsubscript{+8.0} & 27.9\textsubscript{+2.1} & 32.9\textsubscript{+7.1} & 40.2\textsubscript{+14.4} & 41.7\textsubscript{+15.8} \\
\hline
\multicolumn{9}{c}{chrF scores for de-en (Baseline: 49.27)} \\
\hline
Op. & I$\times$G & Saliency & LG$\times$A & IG & GSHAP & DeepLIFT & Attention & ValueZeroing \\
\hline
$+$ & 53.6\textsubscript{+4.4} & 53.4\textsubscript{+4.2} & 55.8\textsubscript{+6.6} & 54.3\textsubscript{+5.1} & 50.3\textsubscript{+1.0} & 53.8\textsubscript{+4.5} & 58.3\textsubscript{+9.1} & 60.0\textsubscript{+10.7} \\
$\mu$ & 51.5\textsubscript{+2.3} & 51.3\textsubscript{+2.0} & 54.1\textsubscript{+4.8} & 53.1\textsubscript{+3.8} & 49.1\textsubscript{-0.2} & 52.0\textsubscript{+2.8} & 55.5\textsubscript{+6.2} & 58.9\textsubscript{+9.6} \\
$\odot$ & 54.9\textsubscript{+5.6} & 55.1\textsubscript{+5.8} & 57.7\textsubscript{+8.4} & 56.0\textsubscript{+6.7} & 51.7\textsubscript{+2.4} & 55.0\textsubscript{+5.7} & \cellcolor{green!20}63.6\textsubscript{+14.4} & 62.0\textsubscript{+12.7} \\
R & 53.5\textsubscript{+4.2} & 53.4\textsubscript{+4.1} & 56.0\textsubscript{+6.7} & 54.4\textsubscript{+5.1} & 50.4\textsubscript{+1.1} & 53.8\textsubscript{+4.6} & 58.5\textsubscript{+9.2} & 59.9\textsubscript{+10.6} \\
\hline
\end{tabular}
\end{table}

\begin{table}[t]
\centering
\caption{BLEU and chrF scores for fr-en Marian-MT 4 head attribution injection. Scores followed by $\Delta$ over the baseline.}
\label{tab:fr_en_bleu_enc_4h}
\setlength{\tabcolsep}{4.5pt} 
\begin{tabular}{lcccccccc}
\hline
\multicolumn{9}{c}{BLEU scores for fr-en (Baseline: 27.01)} \\
\hline
Op. & I$\times$G & Saliency & LG$\times$A & IG & GSHAP & DeepLIFT & Attention & ValueZeroing \\
\hline
$+$ & 34.9\textsubscript{+7.9} & 34.6\textsubscript{+7.6} & 38.5\textsubscript{+11.5} & 37.5\textsubscript{+10.5} & 31.2\textsubscript{+4.2} & 35.1\textsubscript{+8.1} & 44.3\textsubscript{+17.3} & 47.5\textsubscript{+20.5} \\
$\mu$ & 32.1\textsubscript{+5.1} & 33.1\textsubscript{+6.1} & 34.9\textsubscript{+7.9} & 34.1\textsubscript{+7.1} & 28.5\textsubscript{+1.5} & 32.8\textsubscript{+5.8} & 39.9\textsubscript{+12.9} & 44.9\textsubscript{+17.9} \\
$\odot$ & 38.2\textsubscript{+11.2} & 37.9\textsubscript{+10.9} & 42.3\textsubscript{+15.3} & 40.8\textsubscript{+13.8} & 34.1\textsubscript{+7.1} & 38.4\textsubscript{+11.4} & \cellcolor{green!20}56.1\textsubscript{+29.1} & 53.7\textsubscript{+26.7} \\
R & 35.3\textsubscript{+8.3} & 35.4\textsubscript{+8.4} & 38.9\textsubscript{+11.9} & 37.8\textsubscript{+10.8} & 31.1\textsubscript{+4.1} & 35.7\textsubscript{+8.7} & 43.5\textsubscript{+16.5} & 46.8\textsubscript{+19.8} \\
\hline
\multicolumn{9}{c}{chrF scores for fr-en (Baseline: 53.01)} \\
\hline
Op. & I$\times$G & Saliency & LG$\times$A & IG & GSHAP & DeepLIFT & Attention & ValueZeroing \\
\hline
$+$ & 57.2\textsubscript{+4.1} & 58.6\textsubscript{+5.6} & 59.7\textsubscript{+6.7} & 58.8\textsubscript{+5.8} & 54.8\textsubscript{+1.8} & 57.4\textsubscript{+4.4} & 62.9\textsubscript{+9.9} & 65.3\textsubscript{+12.3} \\
$\mu$ & 55.4\textsubscript{+2.4} & 55.9\textsubscript{+2.9} & 57.2\textsubscript{+4.2} & 56.6\textsubscript{+3.6} & 53.1\textsubscript{+0.1} & 55.7\textsubscript{+2.7} & 59.8\textsubscript{+6.7} & 63.4\textsubscript{+10.4} \\
$\odot$ & 59.4\textsubscript{+6.4} & 60.8\textsubscript{+7.8} & 62.1\textsubscript{+9.1} & 61.0\textsubscript{+7.9} & 56.6\textsubscript{+3.6} & 59.4\textsubscript{+6.4} &\cellcolor{green!20} 70.8\textsubscript{+17.8} & 68.9\textsubscript{+15.9} \\
R & 57.4\textsubscript{+4.4} & 57.5\textsubscript{+4.5} & 59.9\textsubscript{+6.9} & 59.0\textsubscript{+6.0} & 54.8\textsubscript{+1.8} & 57.7\textsubscript{+4.7} & 62.3\textsubscript{+9.3} & 65.0\textsubscript{+11.9} \\
\hline
\end{tabular}
\end{table}

\begin{table}[t]
\centering
\caption{BLEU and chrF scores for ar-en Marian-MT 4 head attribution injection. Scores followed by $\Delta$ over the baseline.}
\label{tab:ar_en_bleu_enc_4h}
\setlength{\tabcolsep}{4.5pt} 
\begin{tabular}{lcccccccc}
\hline
\multicolumn{9}{c}{BLEU scores for ar-en (Baseline: 40.68)} \\
\hline
Op. & I$\times$G & Saliency & LG$\times$A & IG & GSHAP & DeepLIFT & Attention & ValueZeroing \\
\hline
$+$ & 51.9\textsubscript{+11.1} & 51.6\textsubscript{+10.8} & 54.9\textsubscript{+14.1} & 51.8\textsubscript{+11.0} & 46.1\textsubscript{+5.3} & 52.7\textsubscript{+11.9} & 60.1\textsubscript{+19.3} & 63.0\textsubscript{+22.2} \\
$\mu$ & 47.5\textsubscript{+6.7} & 47.3\textsubscript{+6.5} & 50.3\textsubscript{+9.5} & 47.0\textsubscript{+6.2} & 43.8\textsubscript{+3.0} & 47.8\textsubscript{+7.0} & 54.6\textsubscript{+13.8} & 59.8\textsubscript{+19.0} \\
$\odot$ & 54.9\textsubscript{+14.1} & 55.0\textsubscript{+14.2} & 58.1\textsubscript{+17.3} & 54.9\textsubscript{+14.0} & 49.0\textsubscript{+8.2} & 55.1\textsubscript{+14.3} & 66.9\textsubscript{+26.1} & \cellcolor{green!20}68.1\textsubscript{+27.3} \\
R & 51.9\textsubscript{+11.1} & 51.5\textsubscript{+10.7} & 55.1\textsubscript{+14.3} & 51.8\textsubscript{+11.0} & 46.2\textsubscript{+5.4} & 52.3\textsubscript{+11.5} & 60.7\textsubscript{+19.9} & 63.3\textsubscript{+22.5} \\
\hline
\multicolumn{9}{c}{chrF scores for ar-en (Baseline: 66.07)} \\
\hline
Op. & I$\times$G & Saliency & LG$\times$A & IG & GSHAP & DeepLIFT & Attention & ValueZeroing \\
\hline
$+$ & 71.4\textsubscript{+5.3} & 71.2\textsubscript{+5.1} & 73.2\textsubscript{+7.1} & 71.3\textsubscript{+5.3} & 68.1\textsubscript{+2.0} & 71.8\textsubscript{+5.8} & 75.8\textsubscript{+9.7} & 77.8\textsubscript{+11.7} \\
$\mu$ & 68.9\textsubscript{+2.8} & 68.8\textsubscript{+2.7} & 70.4\textsubscript{+4.4} & 68.6\textsubscript{+2.6} & 66.9\textsubscript{+0.9} & 69.1\textsubscript{+3.0} & 72.6\textsubscript{+6.5} & 75.8\textsubscript{+9.7} \\
$\odot$ & 73.0\textsubscript{+6.9} & 73.1\textsubscript{+7.0} & 74.9\textsubscript{+8.8} & 72.9\textsubscript{+6.9} & 69.7\textsubscript{+3.6} & 73.1\textsubscript{+7.0} & 79.9\textsubscript{+13.9} & \cellcolor{green!20}80.7\textsubscript{+14.7} \\
R & 71.4\textsubscript{+5.3} & 71.2\textsubscript{+5.1} & 73.3\textsubscript{+7.2} & 71.3\textsubscript{+5.2} & 68.1\textsubscript{+2.0} & 71.6\textsubscript{+5.5} & 76.2\textsubscript{+10.1} & 77.9\textsubscript{+11.8} \\
\hline
\end{tabular}
\end{table}
\FloatBarrier

\begin{table}[t]
\centering
\small
\setlength{\tabcolsep}{4pt}
\begin{tabular}{l l r r}
\hline
\textbf{Model} & \textbf{Text} & \textbf{BLEU} & \textbf{chrF} \\
\hline
\multicolumn{4}{l}{\textbf{Marian-MT}} \\
\hline
Source:  & Ich dachte, dass es damit beendet wäre. & -- & -- \\
\hline
Reference:  & I thought that would be the end of it. & -- & -- \\
Baseline:  & I thought it was finished. &  9.91 & 29.19 \\
I$\times$G &  I thought it would have been finished with it. & 12.55 & 43.76 \\
Saliency &  I thought it would have been finished with it. & 12.55 & 43.76 \\
LG$\times$A &  I thought that it had been finished with it. & 19.64 & 47.02 \\
IG &  I thought it would have been over with it. & 12.55 & 43.29 \\
GSHAP &  I thought it would be finished with it. & 14.78 & 47.72 \\
DeepLIFT &  I thought it would have been finished with it. & 12.55 & 43.76 \\
Attention &  I thought that would have an end to it. & 32.47 & 61.15 \\
ValueZeroing &  I thought this might be the end of it. & 58.14 & 65.49 \\
\hline
\multicolumn{4}{l}{\textbf{mBART}} \\
\hline
Reference: &  I thought that would be the end of it. & -- & -- \\
I$\times$G &  I thought it'd be done to it. & 12.86 & 33.19 \\
Saliency &  I thought it'd be done to it. & 12.86 & 33.19 \\
LG$\times$A &  I thought it'd be done with it. & 12.86 & 33.28 \\
IG &  I thought it'd be done to it. & 12.86 & 33.19 \\
GSHAP &  I thought it was the only way it was. & 10.55 & 29.46 \\
DeepLIFT &  I thought it would end to it for it. & 13.13 & 41.71 \\
Attention &  I thought it would be the end of it. & 70.71 & 80.71 \\
ValueZeroing &  I thought it would be the end over it. & 35.49 & 68.34 \\
\hline
\multicolumn{4}{l}{\textbf{Marian-MT (Generated)}} \\
\hline
Reference: &   I thought it was over. & -- & -- \\
I$\times$G &  I thought it would be. & 32.47 & 57.06 \\
Saliency &   I thought it was over. & 100 & 100 \\
LG$\times$A &  I thought it was over. & 100 & 100 \\
IG &   I thought it would end. &  32.47& 56.33 \\
GSHAP &   I thought it would be. &  32.47 & 57.06 \\
DeepLIFT &   I thought it would be. & 32.47 & 57.06 \\
Attention &   I thought it was over. & 100 & 100 \\
ValueZeroing &  I thought it was over. & 100 & 100 \\
\hline
\end{tabular}
\caption{Sample translations and metrics for Marian-MT and mBART under different attribution/XAI variants for a sample de-en.}
\label{tab:xai_samples_id163}
\end{table}


\end{document}